\documentclass[runningheads]{llncs}

\usepackage{eccv}

\usepackage{eccvabbrv}

\usepackage{graphicx}
\usepackage{booktabs}

\usepackage{multirow}
 \usepackage{subfloat}
\usepackage{graphicx}
\usepackage{amsmath}
\usepackage{amssymb}
\usepackage{booktabs}
\usepackage{enumitem}
\usepackage{times}
\usepackage{epsfig}
\usepackage{enumitem}
\usepackage{graphicx} 
\usepackage{booktabs} 
\usepackage{multirow} 
\usepackage{hyperref}

\usepackage{orcidlink}

\begin{document}

\title{Diffscaler: Enhancing the Generative Prowess of Diffusion Transformers} 

\titlerunning{Diffscaler}


\author{Nithin Gopalakrishnan Nair\inst{1}\orcidlink{0000-1111-2222-3333} \and
Jeya Maria Jose Valanarasu\inst{2} \orcidlink{0000-0001-6424-7529} \and
Vishal M Patel\inst{1}\orcidlink{0000-0002-5239-692X}}

\authorrunning{Nair et al.}

\institute{Johns Hopkins University$^{1}$, Stanford University$^{2}$ \\
\email{\{ngopala2, vpatel36\}@jhu.edu, jmjose@stanford.edu}\\
}

\definecolor{cvprblue}{rgb}{0.21,0.49,0.74}

\maketitle

\begin{abstract}

Recently, diffusion transformers have gained wide attention with the release of SORA from OpenAI, emphasizing the need for transformers as backbone for diffusion models. Transformer-based models have shown better generalization capability compared to CNN-based models for general vision tasks. However, much less has been explored in the existing literature regarding the capabilities of transformer-based diffusion backbones and expanding their generative prowess to other datasets. This paper focuses on enabling a single pre-trained diffusion transformer model to scale across multiple datasets swiftly, allowing for the completion of diverse generative tasks using just one model. To this end, we propose \textit{DiffScaler}, an efficient scaling strategy for diffusion models where we train a minimal amount of parameters to adapt to different tasks. In particular, we learn task-specific transformations at each layer by incorporating the ability to utilize the learned subspaces of the pre-trained model, as well as the ability to learn additional task-specific subspaces, which may be absent in the pre-training dataset. As these parameters are independent, a single diffusion model with these task-specific parameters can be used to perform multiple tasks simultaneously. Moreover, we find that transformer-based diffusion models significantly outperform CNN-based diffusion models methods while performing fine-tuning over smaller datasets. We perform experiments on four unconditional image generation datasets. We show that using our proposed method, a single pre-trained model can scale up to perform these conditional and unconditional tasks, respectively, with minimal parameter tuning while performing as close as fine-tuning an entire diffusion model for that particular task.


\end{abstract}

\section{Introduction}

Denoising Diffusion Probabilistic Models (DDPMs) \cite{ho2020denoising} have transformed the content creation research by achieving photorealistic generation quality across different image generation tasks.  Methods such as Diffusion Transformers (DiTs) \cite{peebles2022scalable}, Latent Diffusion Models (LDM) \cite{rombach2022high}, and Ablated Diffusion Models (ADM) \cite{dhariwal2021diffusion} have achieved state-of-the-art image generation quality on various datasets related to faces (Ex: FFHQ \cite{karras2019style}), buildings (Ex: LSUN-Churches \cite{yu2015lsun}), animals (Ex: LSUN-CATS\cite{yu2015lsun}, CUB-200 \cite{welinder2010caltech}), and more. It should be noted that to achieve optimal performance, these models are trained separately on each dataset. There has also been development in terms of text or modality conditional generative models where image generation is controlled by an  
\begin{center}
\centering
\setlength{\tabcolsep}{0.5pt}
\captionsetup{type=figure}
{\footnotesize
\renewcommand{\arraystretch}{0.5} 
\begin{tabular}{c c c c c c c c c c c c}
    \tabularnewline
        \textit{(a) Incremental Spatial Task Conditioning}\\


 \includegraphics[width=0.16\linewidth]{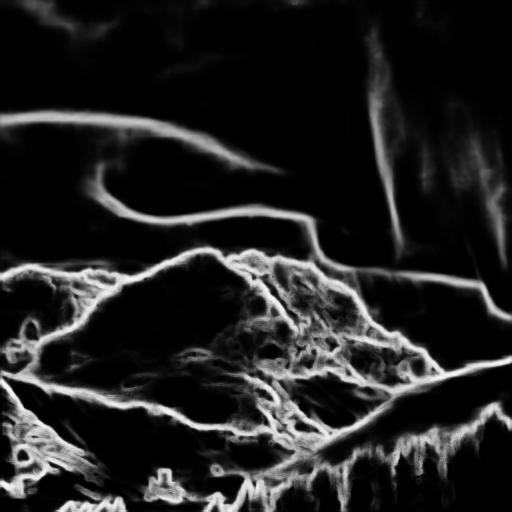}
 \includegraphics[width=0.16\linewidth]{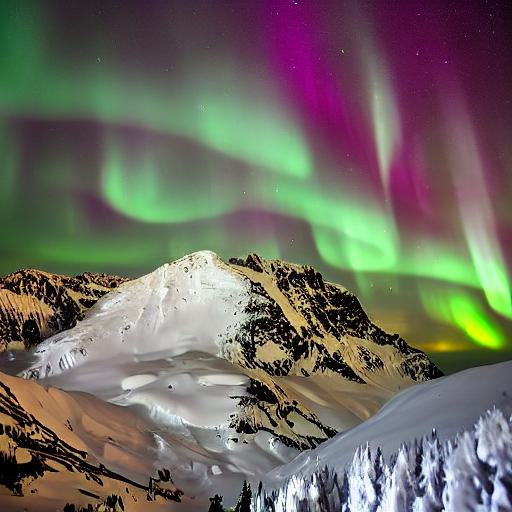}
 \includegraphics[width=0.16\linewidth]{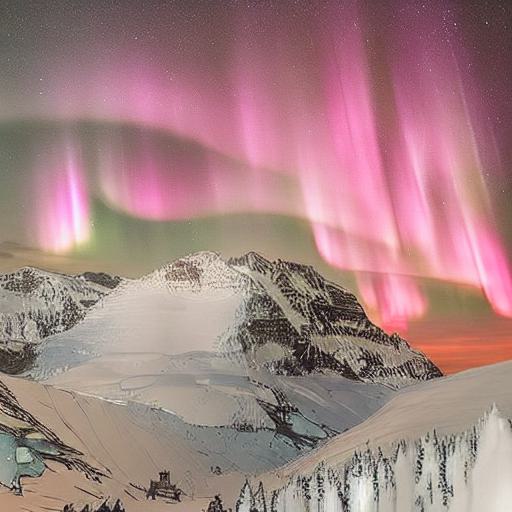}
\hspace{2mm}
 \includegraphics[width=0.16\linewidth]{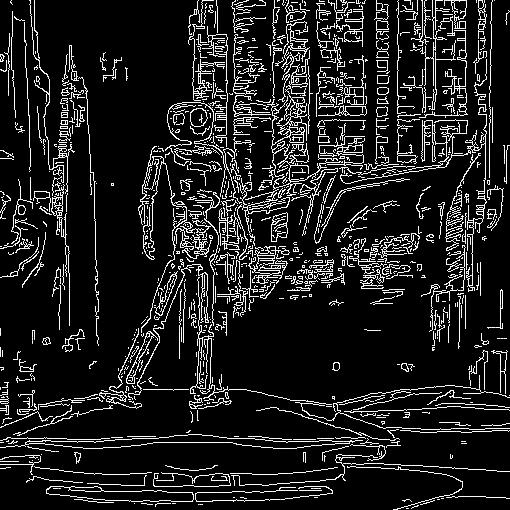}
 \includegraphics[width=0.16\linewidth]{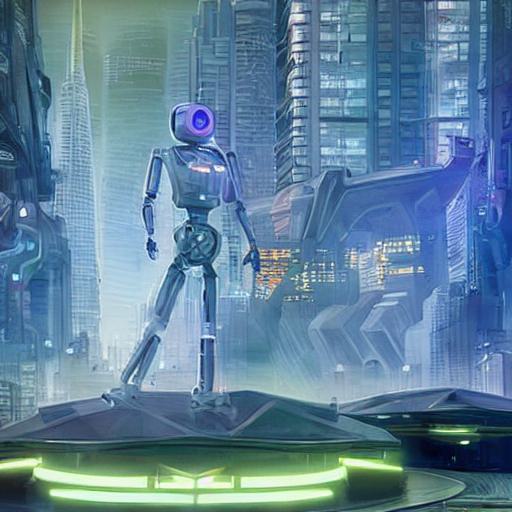}
 \includegraphics[width=0.16\linewidth]{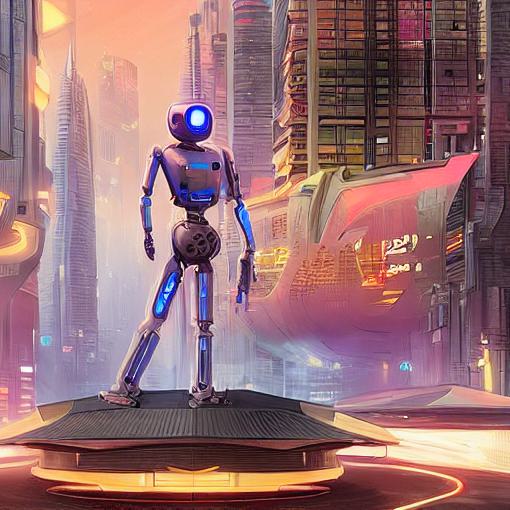} 
    \tabularnewline
 \includegraphics[width=0.16\linewidth]{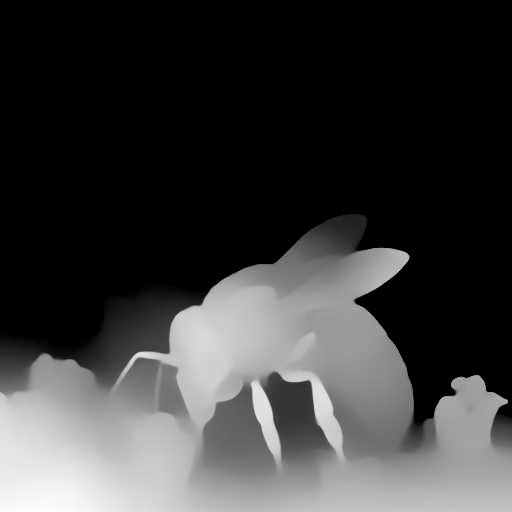}
 \includegraphics[width=0.16\linewidth]{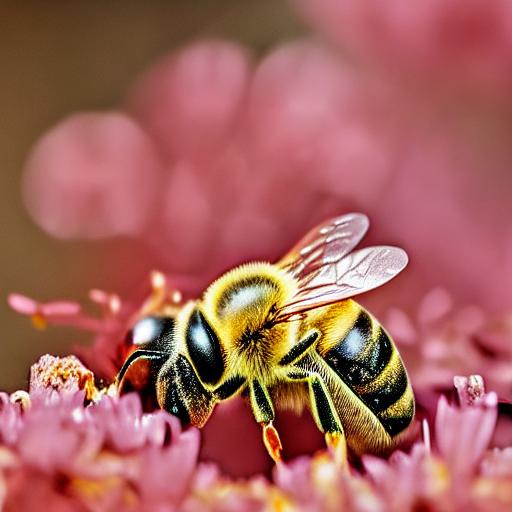}
 \includegraphics[width=0.16\linewidth]{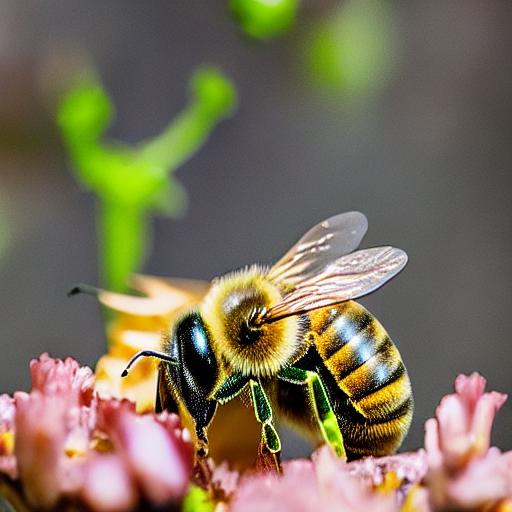} 
\hspace{2mm}

  \includegraphics[width=0.16\linewidth]{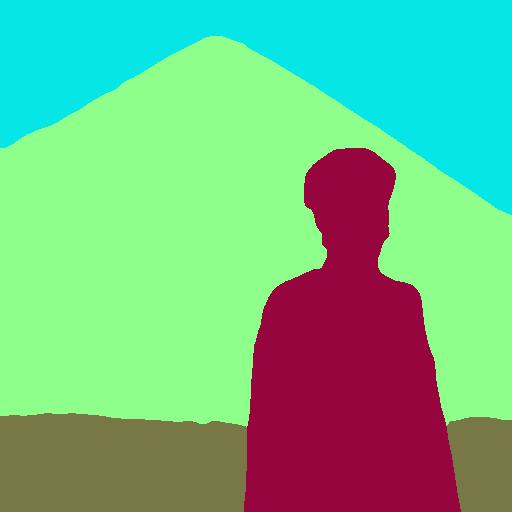}
 \includegraphics[width=0.16\linewidth]{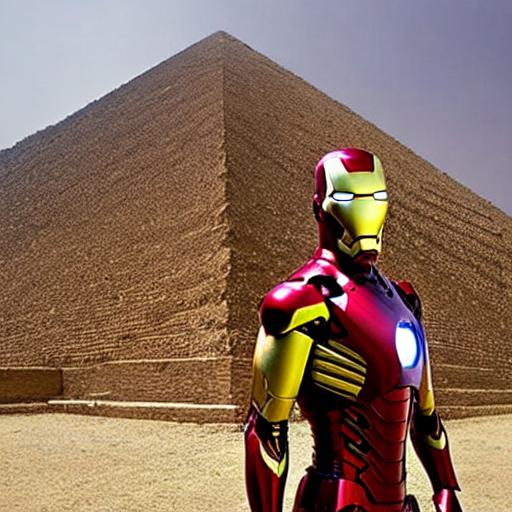}
 \includegraphics[width=0.16\linewidth]{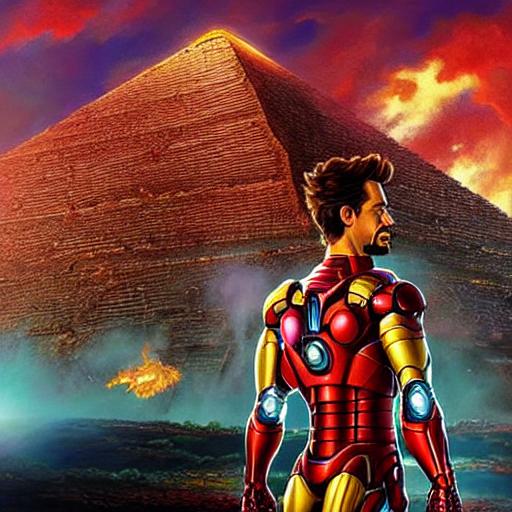}  
 \tabularnewline
%
    \tabularnewline
            \textit{(a) Incremental Dataset addition}\\
 \includegraphics[width=0.094\linewidth]{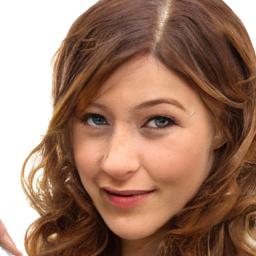}
 \includegraphics[width=0.094\linewidth]{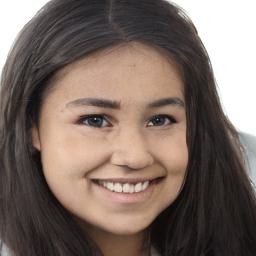}
 \includegraphics[width=0.094\linewidth]{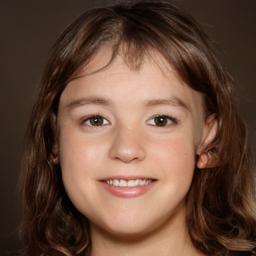}
 \includegraphics[width=0.094\linewidth]{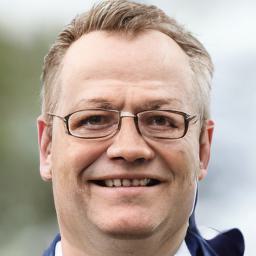}
 \includegraphics[width=0.094\linewidth]{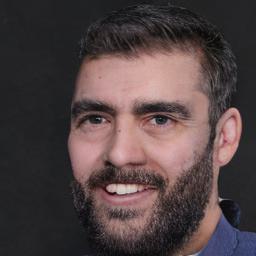}
 \hspace{2mm}
  \includegraphics[width=0.094\linewidth]{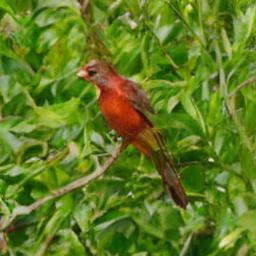}
 \includegraphics[width=0.094\linewidth]{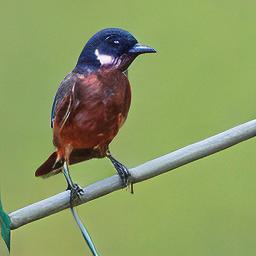}
 \includegraphics[width=0.094\linewidth]{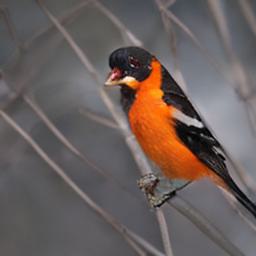}
 \includegraphics[width=0.094\linewidth]{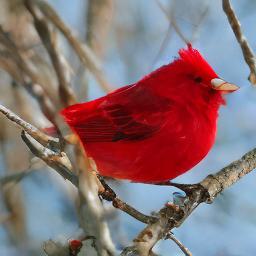}
  \includegraphics[width=0.094\linewidth]{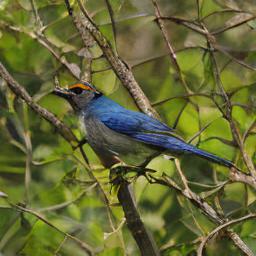}
     \tabularnewline
 \includegraphics[width=0.094\linewidth]{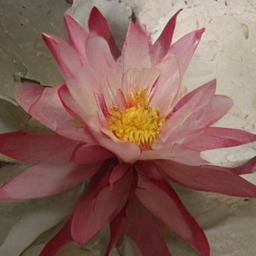}
 \includegraphics[width=0.094\linewidth]{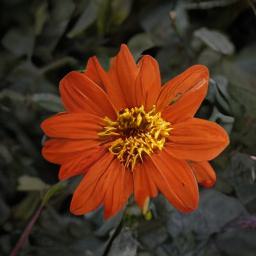}
 \includegraphics[width=0.094\linewidth]{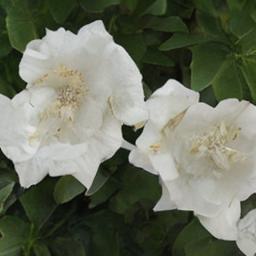}
 \includegraphics[width=0.094\linewidth]{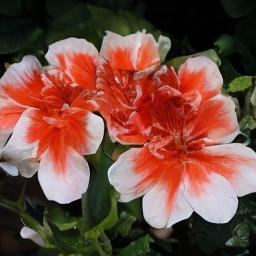}
  \includegraphics[width=0.094\linewidth]{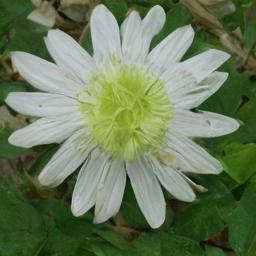}
 \hspace{2mm}
 \includegraphics[width=0.094\linewidth]{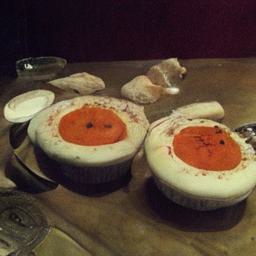}
 \includegraphics[width=0.094\linewidth]{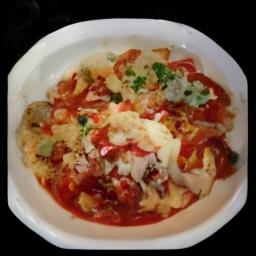}
 \includegraphics[width=0.094\linewidth]{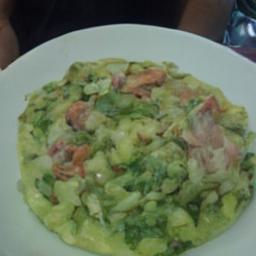}
 \includegraphics[width=0.094\linewidth]{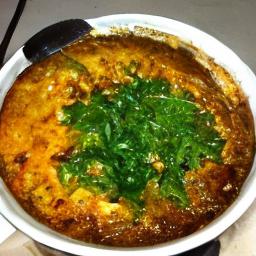}
 \includegraphics[width=0.094\linewidth]{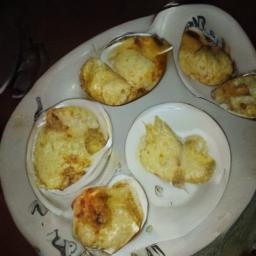}
\tabularnewline
\end{tabular}}
\vspace{-0.5\baselineskip}
\hspace{20pt}\captionof{figure}{\textbf{Applications of our method:} \textbf{Top row:} Samples selected while using DiffScaler to enable a \textcolor{blue}{\textit{single}} text conditioned diffusion model to perform multiple spatial conditioning tasks simultaneously. \textbf{Bottom row:} Images generated while using DiffScaler to enable a  \textcolor{blue}{\textit{single}} diffusion model for image generation across diverse datasets}
\label{fig:introfig}
\vspace{-2mm}
\end{center}%

 \noindent additional condition \cite{saharia2022photorealistic,nichol2021glide}. For example, text-to-image generation models such as DALL-E 2 \cite{ramesh2022hierarchical} and ImageGen \cite{saharia2022photorealistic} have shown impressive performance. Further works have explored finetuning these models for task-specific applications like sketch to image \cite{voynov2022sketch}, depth to image \cite{zhang2023adding, kim2022dag}, segmentation maps to image \cite{li2023guiding} etc. At present, there is no straightforward method to utilize a single diffusion model for generating high-quality images of multiple datasets/tasks  simultaneously. Although different models can be used,  it would be more desirable to have a single model that can generate high-quality images for any dataset/task as required. Hence, we focus on addressing this challenge by aiming to answer the question: \textit{"How can we enable a single diffusion model to generate high-quality images for multiple diverse tasks or datasets?"}

The emergence of foundation models has sparked considerable interest in research focused on efficient transfer of knowledge from these large models to specific downstream tasks \cite{hu2021lora,zaken2021bitfit,lester2021power,su2022transferability,jia2022visual}. These strategies enable the utilization of language understanding capabilities found in the large language models (LLMs) (Ex: GPT-3 \cite{brown2020language}) and vision capabilities in the large vision models (Ex: CLIP \cite{radford2021learning}) for fine-tuning purposes in specific domains. For the case of diffusion models, a straightforward technique would be to finetune an existing pre-trained model with new data. For example, to enable a stable diffusion model \cite{rombach2022high} for the task of mask conditioned text to image generation, one can train it on  masks paired along with the original $\{text, image\}$ data. Performing this at a small scale for datasets is difficult because re-training a large model is computationally expensive while also leads to catastrophic forgetting. Efficient transfer learning techniques for LLMs like  LORA \cite{hu2021lora} and Bitfit \cite{ben2021bitfit} have not been explored for diffusion models. We conducted experiments using these techniques and observed that they obtain a decent performance although not being able to enable the diffusion model to scale to multiple tasks. Due to their inherent design, these methods either do not utilize the pre-trained spaces well or not provide a flexibility of learning new subspaces. Also, ControlNet \cite{zhang2023adding}  proposed a strategy where a task-specific encoder is added to a pretrained stable diffusion model and connected it with zero convolutional layers. Zero convolutional layers were originally formulated in \cite{zhang2023adding} where they enable a stable starting point for fine-tuning a pre-trained network. ControlNet thus offers the capability to condition a large model on relatively small datasets. However, ControlNet still cannot scale a single model to multiple datasets as it utilizes a separate encoder for each task which increases the model size and makes it difficult to host the full model while dealing with multiple tasks. 

In this work, we propose a new learning strategy called \textit{DiffScaler} where we add and optimize minimal parameters in the network to learn a new task. Please note that in contrast to existing methods,  we keep the pre-trained weights frozen and learn only these additional parameters for adapting to a new task.  Diffscaler enables scaling the model to multiple tasks by conditioning the model in parallel 
 or one after the other. For example, one can opt to simultaneously train independent sets of parameters for all new tasks in parallel. This would save training time but would need the availability and knowledge of all the tasks. Alternatively, one can opt to train independent sets of parameters for new tasks sequentially, allowing more flexibility. For example, initial training can involve conditioning stable diffusion on depth maps, followed by training on conditioning with segmentation masks, sketch, etc.  Either way, we scale the model while utilizing less than $\bf{1\%}$ of its parameters for each new task ultimately achieving a computationally efficient model capable of multitasking. 
To enable this compute-efficient transfer through DiffScaler, we introduce a  light weight block \textit{Affiner} to each trainable layer of the diffusion model and train only these parameters to adapt to a new task. Specifically, we introduce a scaling parameter for each weight in the network and a new bias to every learnable bias layer. 
We also add a block that learns new dataset-specific features which might be absent in the pre-trained model. We apply DiffScaler on  existing  transformer-based diffusion models \cite{peebles2022scalable} and CNN-based diffusion models \cite{nichol2021improved,dhariwal2021diffusion}, text-to-image models \cite{rombach2022high} and achieve remarkable generation quality for diverse tasks across different datasets. We show the usefulness of DiffScaler for both conditional image generation across multiple tasks and also unconditional image generation across multiple datasets (with some examples in Figure \ref{fig:introfig}). To further reduce the computational overhead for transformer-based diffusion models, we utilize masking for memory-efficient optimization of the new parameters. To summarize, our main contributions are as follows:
\begin{itemize}
\item We present DiffScaler, an approach which allows generative models can be scaled across diverse datasets efficiently. 
\item We show that transformer backbone diffusion models adapt better to smaller datasets while performing parameter efficient finetuning.
\item We propose a lightweight module, Affiner, which performs learnable scaling of weights and biases of latent subspaces while also learning additional task-specific lower-dimensional subspaces to scale effectively.
\item We illustrate the working of our method for both conditional and unconditional generative models across multiple tasks and datasets while also showing its generalizability on both convolution and transformer-based diffusion models.

\end{itemize}

\section{Related Works}

\subsection{Scaling and Controlling Diffusion Models}
 DDPMs \cite{ho2020denoising} have obtained impressive performance for various generation tasks like text-conditioned image generation \cite{rombach2022high,nichol2021glide,saharia2022photorealistic}, video generation \cite{ho2022imagen,ho2022video}, and multimodal generation \cite{rombach2022high,ruiz2022dreambooth}. 
  ControlNet \cite{zhang2023adding} was proposed to control these pretrained large
diffusion models to support additional input conditions. For this, paired $\{image, 
\newline
text, conditoning\}$ triplets are synthetically generated. To add an additional task, an additional task-specific encoder initialized with the original encoder weights is added to the  network and connected with convolutional layers initialized with zeroes. The purpose of this zero initialization is to ensure consistent intermediate latent outputs at the beginning of the training process, thereby providing a stable initialization for effective training. Only the parameters of the additional encoder are trained for adding the conditional functionality. Please note that naive finetuning-based approaches do not retain the pre-trained information and so lead to catastrophic forgetting \cite{kirkpatrick2017overcoming}. Although ControlNet prevents this from happening, it cannot  be scaled for task-specific conditioning for multiple datasets. Recently DiffFIt\cite{xie2023difffit} introduced a method to fine-tune diffusion models by fine-tuning the bias terms in the diffusion model. Howoever, learning only the basis terms does not let the model adapt to vast diverse datasets. We present the results in the experiments section.
\vspace{-0.1mm}
\subsection{Transfer learning  of Large Pre-trained models }
Transfer learning has been a significant focus of research in the fields of computer vision and large language models \cite{bapna2019simple,ben2021bitfit,chen2022adaptformer,caron2021emerging,he2022masked,pfeiffer2020adapterfusion, jia2022visual,ju2022prompting}. The most prominent transfer learning technique is linear probing, where a part of the pre-trained network is frozen, and only an additional task-specific layer is trained for the task at hand. Another approach to transfer training is to learn the whole method from scratch by fine-tuning all layers in the network. 
Recently, several transfer learning approaches have been proposed for language models like BitFit \cite{ben2021bitfit} where the bias of the query layer in transformer and the bias of the second MLP is tuned. LORA \cite{hu2021lora} proposes a low-rank adaptation technique to add an additional weight to the query and value matrices of the network. Adaptor \cite{houlsby2019parameter} proposes adding a parallel branch to the MLP layer of the network to improve the performance. For large vision models, adaptformer \cite{chen2022adaptformer} uses additional lightweight parallel MLP layer to the transformer network to enable downstream tasks. Visual prompt tuning \cite{jia2022visual,ju2022prompting} approaches add learnable tokens in the input space to shift the input  to accommodate the large model.

\section{Method}

In this section, we explain our proposed method, DiffScaler. Regardless of the model type like GANs, VAEs or Diffusion models,  through a close look at the inference process of diffusion models one can see that the learned context is different along the sampling process. In particular, the initial steps of the inference process tune the broad context of the dataset which then gradually progresses to become a denoising task as the time steps progress. Recent works \cite{balaji2022ediffi} have utilized this characteristic to ensemble multiple models for learning time step specific context. 
The denoising capability of a network for different noise levels is attributed mainly to the weight layers  as shown in previous works \cite{mohan2021adaptive,ho2020denoising}. Also, the dataset-specific information  in generative models is attributed to the bias in the network \cite{ben2021bitfit}. Therefore, to enable a model to learn contextual information and generate photorealistic outputs, additional weights and biases are required to facilitate dataset-specific context generation.

\subsection{Unlocking incremental dataset addition}

 Given a diffusion model that can generate N classes, and we want to expand the capability of this model to extend to an additional dataset that contain M classes. At the same time, we also need to ensure that after the addition of the new classes, the model's capability to generate the old N classes should not deteriorate. 
 We address this problem by freezing the original model parameters for the N classes and introducing a small number of new parameters for learning the new M classes. Moreover, to map the model to the new dataset, we need to add new embedding layers that can map to the new classes.  Consider the embedding matrix of a class conditioned diffusion model with N classes and the embedding matrix $W_{emb} \in R^{(N+1)\times d}$ where $d$ is the embedding dimension of the matrix. The extra one class denotes to the unconditional class that models the distribution of all the other classes. We denote this by $W_{uc}$. The embedding matrix $W_{emb}$ can be expressed as follows:
 \begin{equation}
     W_{emb} = [W_1,W_2,...,W_N,W_{uc}, W_{N+2},...,W_{N+M+1}]
 \end{equation}
We introduce a new training scheme to incrementally add these new classes as well as keep the training process stable as we want the network to work well and denoise when the training process starts. To accommodate this setting, we initialize $W_{N+2},..,W_{N+M+1}$ as $W_{uc}$. This setting ensures that when the training starts, the model is able to satisfy the denoising objective, even though it is not able to generate the new classes. The same principle underlies in ControlNet \cite{zhang2023adding} where the initial feature embeddings that are added are zero at the start of training, hence ensuring a stable prediction that satisfies the denoising objective at the start of the training process. Changing the weights of the embedding matrix alone can only give a combination of all the previously seen classes for the network.
We validate this claim in the experiments section, where we see that a model  pre-trained on ImageNet classes is unable to generate facial images by training the embedding matrices alone. To enable the network to learn new classes, the basis functions of these new classes need to be learned by the network and in cases where the new classes vary drastically from the original pre-trained model classes, new parameters need to be added to accommodate this generation capability. In the next section, we describe the new weight addition mechanism we propose to enable scaling of existing models to new datasets.

\subsection{Affiner} Let us consider any trainable layer in a neural network. The output $y$ for an input $x \in R^n$ can be modelled as 
\begin{equation}
    y= Wx+\hat{b},
\end{equation}

where $W \in R^{m\times n}$ and $b \in R^m$ are the trainable weights and bias in that layer, respectively. In most cases, the memory consumption during training occurs while storing the weight matrix $W \in R^{m\times n}$  and its corresponding gradients.  Given the formulation, LORA serves as an ideal candidate to increment the generative prowess of the network, as it can add new basis vectors to the weight matrices. But the rank of $W$   The LORA modules can be thought of as learning additional basis functions  to learn the lower dimensional fine-tuning subspace. In the case where  $R = min(m,n)$, LORA blocks become redundant. Moreover, LORA doesn't have the capability to utilize the basis functions of $W$ effectively and may have to learn the same column vectors as in $W$ since no scaling of $W$ is present in LORA. Hence, the model still lacks the capability for modelling data which is totally out of distribution from the ones in the training data and as per our experiments in multiple cases this led to explosive gradients. Hence, we add an additional weight re-parameterization to account for these additional basis spaces. 
\begin{figure*}[htbp]
    \centering
    \includegraphics[width=0.8\linewidth]{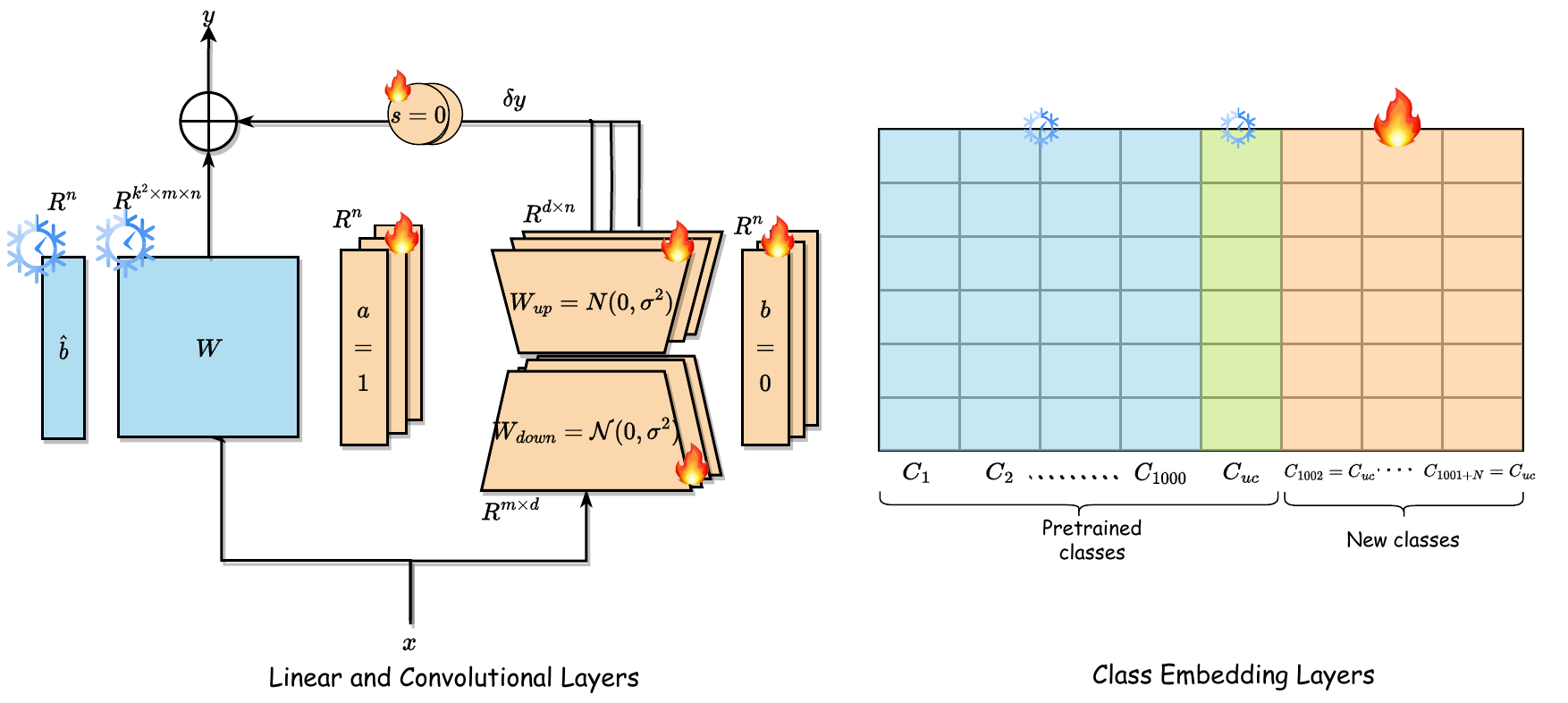}
    \vspace{-4mm}
  \caption{ An illustration of the proposed Affiner block. We scale each weight matrix in the network and shift the trainable bias layer thus enabling the ability to utilize any subspace of the learned matrix. Moreover, to include additional subspaces, we add a parallel low-rank decomposition branch.}
    \label{fig:intro2}
\end{figure*}

 It can be noted that in a pretrained diffusion model with weights $W$, the model inherently can perform denoising. Hence, the task at hand is to learn the contextual information depicting the distribution of the dataset. Motivated by the role of style embedding in StyleGAN \cite{karras2019style} and BitFit \cite{ben2021bitfit}, we add two lightweight trainable parameters, a scaling parameter $a$ and shifting parameter $b$ initialized to $a$ and $b$, respectively. Specifically, to enable infusing of new styles into the pre-trained network, we re-parameterize the network layer to 
\begin{equation}
    y= (1+a)Wx+\hat{b}+b,
\end{equation}
where $a$ and $b$ are learnable parameters which scale and shift the weights, respectively. 

In many cases of diffusion models, one key aspect to consider is its distribution-dependent denoising capability. 
Although the above formulation helps to parameterize the weights of the network efficiently, 
one key point to notice here is that if the weight matrix $W \in R^{m \times n} $ has a rank $R$ such that $R < min\{m,n\}$, then the proposed scaling and shifting of the bias cannot model basis vectors as that of a full rank weight matrix $R=min\{m,n\}$. Hence, we try to provide the network with the capability to model these additional basis functions for the weight matrix. We perform low-rank decomposition of the original weight matrix denoted as follows:
\begin{equation}
    \delta y=s(W_{up}ReLU(W_{down}.x)),
\end{equation}
where $W_{up} \in R^{m \times d}$ and  $W_{down} \in R^{d \times n}$ , $d \ll \{m,n\}$, $s$ is a trainable scaling parameter initialized with zero. We initialize $W_{down},W_{up}$ with Kaiming normal initialization \cite{he2015delving} to satisfy the trivial condition. Our initialization strategy differs from approaches in \cite{hu2021lora,chen2022adaptformer} where the weights are initialized with zeroes.

 Including the new weight formulation to our previous learnable scaling and shifting parameters, the overall parameterization of each trainable layer becomes
\begin{equation}
    y=(1+a)Wx+sW_{up}ReLU(W_{down}.x)+\hat{b}+b.
\end{equation}

This becomes the plug-in layer we introduce called Affiner. To summarize,  $a\in R^n$ is initialized as ones, $b\in R^n$ is initialized with zeros, and $W_{down}$ and $W_{up}$ are initialized with Kaiming normal, and $s$ is initialized with zero. Note that Affiner can be plugged in to both convolution and transformer layers.
In the case of 2-D convolutional layer, we model $\{W_{down},W_{up}\} $ with  $1\times 1$ convolutions layers and additional bias and scaling 
 are just learnable parameters. The same holds true for the case of 1D and 3D convolutions. In the case of vision transformers \cite{vaswani2017attention, dosovitskiy2020image}, the comprising transformer blocks include multi-head attention and a feed forward block with MLPs and layer normalization. In both self-attention and MLP layers, the weight transformation occurs through linear layers. Hence, our formulation still holds. For transformer models, we add our Affiner block for each key, query, value weights and bias parameters as well as the MLP block. Affiner has been illustrated in Figure \ref{fig:intro2}.

\subsection{Scaling up affiner to multiple datasets}



Note that DiffScaler can be used to scale a model across multiple datasets or tasks. We utilize the term datasets while dealing with unconditional generation in specific datasets and utilize the term task while dealing with conditional generative tasks. For simplicity, in this section we explain scaling in terms of multiple datasets but note that the same strategy works for scaling across multiple tasks as well. To scale a diffusion model to $N$ different datasets, we first create a new parameter set $\{W_{up}^i,W_{down}^i,b^i,a^i,s^i\}$  for each $i$ of the $N$ datasets. We train these new parameters for each task individually. Then, these parameters are stacked together to perform generation tasks using a single diffusion model. One can also simply train the model on all the datasets in parallel to save time. In this case, there will be $N$ sets of these new parameters which are trained together across all datasets simultaneously. Alternatively, to add a new dataset to a diffusion model that has already been trained on $N$ datasets, one can simply add a new set of parameters and train that model for the new task. Further once a model is trained in this proposed manner, we propose a new weight merging scheme to fuse these weights for zero shot task generalization of the composition of any of these tasks on the next section


\subsection{Conditional generation without zero convolutions}
Recently ControlNet model added the capability of finetuning stable diffusion with a desired conditioning by first obtaining \{\textit{Image, Caption, Condition}\} pairs and finetuning a parallel encoder of the stable diffusion according to the image-caption pairs. The parallel encoder is connected to the base networks with convolutional layers initialized with zeroes which account for a large number of parameters. More often than not, the conditions are independent and align  with the text caption. As an upgrade to ControlNet, DiffScaler can be used for conditional generation across multiple tasks by using the proposed re-parametrization technique and modifying each trainable layer and train the newly added parameters for the new task at hand. With our formulation there is no need for a separate encoder or zero convolutions like in ControlNet. and we obtain comparable performance  $7M$ trainable parameters whereas that of ControlNet is about $300M$.

\section{Experiments}
In this section, we conduct extensive experiments to show the usefulness of our proposed method. We conduct experiments on both conditional and unconditional generation tasks. We start by taking a pre-trained diffusion model trained on a large dataset and enable it to synthesize images across diverse set of new tasks or datasets. We add a new set of parameters for each dataset and train them alone while retaining the weights, biases and normalization of the original network during the training process.

\subsection{Datatsets}
For unconditional image generation, we conduct experiments on diverse datasets for Faces (FFHQ)\cite{karras2019style}, Flowers (Oxford-flowers) \cite{jia2022visual}, birds (CUB-200) \cite{wah2011caltech}, and Caltech-101 \cite{fei2006one} datasets. 

For conditional image generation, we follow similar protocols like in  ControlNet to generate synthetic datasets. We use COCO dataset \cite{lin2014microsoft} and obtain the corresponding annotated data for the conditions using different pre-trained networks. We pick  canny edge maps, hed maps, depth maps and segmentation maps as the conditional inputs for our experiments. We use BLIP \cite{li2022blip} to obtain corresponding text captions. To obtain hed maps, we utilize the method from Xie et al. \cite{xie2015holistically}. We obtain the depth maps by following the process in \cite{ranftl2020towards} and obtain the segmentation maps by using Uniformer  \cite{li2022uniformer}. We obtain the  edge maps by using a canny edge detector \cite{rong2014improved}.

\subsection{Implementation Details}

 For unconditional image generation, we use a batch size of 16 and train it for 50000 iterations on one NVIDIA A5000 GPU. For conditional image generation, we use a batch size of 4 and train the model for 10000 iterations on a single A6000 GPU. For transformer-based experiments, the masking ratio of experiments to 0.5 unless stated otherwise. Note that for our main experiments we scale the model in parallel with different tasks or datasets and utilize a single diffusion model during inference, one for unconditional and one for conditional. Please look at the supplementary material for experiments on sequential scaling of models. The learning rate for all our experiments was set to $1e^{-3}$.

\subsection{Training Details}

\begin{figure}[t]
\centering
\begin{subfigure}{0.46\textwidth}
  \centering
  \includegraphics[width=\linewidth]{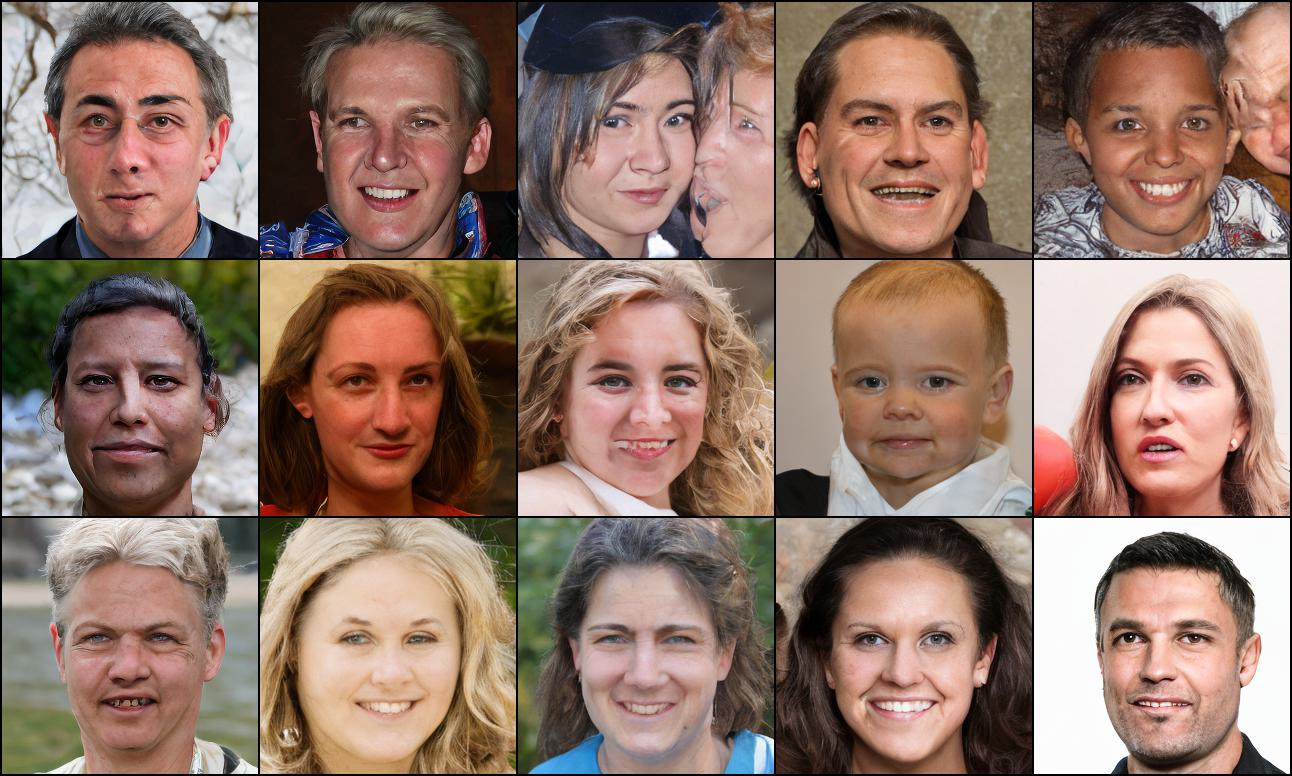}
\end{subfigure}
\begin{subfigure}{0.46\textwidth}
  \centering
  \includegraphics[width=\linewidth]{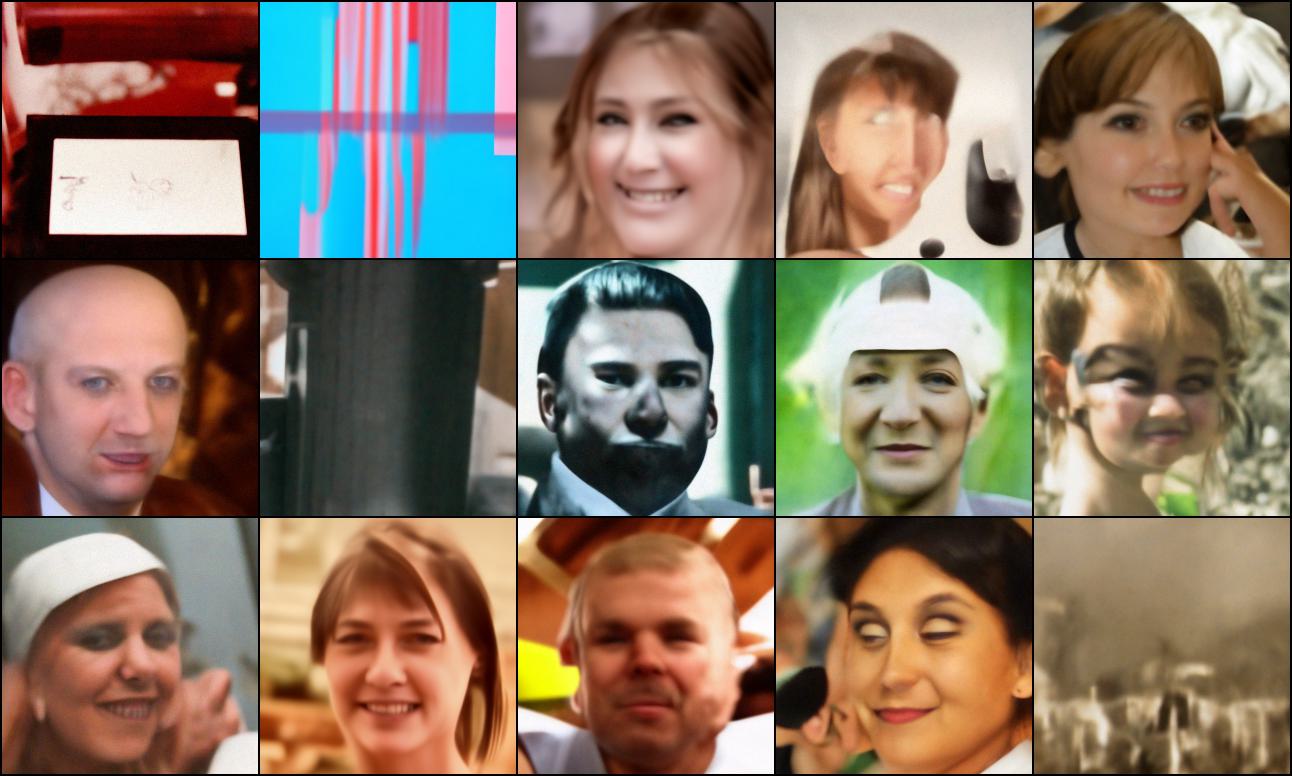}
\end{subfigure}\\
\vspace{-3mm}
\begin{subfigure}{0.46\textwidth}
  \centering
  \includegraphics[width=\linewidth]{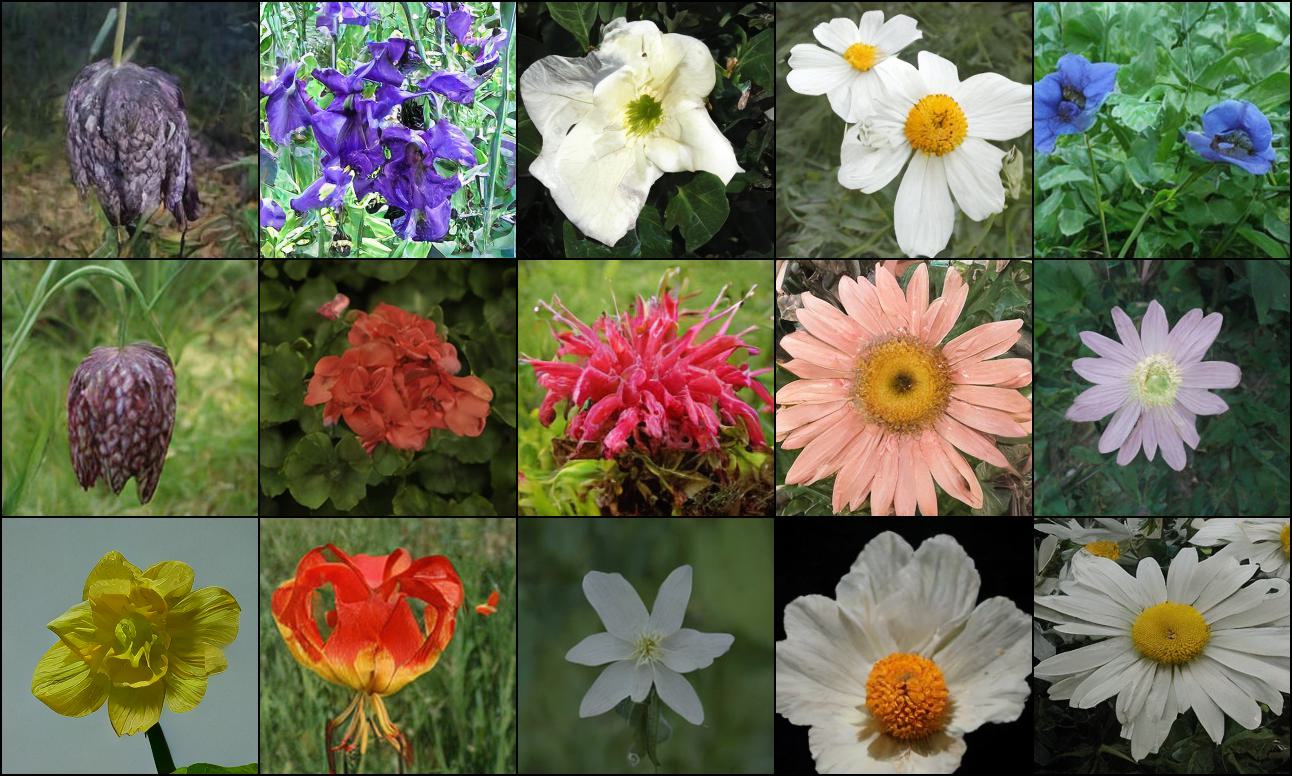}
  \caption{DiT backbone}
\end{subfigure}
\begin{subfigure}{0.46\textwidth}
  \centering
  \includegraphics[width=\linewidth]{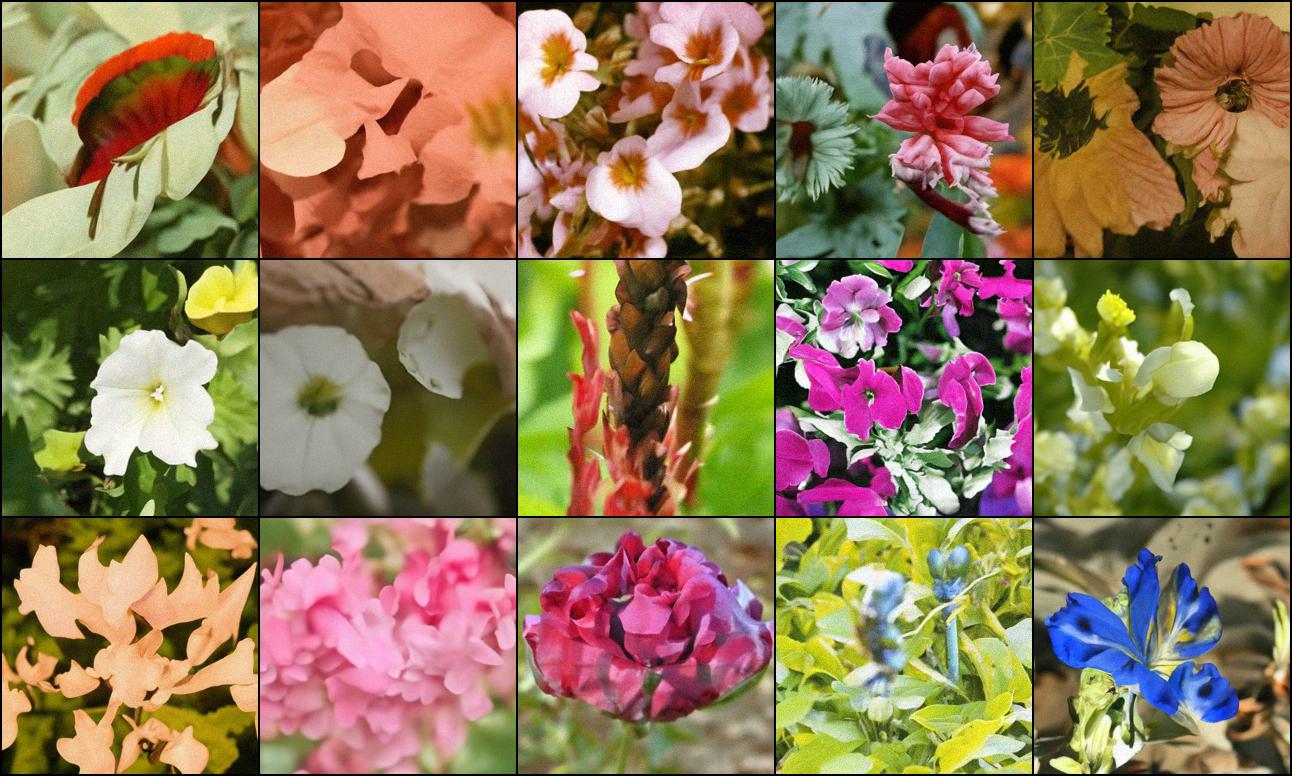}
  \caption{CNN backbone}
\end{subfigure}
\vspace{-1em}
\caption{Parameter efficient fine-tuning on Transformer and CNN backbone. Both the models are pre-trained on ImageNet Dataset  fine-tuned on FFHQ and Flowers  datasets respectively. }
\label{fig:uncond2}
\vspace{-1em}
\end{figure}

\noindent\textbf{Sampling Technique:} For all unconditional experiments, DDIM sampling \cite{song2020denoising} was used, and 100 sampling steps were performed.  For conditional cases, 50 steps of sampling and DDIM sampling were used.

\noindent\textbf{Resolution:} The reported FID scores correspond to FID obtained using 5000 samples. All unconditional samples were synthesized at a resolution of $256\times 256$. Stable Diffusion v1.5 was used as the backbone and the images were synthesized at a resolution of $512 \times 512$. For metric computation, 5000 samples were synthesised  according to condtioning generated from COCO-Stuff dataset. 

\noindent\textbf{Intermediate dimension for DiT and CNN backbone.} For DiT, the intermediate backbone was chosen as 64 as the hidden dimension of DiT is 1152. For Stable Diffusion and other CNN backbones, the intermediate dimension was chosen as 32 since, the number of CNN channels is often less than or equal to 512.

\noindent\textbf{Images for Conditional Generation:} For the examples illustrated in the paper, the conditioning maps were obtained by passing the images generated through Midjourney through the proposed annotation schemes. The images shown in the supplementary material are from COCO-stuff dataset.

\noindent\textbf{Training only the transformer Layers for DiT:} In the case of DiT, most of the trainable parameters are present in the transformer layers, moreover there exist a patch embedding layer which is a convolutional layer and a unpatch layer. Both of these layers have a very low number of parameters and are inherently used for obtaining the embeddings. Since the base model itself can give representative embeddings, we do not retrain this layer. Regarding time step embedding,   these are embedded into the network through adaptive instance normalization layers, hence we train adaptive normalization layers in every transformer block along with the transformer parameters.

\begin{figure}[tp!]
\centering
\begin{subfigure}{0.46\textwidth}
  \centering
  \includegraphics[width=\linewidth]{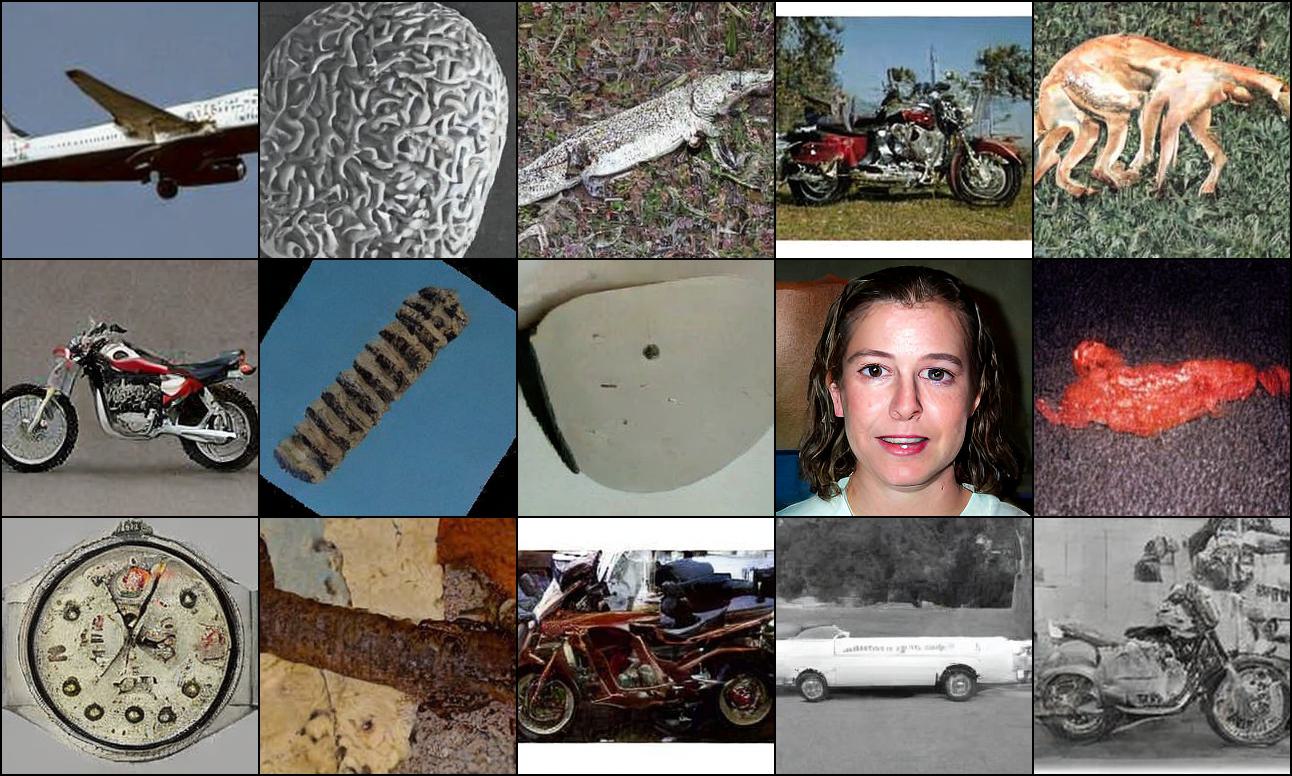}
\end{subfigure}
\begin{subfigure}{0.46\textwidth}
  \centering
  \includegraphics[width=\linewidth]{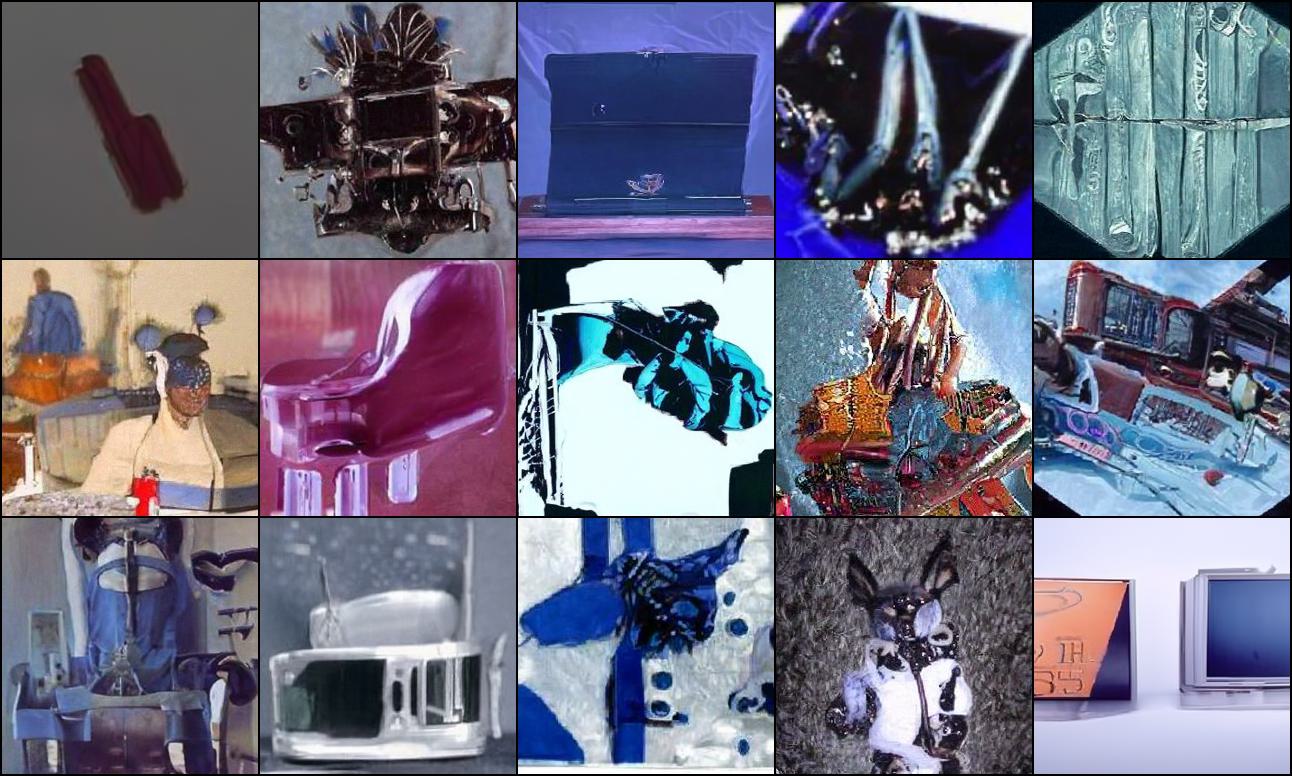}
\end{subfigure}\\
\begin{subfigure}{0.46\textwidth}
  \centering
  \includegraphics[width=\linewidth]{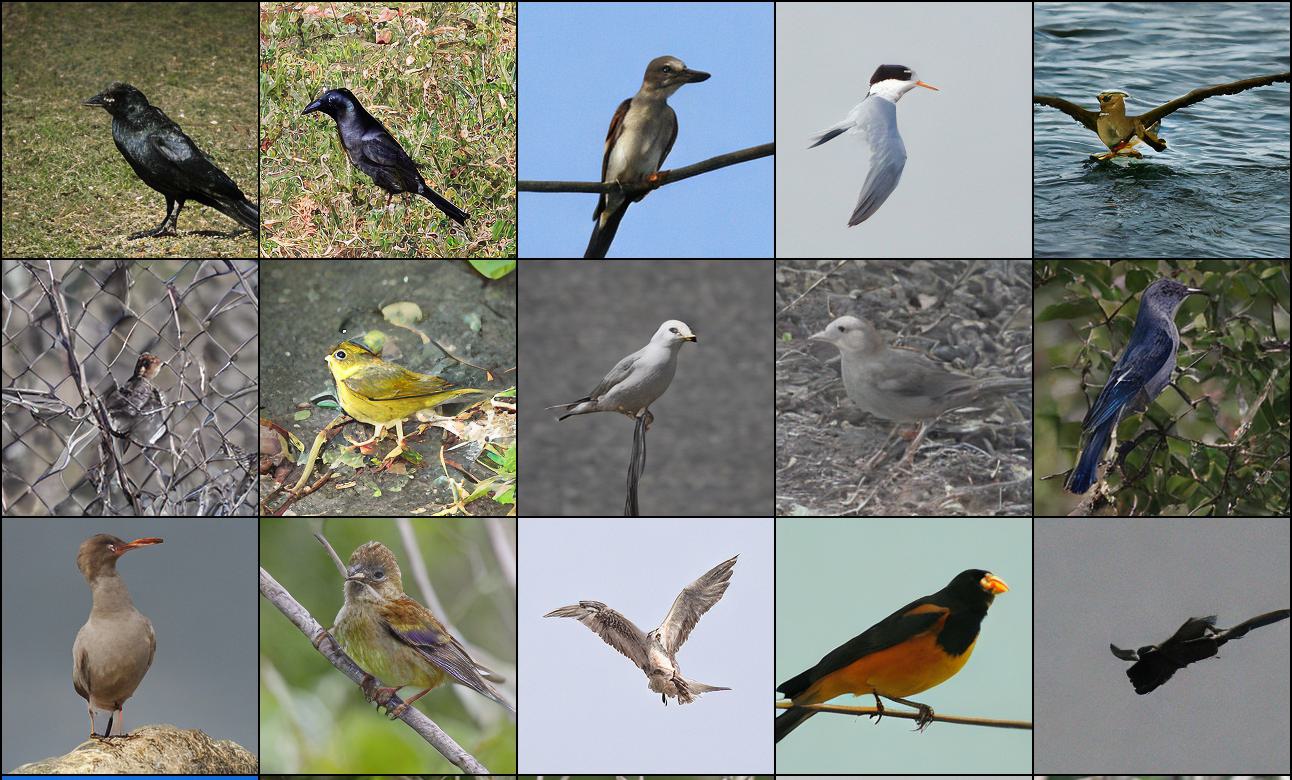}
  \caption{DiT backbone}
\end{subfigure}
\begin{subfigure}{0.46\textwidth}
  \centering
  \includegraphics[width=\linewidth]{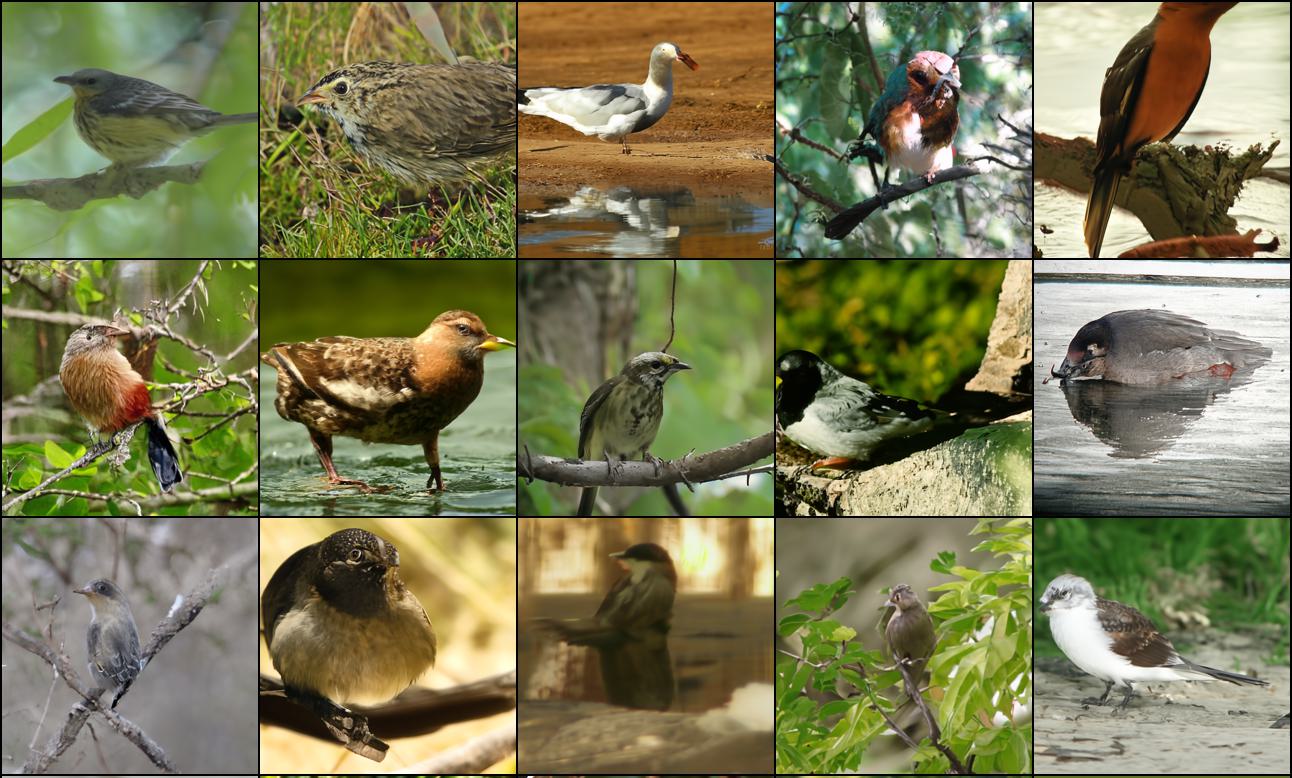}
  \caption{CNN backbone}
\end{subfigure}
\vspace{-1em}
\caption{Parameter efficient fine-tuning on Transformer and CNN backbone. Both the models are pre-trained on ImageNet Dataset and fine-tuned on Caltech, CUB  datasets respectively. }
\label{fig:uncond1}
\vspace{-1em}
\end{figure}

\subsection{Results and Discussions}
\textbf{Unconditional Generation:} It can be noted that, the only work for  efficient finetuning for diffusion-based generative models is DiffFit\cite{xie2023difffit}. For comparison, wWe utilize DiffFit\cite{xie2023difffit}, but train from scratch using the same compute resources for the same number of iterations. We also adopt LORA as our baseline although it was originally proposed for fine-tuning LLMs.  In DiffFit, only the biases of the MLP layer and the second MLP layer of the attention network are trained. In LORA, only the attention weights of the transformer are trained. As seen in Table \ref{table:uncond}, the efficient fine-tuning techniques obtain much higher FID scores compared to full-finetuning as a large number of parameters are present for the optimization process and the model easily overfits on smaller datasets. The FID scores for LORA and Bitfit are low owing to the incapability of DiffFit for high-resolution generation. If the dataset to generate requires higher latent dimension than the latent dimension in LORA, these models perform worse and the FID and inception scores reduce. Qualitative results for unconditional image generation can be seen in Fig \ref{fig:uncond2} and \ref{fig:uncond1}. Our method produces high fidelity photorealistic images across multiple datasets using a single model.

\noindent \textbf{Transformer vs CNN backbone for Parameter efficient tuning:} We present the results of incrementing with a new dataset for Transformer backbone vs CNN backbone in Table \ref{table:uncond} and Figures Fig \ref{fig:uncond2} and \ref{fig:uncond1} respectively. As we can see, the transformer backbone models adapt well to the different datasets efficiently producing realistic samples whereas the CNN backbones fails to produced meaningful samples for Caltech, FFHQ and Flowers dataset. Surprisingly, both models produce good results for the CUB dataset. We argue that this occurs because of the bird classes that is present in the ImageNet dataset. Hence adapting to CUB datasets is easier, whereas for less diverse cases, the CNN-based backbone fails and transformer backbone produces high quality results with better FID scores. The corresponding quantitative metrics can also be seen in Table \ref{table:uncond}, as we can see the quality of samples reflect in the FID scores as well.

\begin{figure}[htp!]
\centering
\begin{subfigure}{0.32\textwidth}
  \centering
  \includegraphics[width=\linewidth]{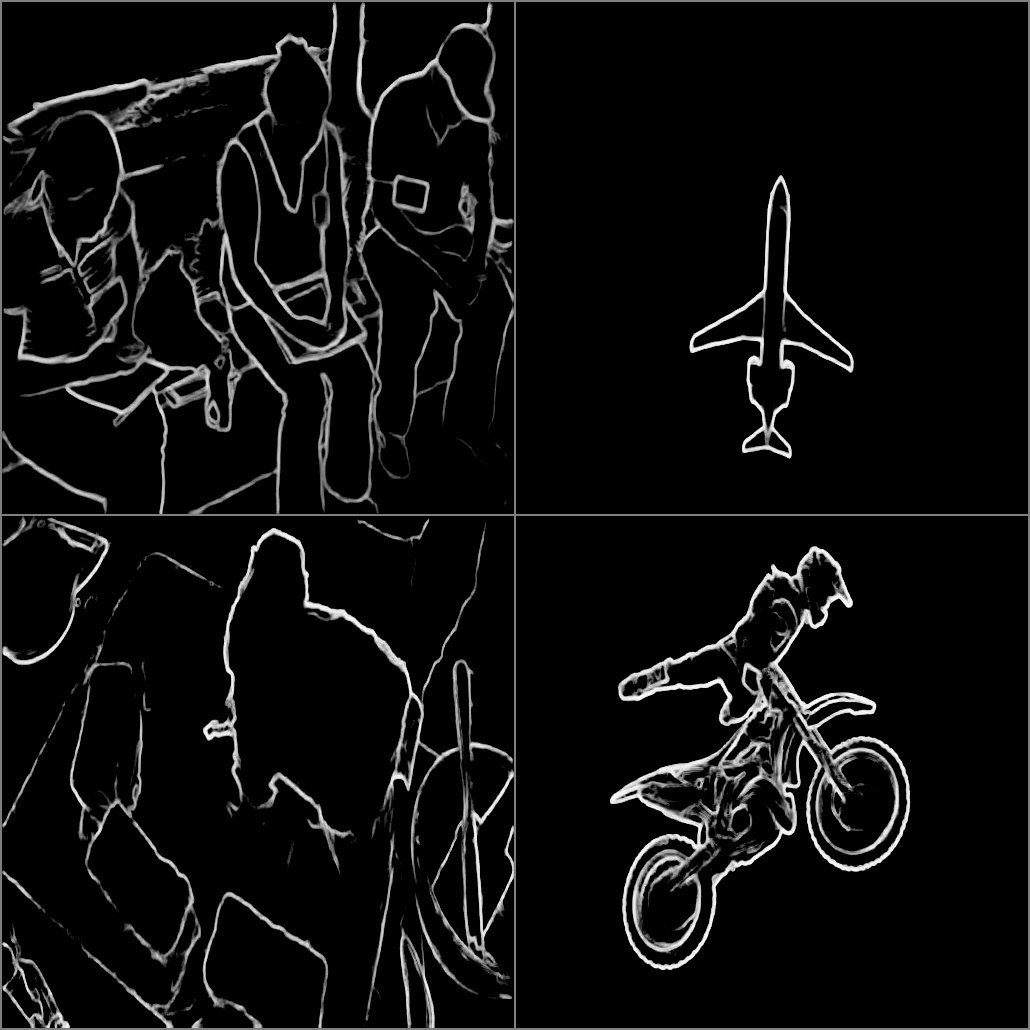}
\end{subfigure}
\begin{subfigure}{0.32\textwidth}
  \centering
  \includegraphics[width=\linewidth]
  {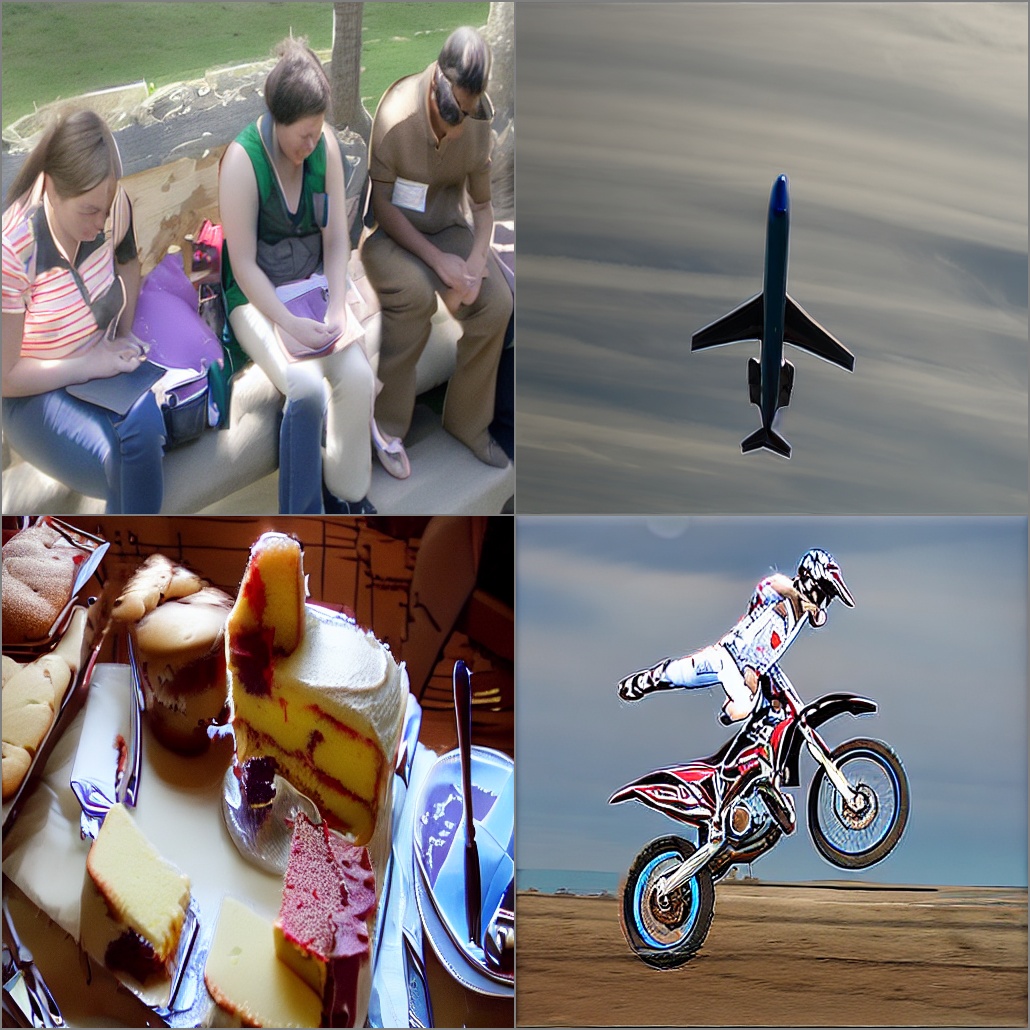}
\end{subfigure}
\begin{subfigure}{0.32\textwidth}
  \centering
  \includegraphics[width=\linewidth]  {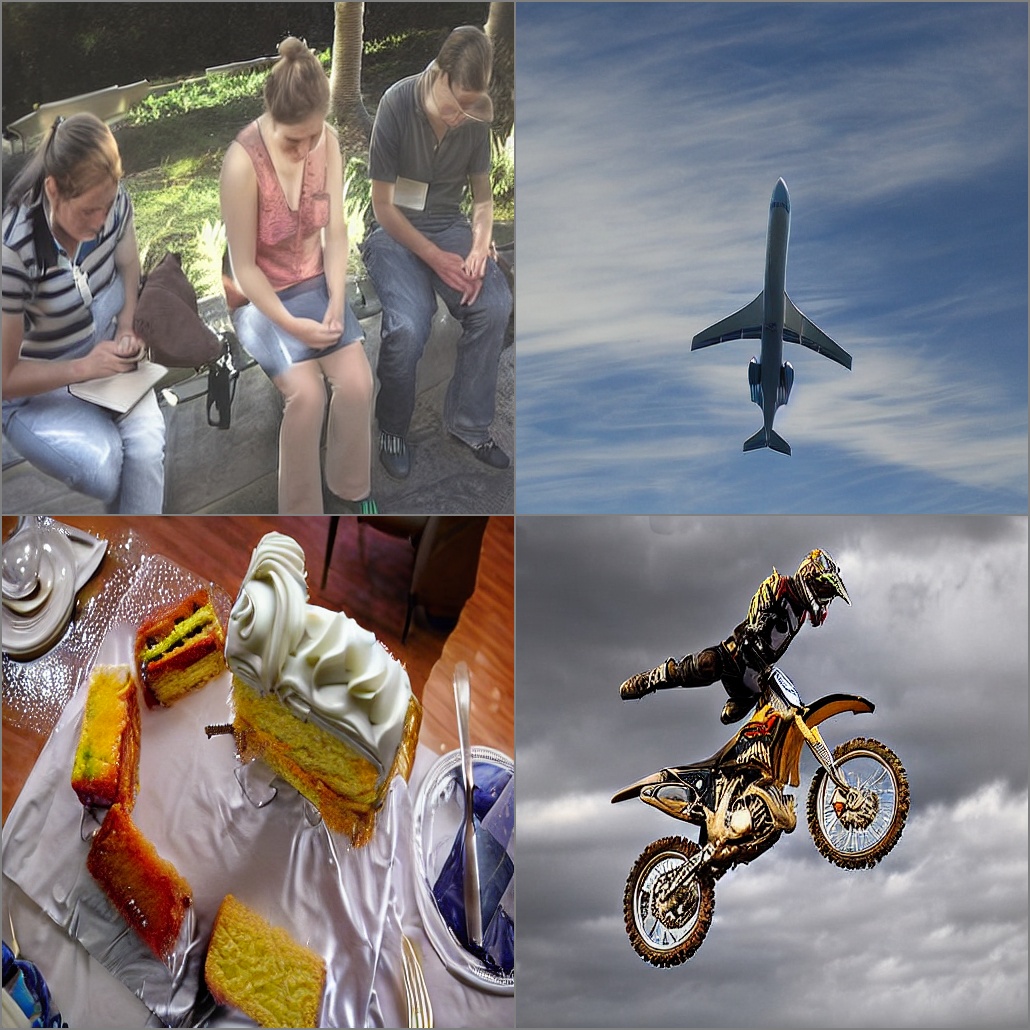}
\end{subfigure}%
\\
\centering
\begin{subfigure}{0.32\textwidth}
  \centering
  \includegraphics[width=\linewidth]  {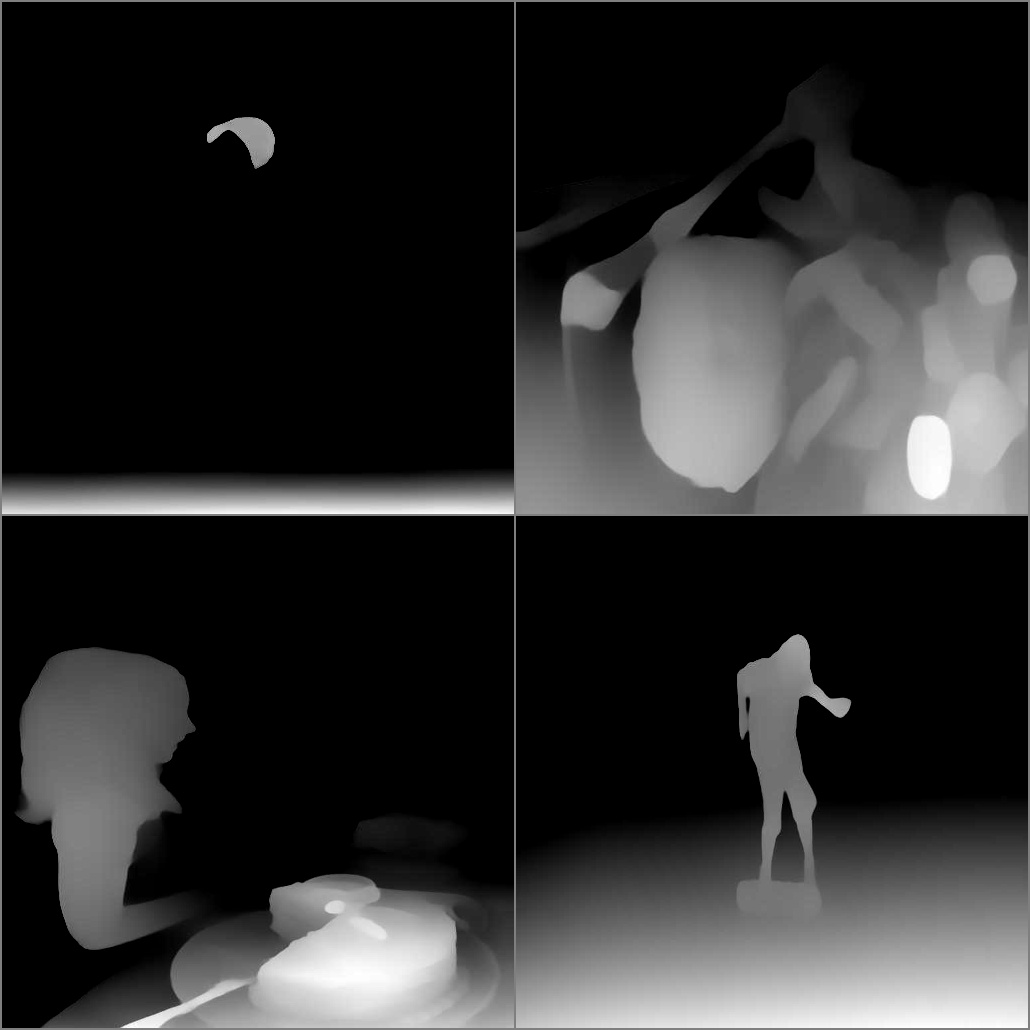}
  \caption{Canny Maps}
\end{subfigure}
\begin{subfigure}{0.32\textwidth}
  \centering
  \includegraphics[width=\linewidth]  {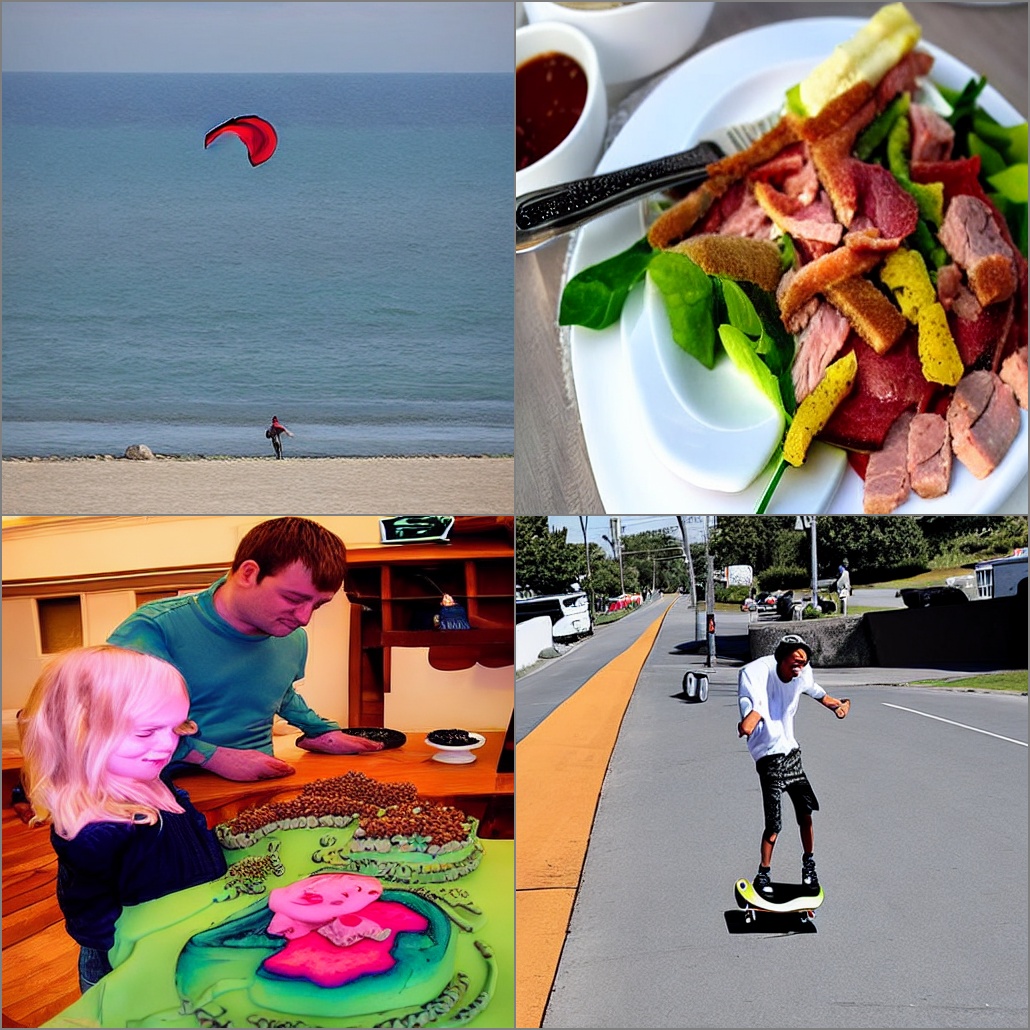}
  \caption{DiffFit\cite{xie2023difffit}}
\end{subfigure}
\begin{subfigure}{0.32\textwidth}
  \centering
  \includegraphics[width=\linewidth]{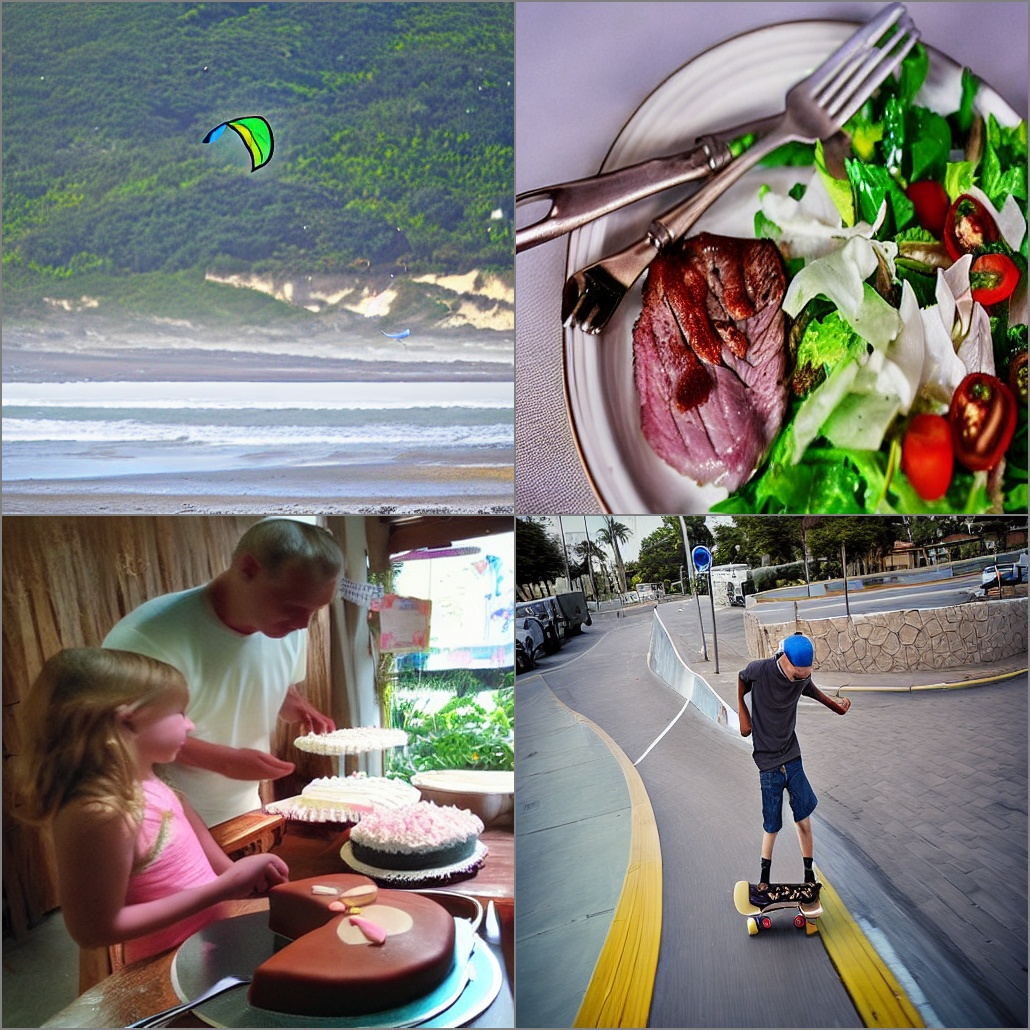}
  \caption{OURS}
\end{subfigure}%
\vspace{-2mm}
\caption{ Qualitative comparisons  for spatial conditioning with HED maps and Depth Maps. }
\label{fig:cond1}
\vspace{-5mm}
\end{figure}

\begin{figure}
\vspace{-0.5cm}
\centering
\begin{subfigure}{0.32\textwidth}
  \centering
  \includegraphics[width=\linewidth]{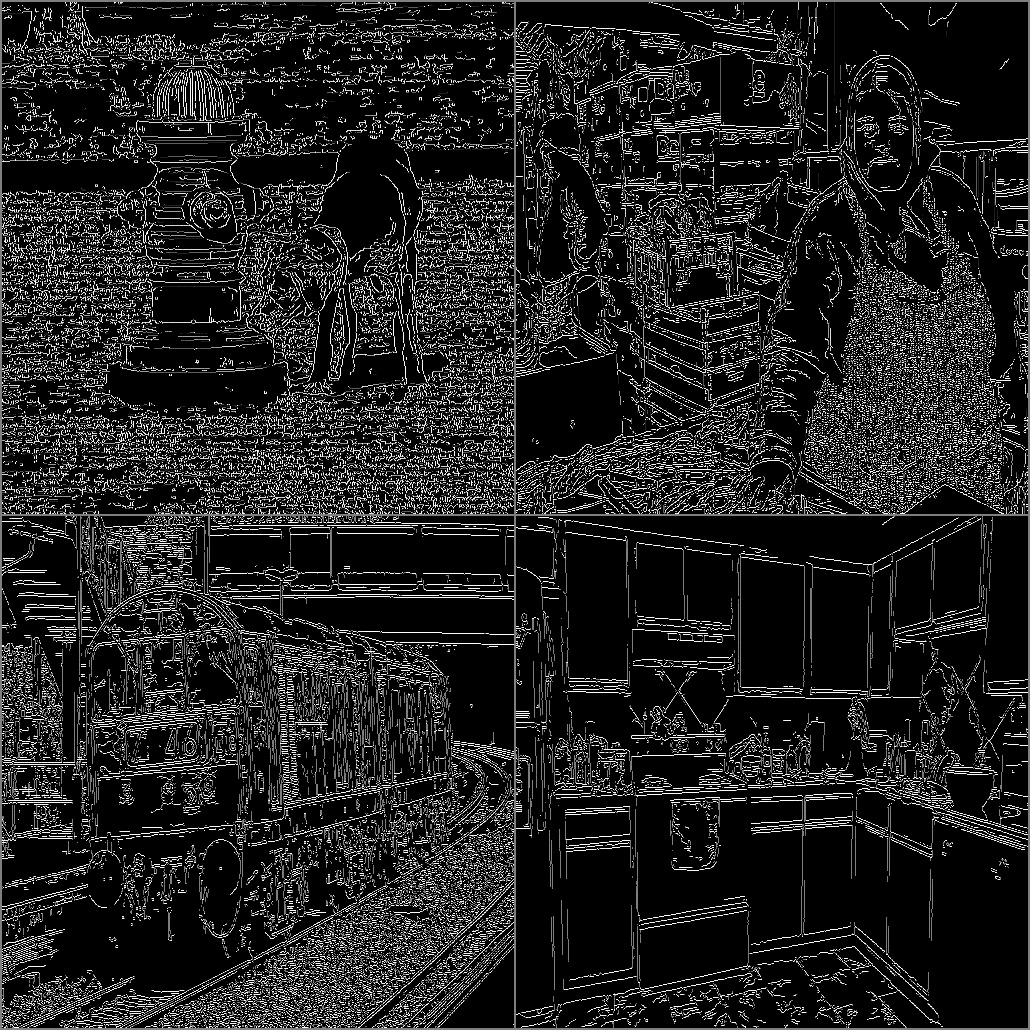}
\end{subfigure}
\begin{subfigure}{0.32\textwidth}
  \centering
  \includegraphics[width=\linewidth]
  {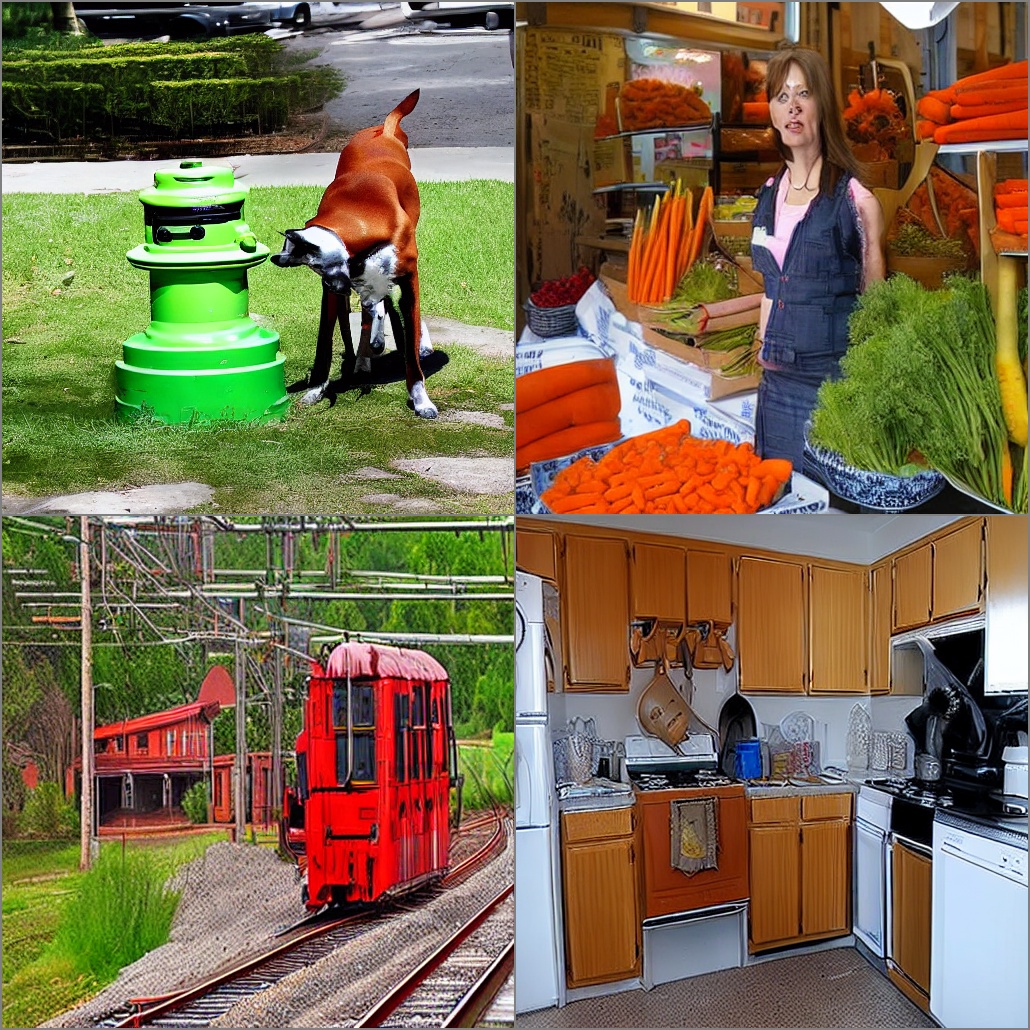}
\end{subfigure}
\begin{subfigure}{0.32\textwidth}
  \centering
  \includegraphics[width=\linewidth]  {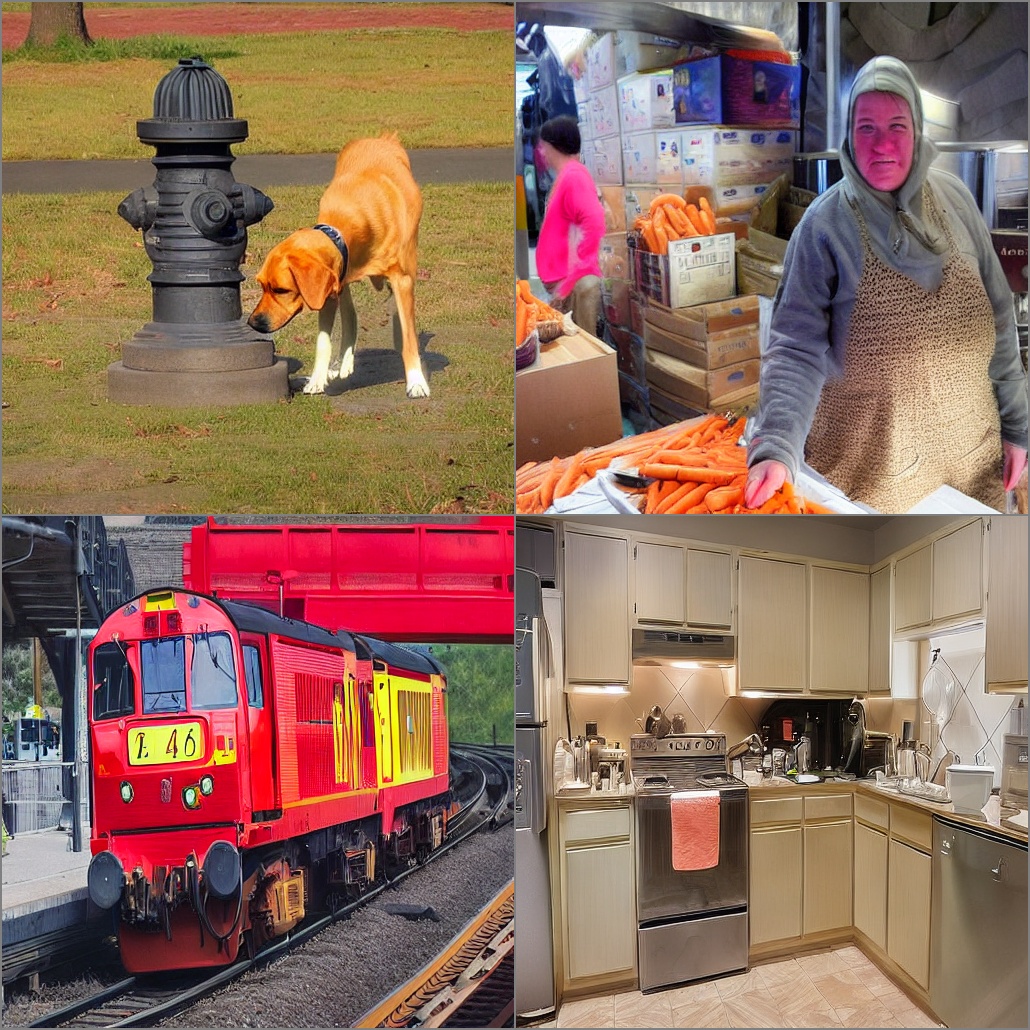}
\end{subfigure}%
\\
\centering
\begin{subfigure}{0.32\textwidth}
  \centering
  \includegraphics[width=\linewidth]  {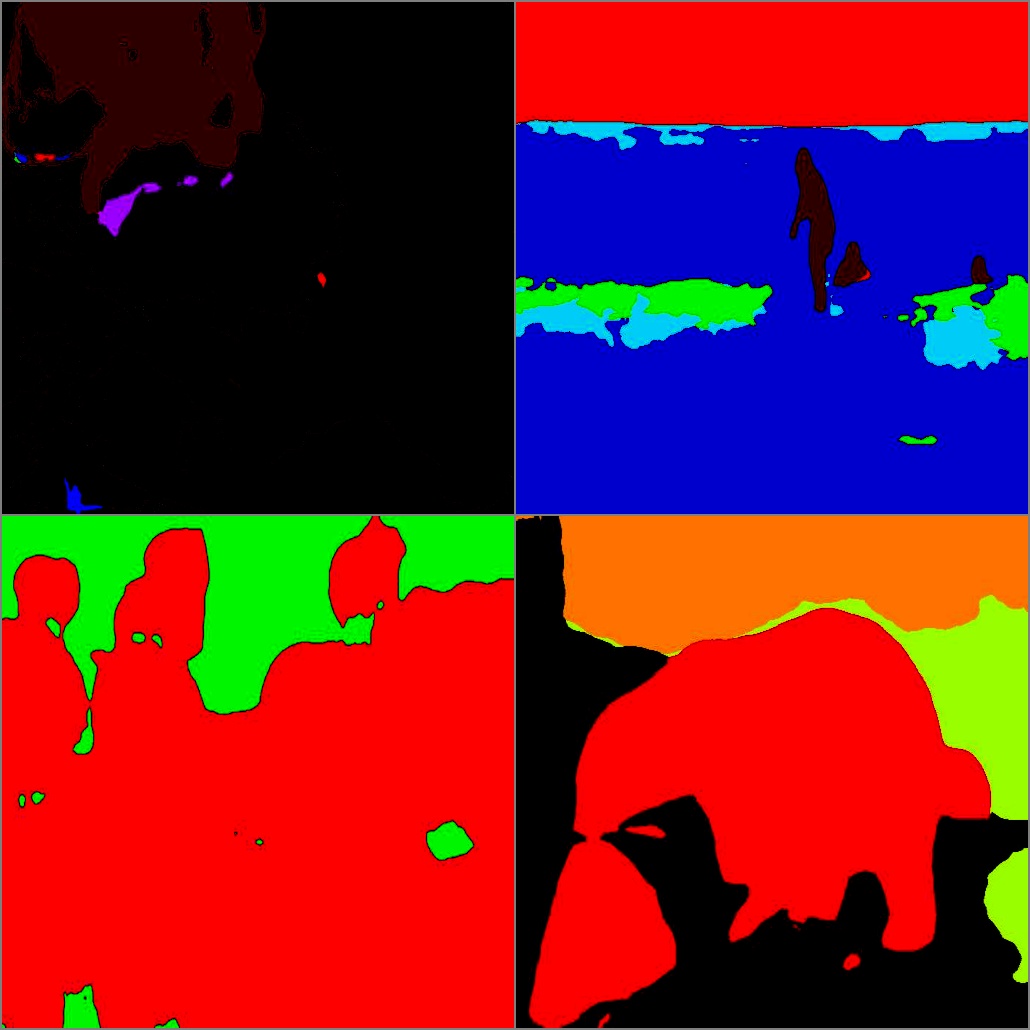}
  \caption{Canny Maps}
\end{subfigure}
\begin{subfigure}{0.32\textwidth}
  \centering
  \includegraphics[width=\linewidth]  {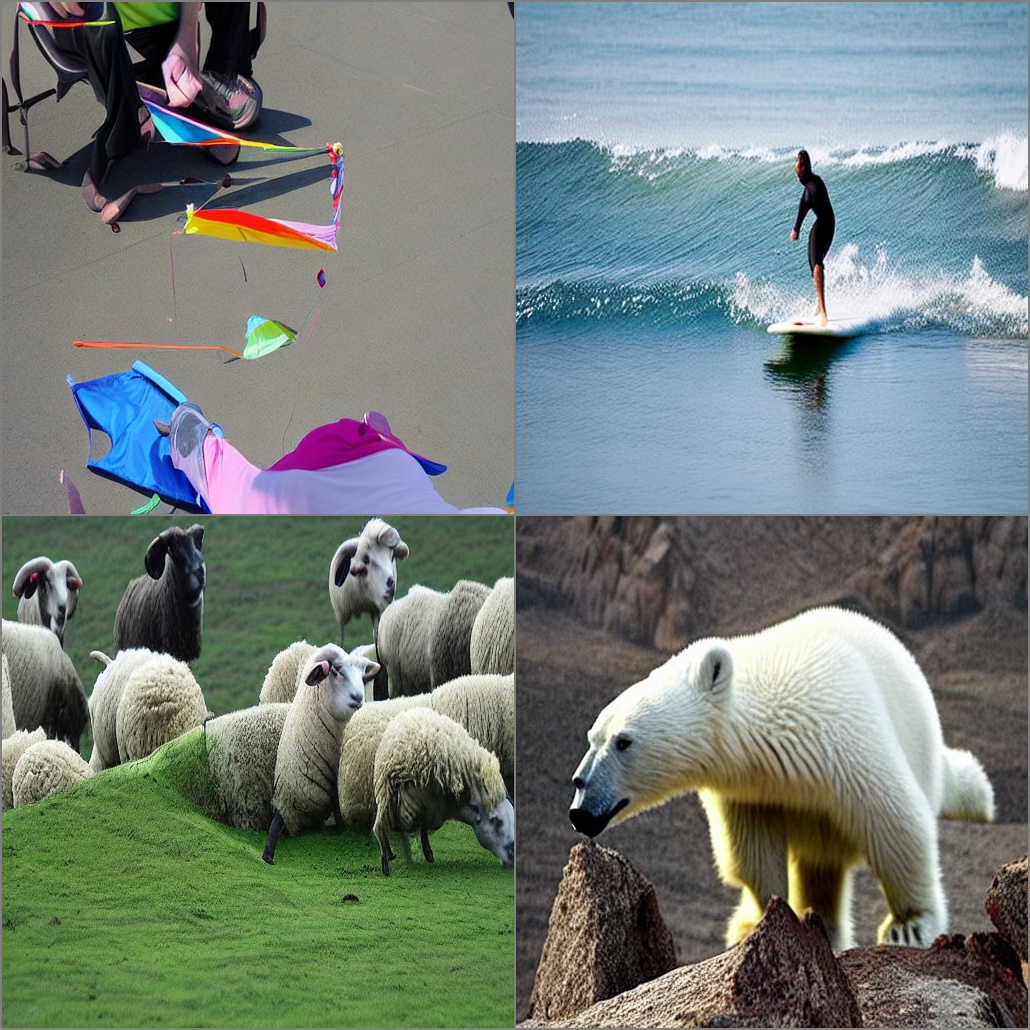}
  \caption{DiffFit\cite{xie2023difffit}}
\end{subfigure}
\begin{subfigure}{0.32\textwidth}
  \centering
  \includegraphics[width=\linewidth]  {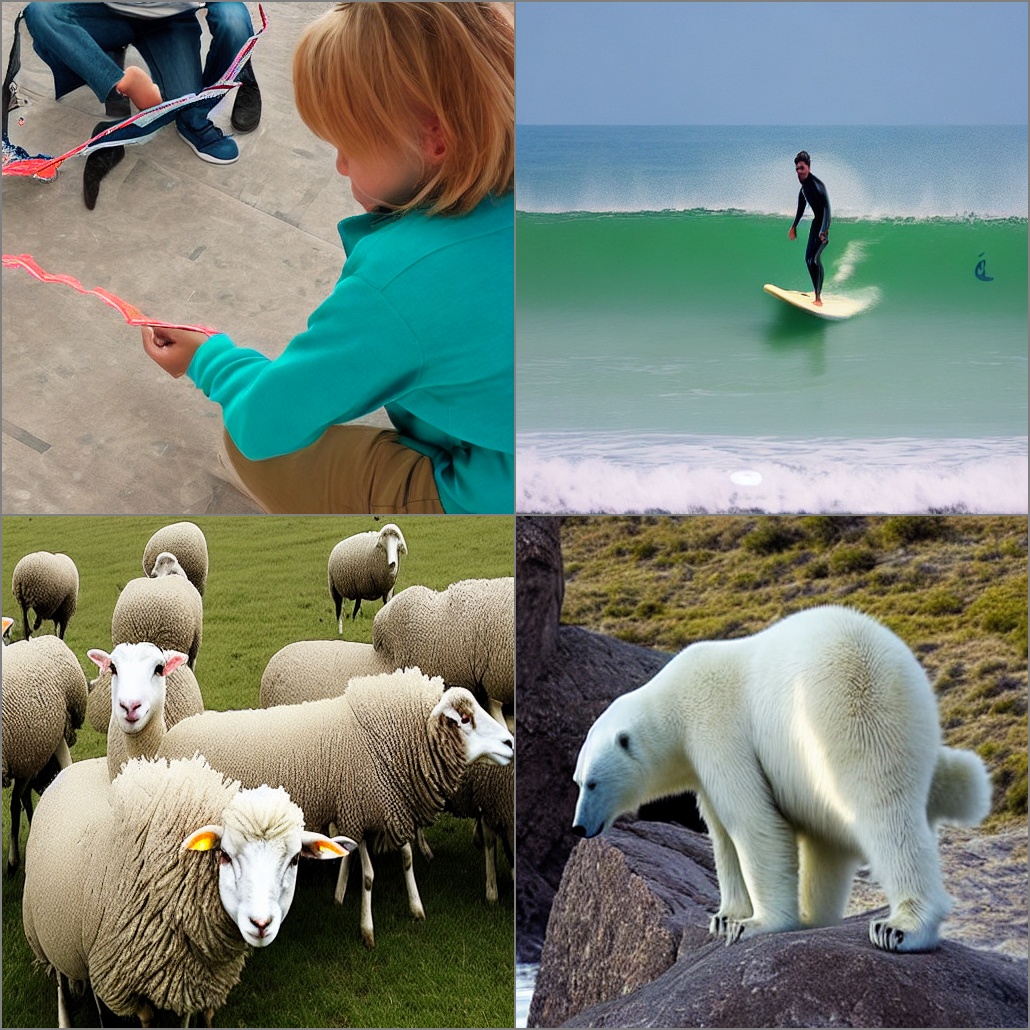}
  \caption{Canny Maps}
\end{subfigure}%
\vspace{-3mm}
\caption{Qualitative comparisons for spatial conditioning with Canny edges maps and Seg Maps. }
\label{fig:cond2}
\vspace{-4mm}
\end{figure}

\noindent \textbf{Spatial Conditioning:} For  spatial conditioning, we compare our performance with ControlNet \cite{zhang2023adding} and DiffFit \cite{xie2023difffit}. We compare the generation quality using the FID score \cite{he2015delving}. We perform this on the COCO stuff validation set \cite{lin2014microsoft}. We compare how close the method matches with its text descriptions using the CLIP score \cite{radford2021learning}. As seen in Table \ref{table:cond}, although our method has only $0.5$ per cent of the trainable parameters as that in Controlnet we achieve similar generation quality in terms of the  performance metrics. Qualitative results for conditional image generation can be seen in Fig \ref{fig:cond2} and \ref{fig:cond1}. More results can be seen in \ref{fig:resintro2} where we also show the diversity in the outputs of our proposed method.in the supplementary material.

\newcommand{\bestcell}{\bfseries}

\begin{table}[tp!]
\begin{center}
\vspace{-1mm}
\caption{Quantitative comparisons for unconditional sampling. Orange  represents best values and blue represents second best values. }
\resizebox{\columnwidth}{!}{%
\begin{tabular}{c|c|cc|cc|cc|cc|}
\toprule
\multirow{2}{*}{Method} & \multirow{2}{*}{Parameters} & \multicolumn{2}{c|}{CUB} & \multicolumn{2}{c|}{Flowers} & \multicolumn{2}{c|}{FFHQ} & \multicolumn{2}{c|}{Caltech} \\
\cmidrule{3-10}
 & & $FID(\downarrow)$ & $IS(\uparrow)$ & $FID(\downarrow)$ & $IS(\uparrow)$ & $FID(\downarrow)$ & $IS(\uparrow)$ & $FID(\downarrow)$ & $IS(\uparrow)$ \\
\midrule
Fine-Tuning & 674M & 18.30 & 5.15 & 20.07 & 3.38 & 15.50 & 4.37 & 45.99 & 14.35 \\
\midrule
LORA\cite{hu2021lora} & 47.7M & 30.97 & \cellcolor{orange!50}6.09 &\cellcolor{blue!25} 18.75 & 3.42 &\cellcolor{blue!25} 22.46 & 4.50 & \cellcolor{blue!25}52.42 & 11.58 \\
DiffFit\cite{xie2023difffit} & 0.48M &  \cellcolor{orange!50}{19.75} & 4.93 & 23.94 & \cellcolor{orange!50}3.61 & 30.00 & \cellcolor{orange!50}4.70 & 54.93 &\cellcolor{blue!25} 12.32 \\
DiffScaler & 48.7M & \cellcolor{blue!25}20.19 &\cellcolor{blue!25} 5.03 & \cellcolor{orange!50}18.07 & \cellcolor{blue!25}3.50 & \cellcolor{orange!50}15.13 & \cellcolor{orange!50}4.70 & \cellcolor{orange!50}51.46 & \cellcolor{orange!50}13.76 \\
\midrule
DiffScaler (CNN backbone) & -& 25.23 & 4.81 & 59.15 & 4.38 & 77.88 & 5.95 & 92.43 & 9.76 \\
\bottomrule
\end{tabular}
}

\label{table:uncond}
\end{center}
\vspace{-3em}
\end{table}

\begin{table*}[tbp]
\centering
\caption{Quantitative comparisons for conditional sampling. All methods were retrained on COCO for equal number of iterations.}
\resizebox{\textwidth}{!}{%
\begin{tabular}{c|c|cc|cc|cc|cc|}
\toprule
\multirow{2}{*}{Method} & \multirow{2}{*}{Parameters} & \multicolumn{2}{c|}{Hed} & \multicolumn{2}{c|}{Depth} & \multicolumn{2}{c|}{Segmentation Map} & \multicolumn{2}{c|}{Canny} \\
\cmidrule{3-10} 
 & & FID$(\downarrow)$ & CLIP$(\uparrow)$ & FID$(\downarrow)$ & CLIP$(\uparrow)$ & FID$(\downarrow)$ & CLIP$(\uparrow)$ & FID$(\downarrow)$ & CLIP$(\uparrow)$ \\
\midrule
ControlNet \cite{zhang2023adding} & 361.2 M (33\%) & 20.99 & 152.03 & 22.09 & 151.15 & 27.56 & 76.18 & 19.22 & 150.59 \\
DiffFit\cite{xie2023difffit} & 11.07M (0.9\%) & 21.20 & 148.06 & 19.51 & \cellcolor{orange!50}155.17 & 23.58 & \cellcolor{orange!50}148.98 & 20.62 & \cellcolor{orange!50}153.00 \\
DiffScaler & 6.85 M (0.5\%) & \cellcolor{orange!50}16.82 & \cellcolor{orange!50}155.26 & \cellcolor{orange!50}17.82 & 153.34 & \cellcolor{orange!50}19.30 & 144.21 & \cellcolor{orange!50}15.80 & 149.46 \\
\bottomrule
\end{tabular}
}

\label{table:cond}
\vspace{-2em}
\end{table*}

\subsection{Ablation Study}

We conduct experiments to understand the contribution of each individual block in our proposed approach.   We conduct all the ablation experiments on the FFHQ dataset using a DiT backbone.
\begin{figure}[tb]
    \centering
    \includegraphics[width=1\linewidth]{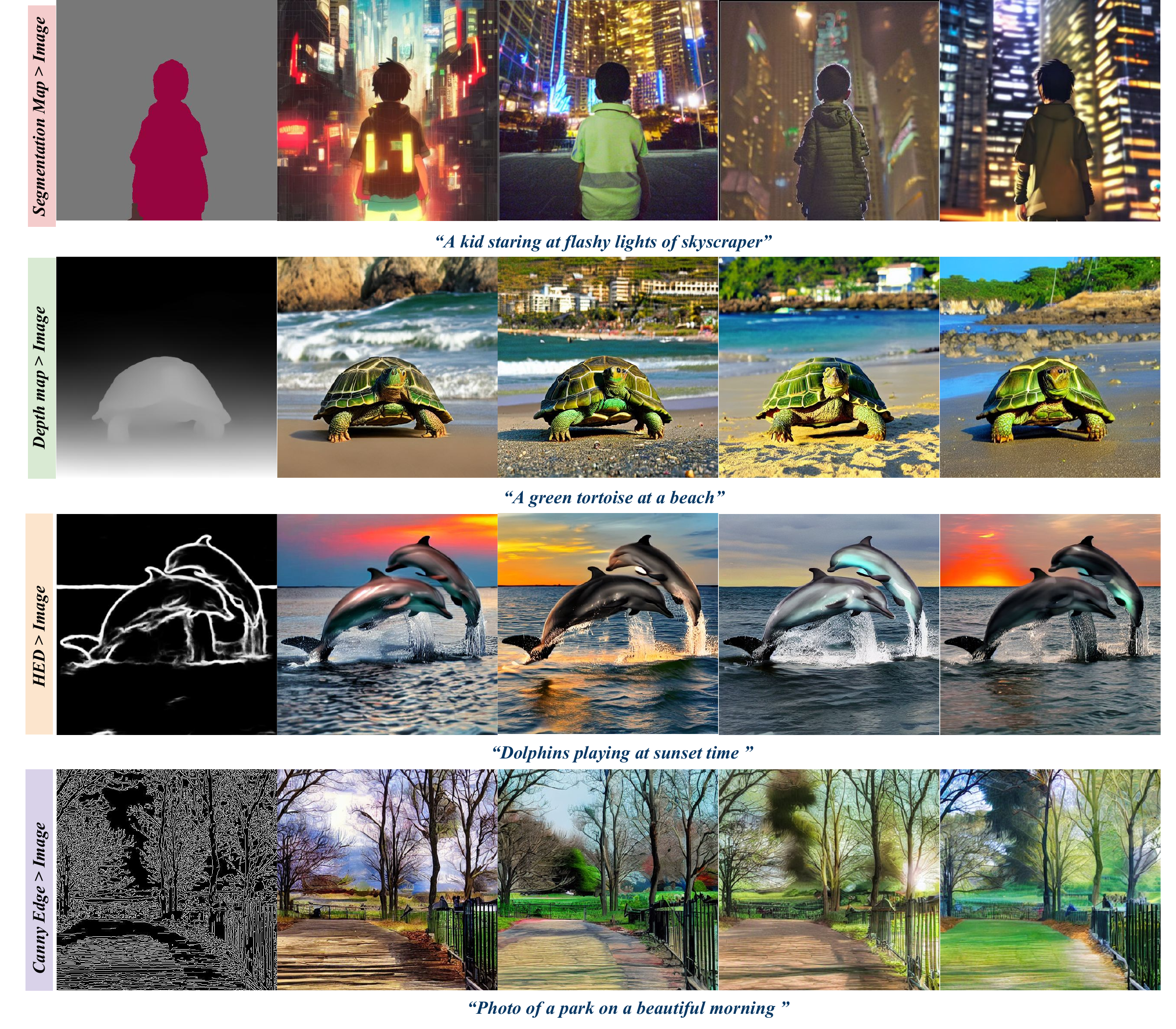}
\caption{Additional results from web scrapped spatial conditionings. }
    \label{fig:resintro2}
\end{figure}
\noindent \textbf{Role of individual components} Table \ref{tab:ablations} (a) provides results for the ablation study conducted to understand the usefulness of each component. For this, we start with the pre-trained model and add the new parameters for bias. We then conduct an experiment by adding scaling parameter on weights. We then add the additional weight basis followed by the scaling parameter for the weight basis. From the performance we can note that each a big boost comes form initializing $W_{up}$ with non-zero values hence reaching better convergence.
\begin{table*}[htbp]
\centering
\caption{Ablation experiments for unconditional sampling in FFHQ dataset showcasing the impact of each term in Affiner.}\label{tab:ablations}
\subfloat[
Individual components\label{tab:insert_form}
]{
\begin{minipage}{0.3\textwidth}
\centering
\setlength{\tabcolsep}{4pt} 
\renewcommand{\arraystretch}{1} 
\begin{tabular}{|c|c|c|c|}
\hline
layers & \#params & FID & IS \\
\hline
b & 0.48M & 26.26 & \cellcolor{orange!50}4.96 \\
a & 0.48M & 20.03 & 4.42 \\
$W_{up}$ & 47.7M & 15.71 & 4.13 \\
Full & 47.7M & \cellcolor{orange!50}15.13 & 4.70 \\ 
\hline
\end{tabular}
\end{minipage}
}
\hspace{3em} 
\subfloat[
Latent dimension $d$\label{tab:s_factor}
]{
\begin{minipage}{0.3\textwidth}
\centering
\setlength{\tabcolsep}{5pt} 
\renewcommand{\arraystretch}{1} 
\begin{tabular}{|c|c|c|c|}
\hline
dim & \#params & FID & IS \\
\hline
1 & 0.77M & 19.22 & 4.91 \\
4 & 3.01M & 20.76 & 3.98 \\
16 & 11.9M & 20.38 & \cellcolor{orange!50}5.07 \\
64 & 48.7M & \cellcolor{orange!50}15.13 & 4.70 \\
\hline
\end{tabular}
\end{minipage}
}
\vspace{-0.5cm}
\end{table*}

\noindent \textbf{Analysis on latent dimension of basis function} In Table \ref{tab:ablations} (b), we conduct experiments to analyze the role of the latent dimension in the introduced basis function. As we can see the FID score initially decreases and then increases as the latent dimension is increased. While it can be seen that increasing the dimensionality improves the FID score while also increasing the number of parameters being introduced for the new task.

\noindent \textbf{Potential Impacts:} Like any other method that enables low compute tuning of large models,  the possibility of using such a technique for harmful uses is also present. Hence, care must be taken while utilizing the model to generate results and while training for new tasks.

\section{Conclusion}
We proposed a scaling strategy for diffusion models such that it can perform multiple tasks using a single model. Our scaling strategy compromises of adding a lightweight block at each layer to efficiently utilize useful subspaces from the pre-trained model as well as learn new subspaces if needed for the new task at hand. The parameters added for each task is very minimal allowing a single diffusion model to scale efficiently. We conduct experiments for both conditional as well as unconditional generative tasks where we show that our approach can generate high quality images over multiple tasks and datasets with a single model. We also show that our approach can be used on both CNN as well as transformer-based methods. In the future, we hope to see how our proposed method would fare for other tasks aside from image generation.

{\small
\bibliographystyle{plain}
\bibliography{references}
}

\title{Supplementary material: Diffscaler: Enhancing the Generative Prowess of Diffusion Transformers
}

\titlerunning{Diffscaler}


\author{Nithin Gopalakrishnan Nair\inst{1}\orcidlink{0000-1111-2222-3333} \and
Jeya Maria Jose Valanarasu\inst{2} \and
Vishal M Patel\inst{1}\orcidlink{0000-0002-5239-692X}}

\authorrunning{Nair et al.}

\institute{Johns Hopkins University$^{1}$, Stanford University$^{2}$ \\
\email{\{ngopala2, vpatel36\}@jhu.edu, jmjose@stanford.edu}\\
}

\maketitle

\section{Difference between Diffscaler and ControlNet}

ControlNet added the capability of finetuning stable diffusion with a desired condition by first obtaining \{\textit{Image, Caption, Condition}\} triplets and finetuning a parallel encoder of the stable diffusion according to the image caption pairs. The parallel encoder is connected to the base networks with convolutional layers initialized with zeroes which account for a large number of parameters. More often than not, the conditions are independent and align  with the text caption. As an upgrade to ControlNet, DiffScaler can be used for conditional generation across multiple tasks by using the proposed reparametrization technique and modifying each trainable layer of the encoder of the model and train the newly added parameters for the new task at hand. {\bf{With our formulation, there is no need for a separate encoder or zero convolutions like in ControlNet}}. We obtain comparable performance with just  $7M$ trainable parameters whereas that of ControlNet is about $300M$.

\section{More Details}

The learning rate for all our experiments was set to $1e^{-3}$.

\textbf{Sampling Technique:}For all unconditional experiments, DDIM sampling \cite{song2020denoising} was used, and 100 sampling steps were performed.  For conditional cases, 50 steps of sampling and DDIM sampling were used.

\textbf{Resolution:} The reported FID scores correspond to FID obtained using 5000 samples. All unconditional samples were synthesized at a resolution of $256\times 256$ Stable Diffusion v1.5 was used as the backbone and the images were synthesized at a resolution of $512 \times 512$. For metric computation, 5000 samples were synthesised  according to condtioning generated from COCO-Stuff dataset. 

\textbf{Images for Conditional Generation:-} For the examples illustrated in the paper, the conditioning maps were obtained by passing the images generated through midjourney through the proposed annotation schemes. The images shown in the supplementary material is from COCO-stuff dataset.

\textbf{Intermediate dimension for DiT and CNN backbone.} For DiT the intermediate backbone was chosen as 64 as the hidden dimension of DiT is 1152. FOr Stable Diffusion and other CNN backbones, the intermediate dimension was chosen as 32 since, the number of CNN channels is often less than or equal to 512.

\textbf{Training only the transformer Layers for DiT:} In the case of DiT, most of the trainable parameters are present in the transformer layers, moreover there exist a patch embedding layer which is a convolutional layer and a unpatch layer. Both of these layers have a very low number of parameters and are inherently used for obtaining the embeddings. Since the base model itself can give representative embeddings, we do not retrain this layer. Regarding timestep embedding  these are embedded into the network through adaptive instance normalization layers, hence we train adaptive normalization layers in every transformer block along with the transformer parameters.

\begin{figure}
\centering
\begin{subfigure}{0.8\textwidth}
  \centering
  \includegraphics[width=0.95\linewidth]{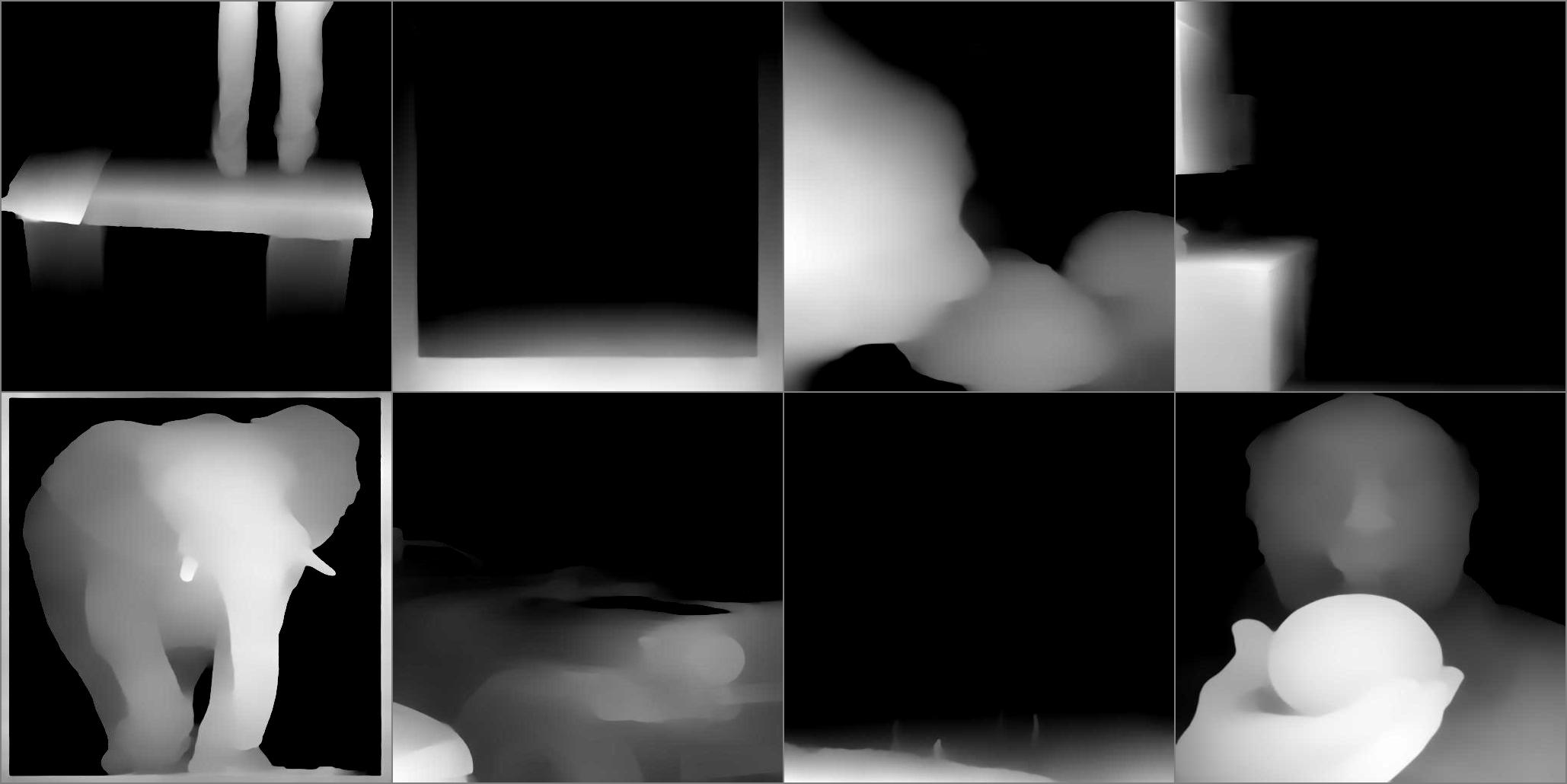}
  \caption{Depthmaps}
  \label{fig:sub11}
\end{subfigure}%
\\
\begin{subfigure}{0.8\textwidth}
  \centering
  \includegraphics[width=0.95\linewidth]{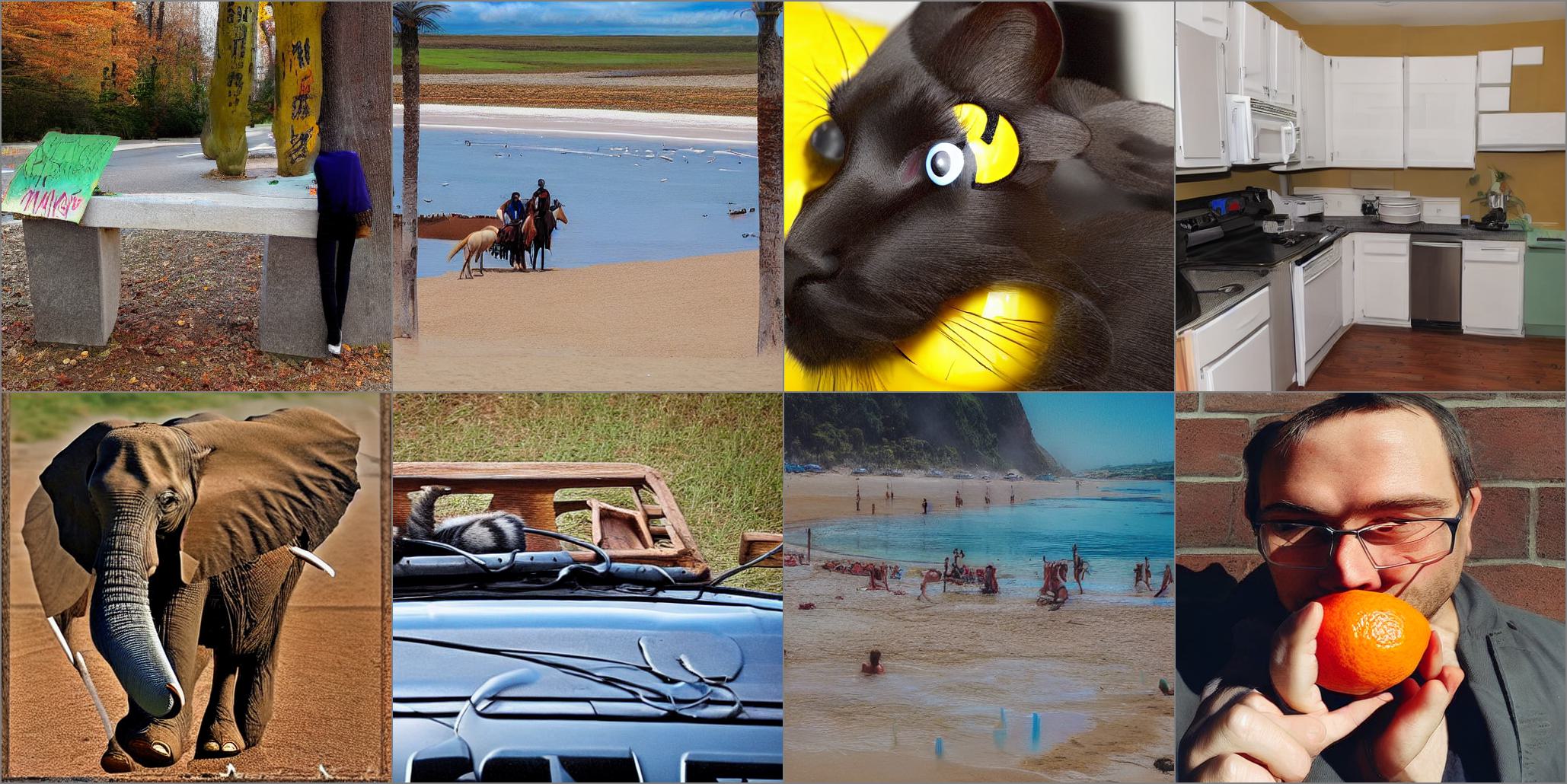}
  \caption{DiffFit}
  \label{fig:sub12}
\end{subfigure}%
\\
\begin{subfigure}{0.8\textwidth}
  \centering
  \includegraphics[width=0.95\linewidth]{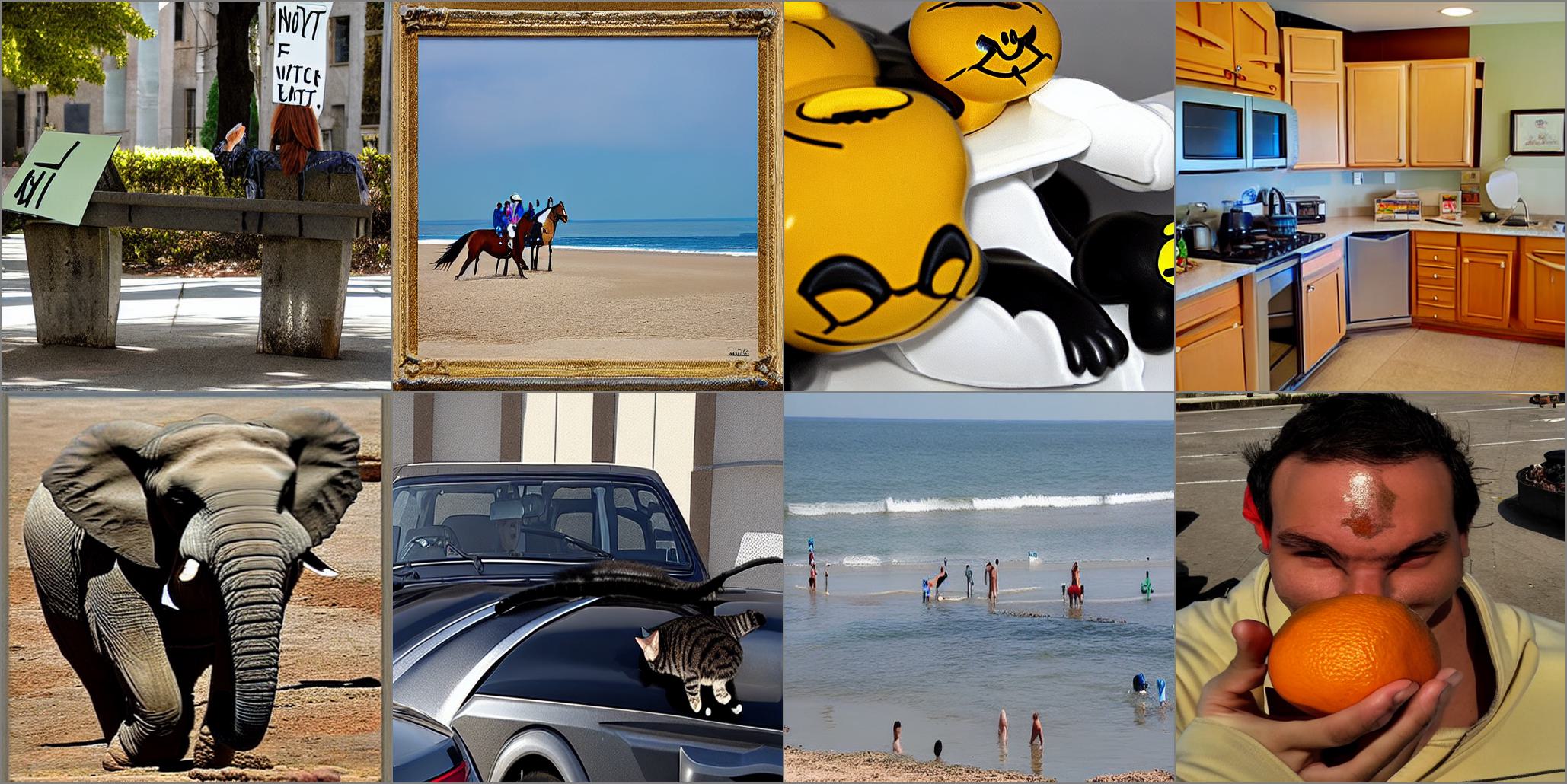}
  \caption{ControlNet}
  \label{fig:sub13}
\end{subfigure}%
\\
\begin{subfigure}{0.8\textwidth}
  \centering
  \includegraphics[width=0.95\linewidth]{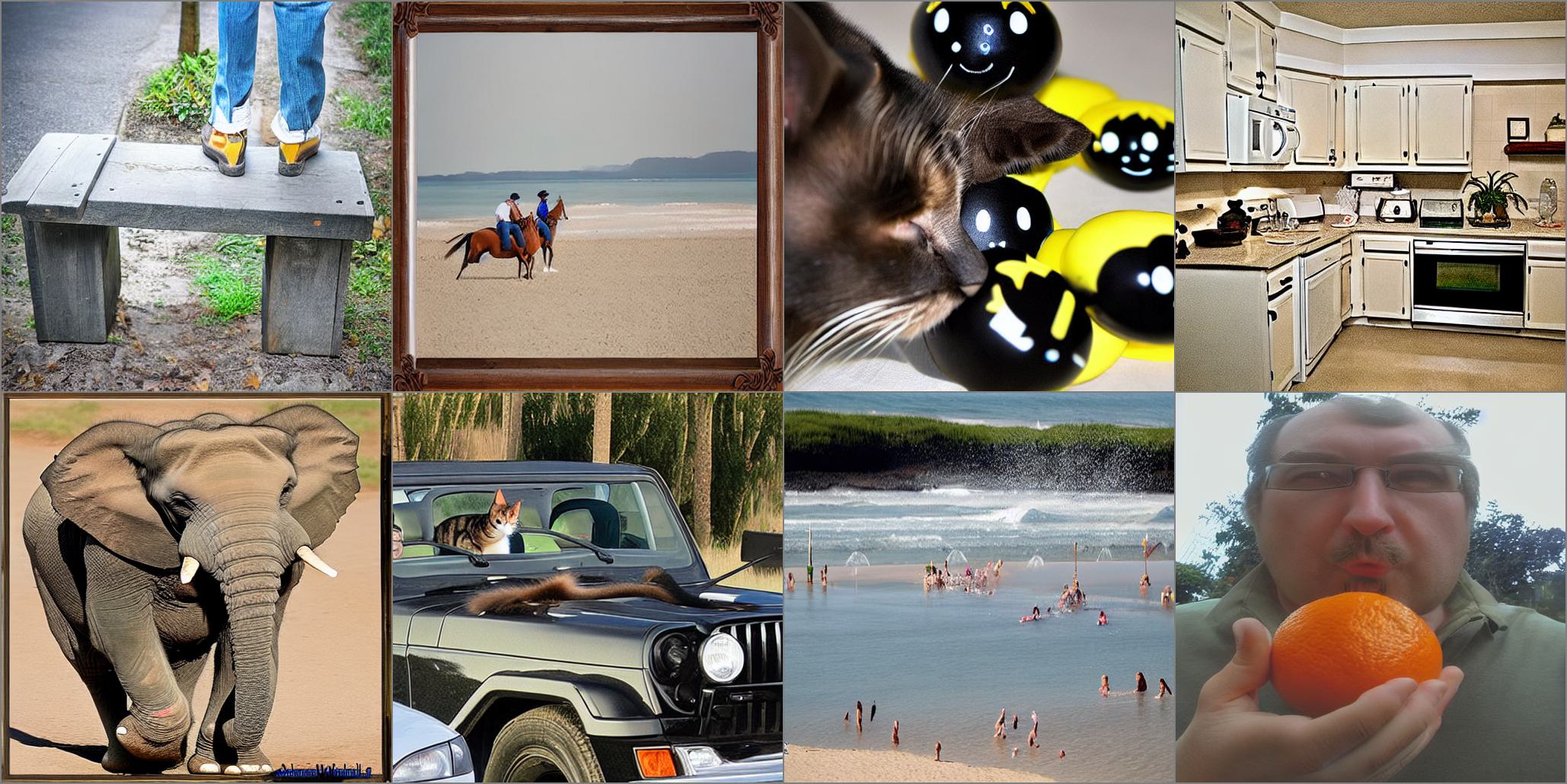}
  \caption{Diffscaler-DiT}
  \label{fig:sub14}
\end{subfigure}%
\caption{Non-cherry picked Conditional comparisons}
\label{fig:test11}
\end{figure}

\begin{figure}
\centering
\begin{subfigure}{0.8\textwidth}
  \centering
  \includegraphics[width=0.95\linewidth]{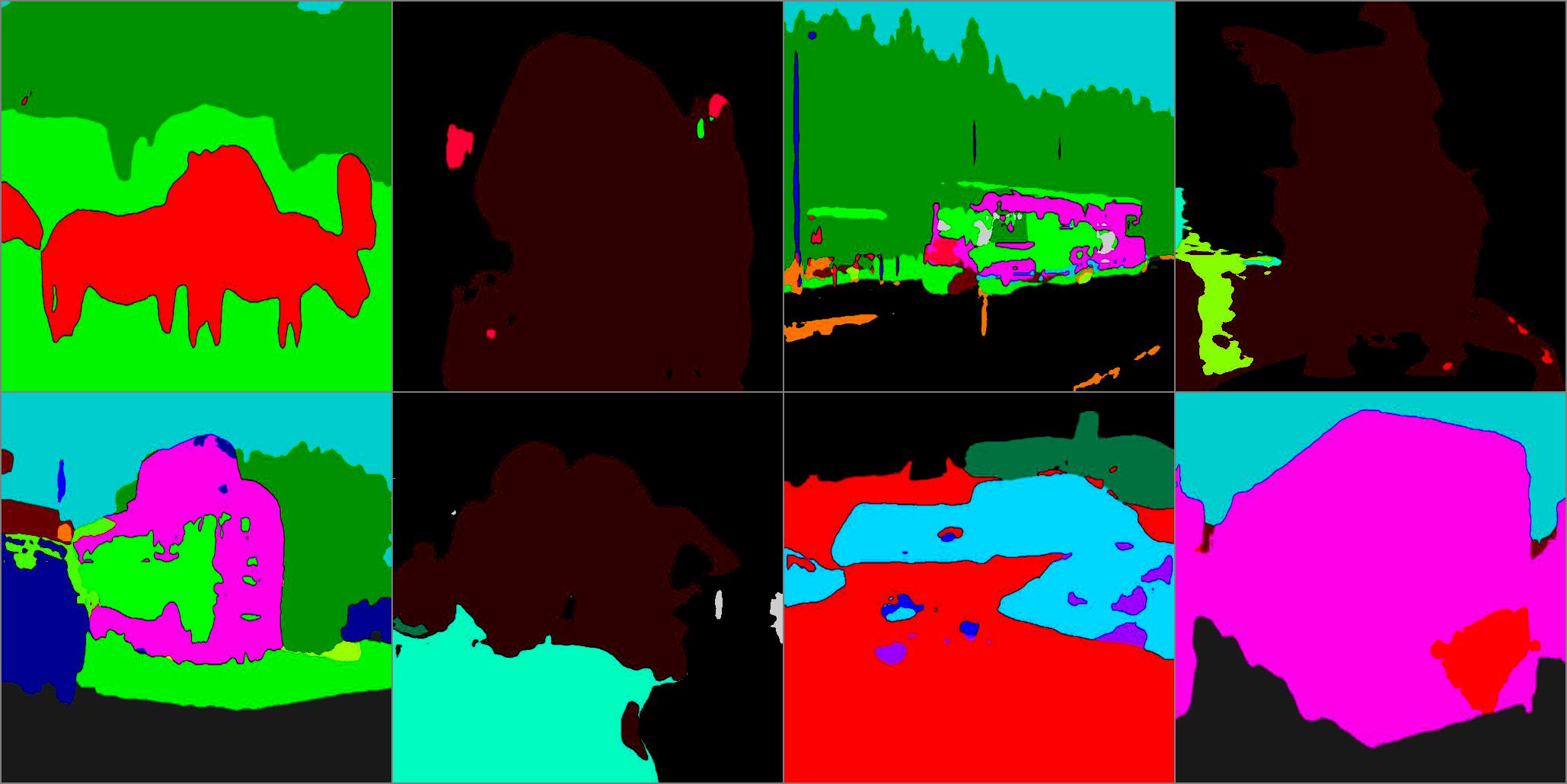}
  \caption{Segmentation maps}
  \label{fig:sub21}
\end{subfigure}%
\\
\begin{subfigure}{0.8\textwidth}
  \centering
  \includegraphics[width=0.95\linewidth]{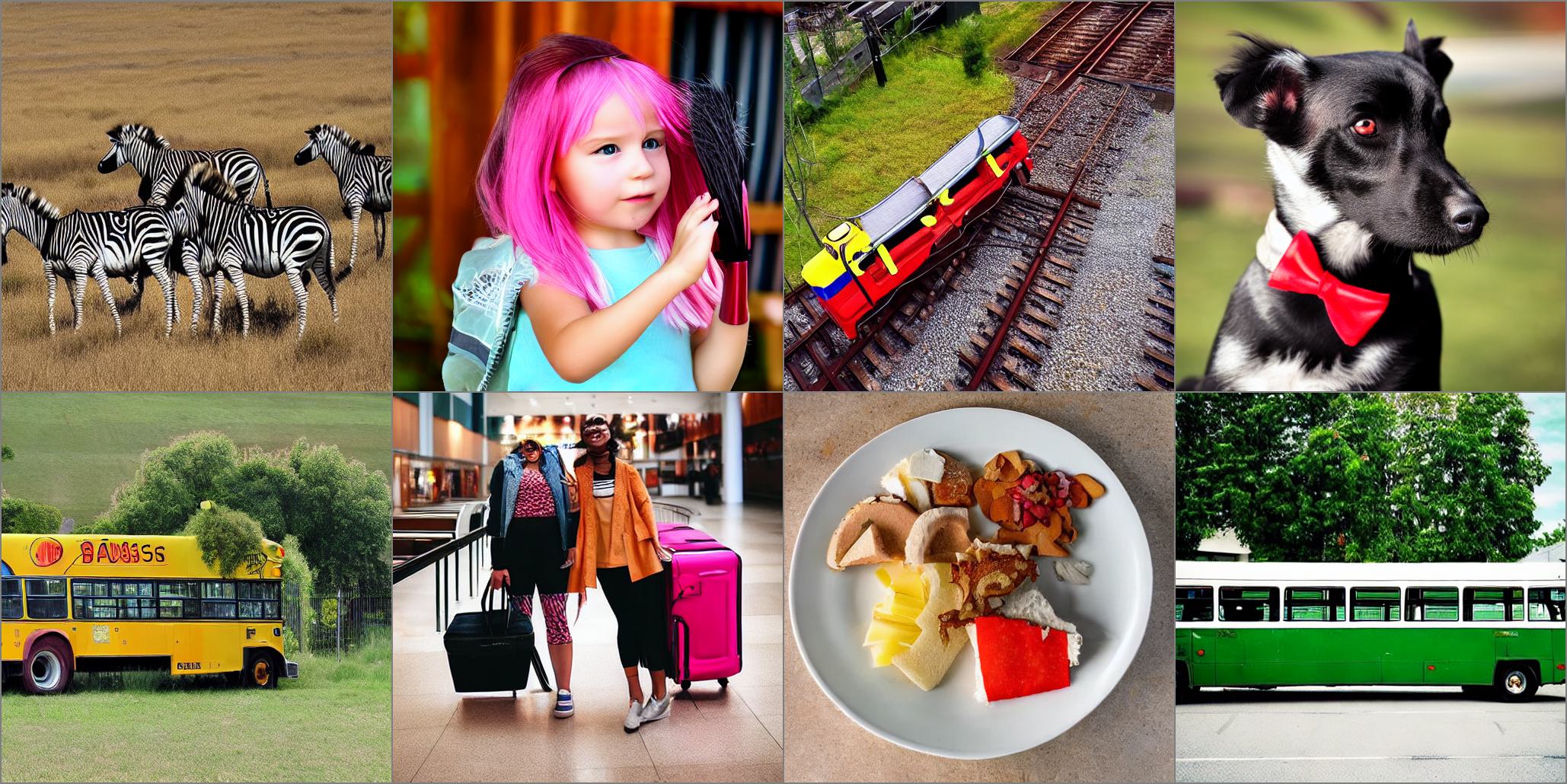}
  \caption{DiffFit}
  \label{fig:sub22}
\end{subfigure}%
\\
\begin{subfigure}{0.8\textwidth}
  \centering
  \includegraphics[width=0.95\linewidth]{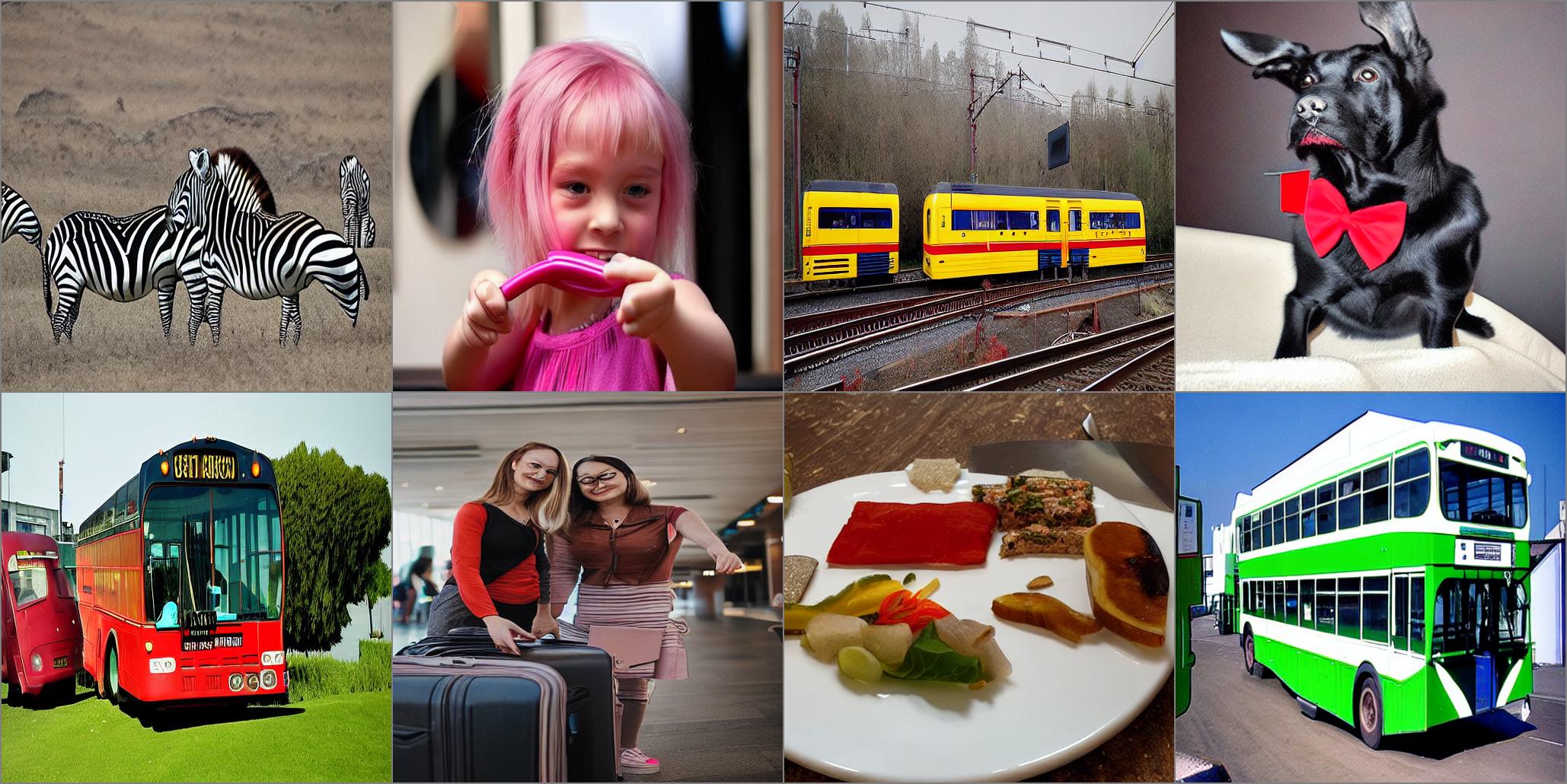}
  \caption{ControlNet}
  \label{fig:sub23}
\end{subfigure}%
\\
\begin{subfigure}{0.8\textwidth}
  \centering
  \includegraphics[width=0.95\linewidth]{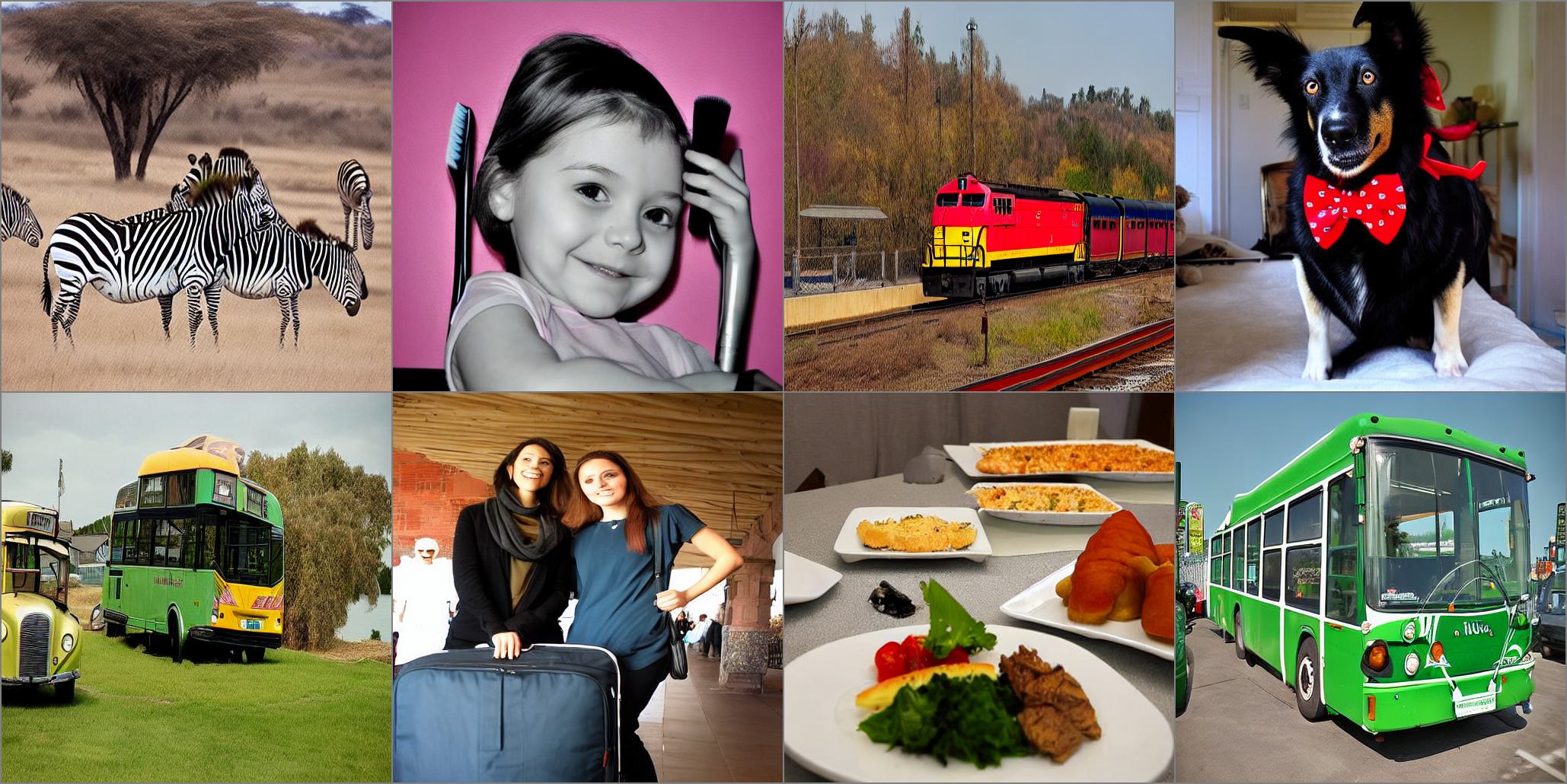}
  \caption{Diffscaler-DiT}
  \label{fig:sub1}
\end{subfigure}%
\caption{Non-cherry picked Conditional comparisons}
\label{fig:test21}
\end{figure}

\begin{figure}
\centering
\begin{subfigure}{0.8\textwidth}
  \centering
  \includegraphics[width=0.95\linewidth]{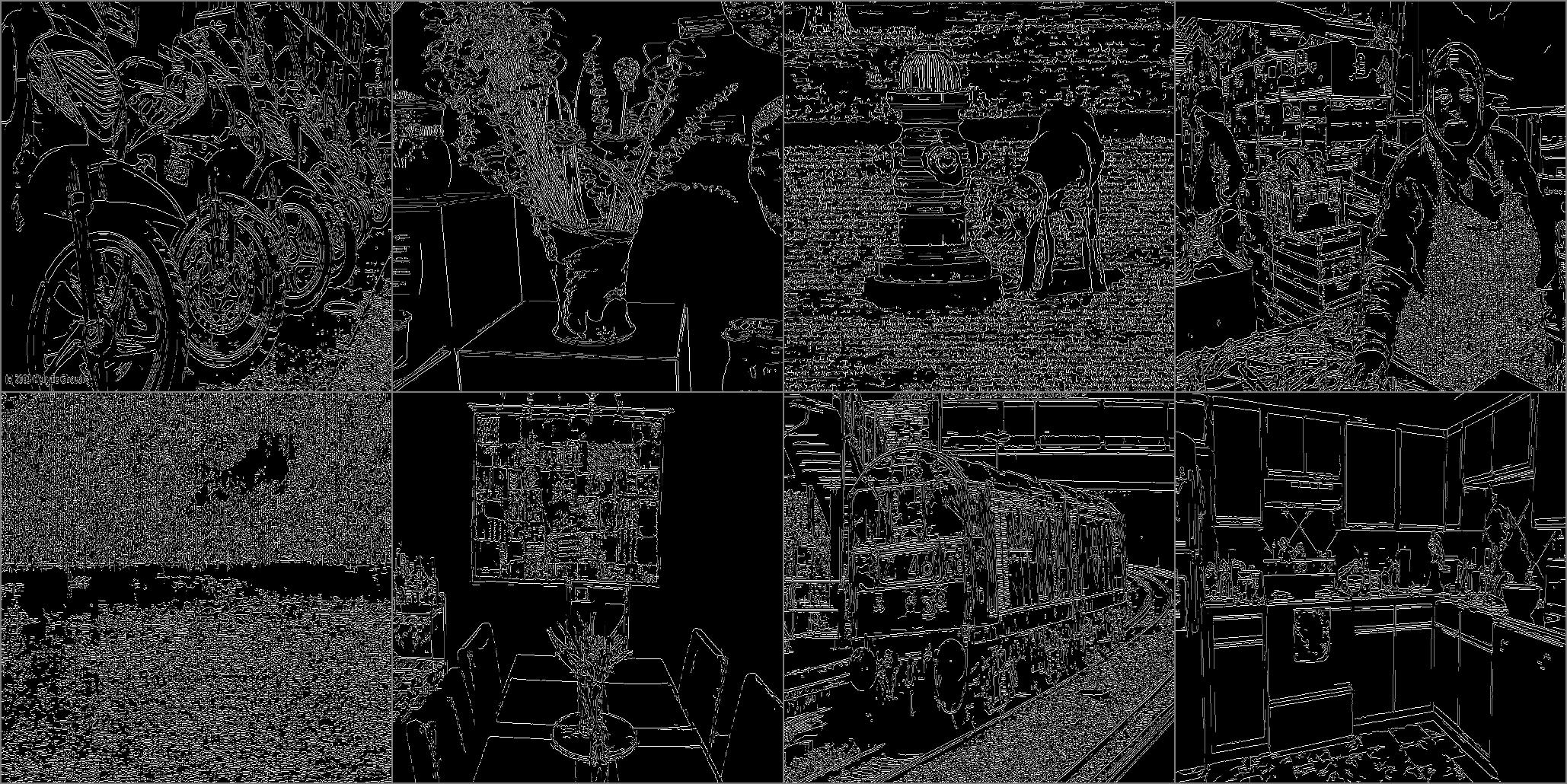}
  \caption{Canny Maps}
  \label{fig:sub31}
\end{subfigure}%
\\
\begin{subfigure}{0.8\textwidth}
  \centering
  \includegraphics[width=0.95\linewidth]{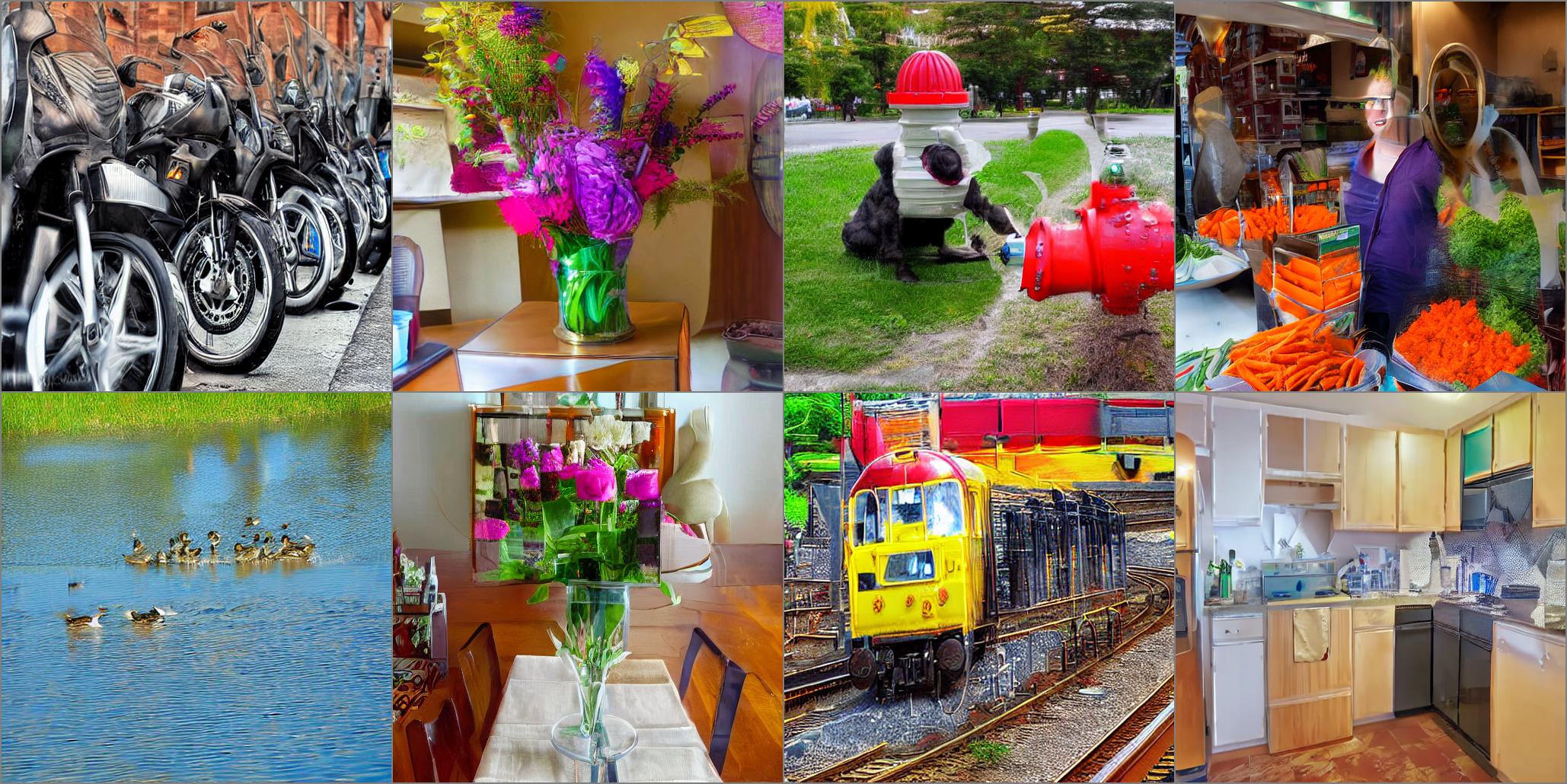}
  \caption{DiffFit}
  \label{fig:sub32}
\end{subfigure}%
\\
\begin{subfigure}{0.8\textwidth}
  \centering
  \includegraphics[width=0.95\linewidth]{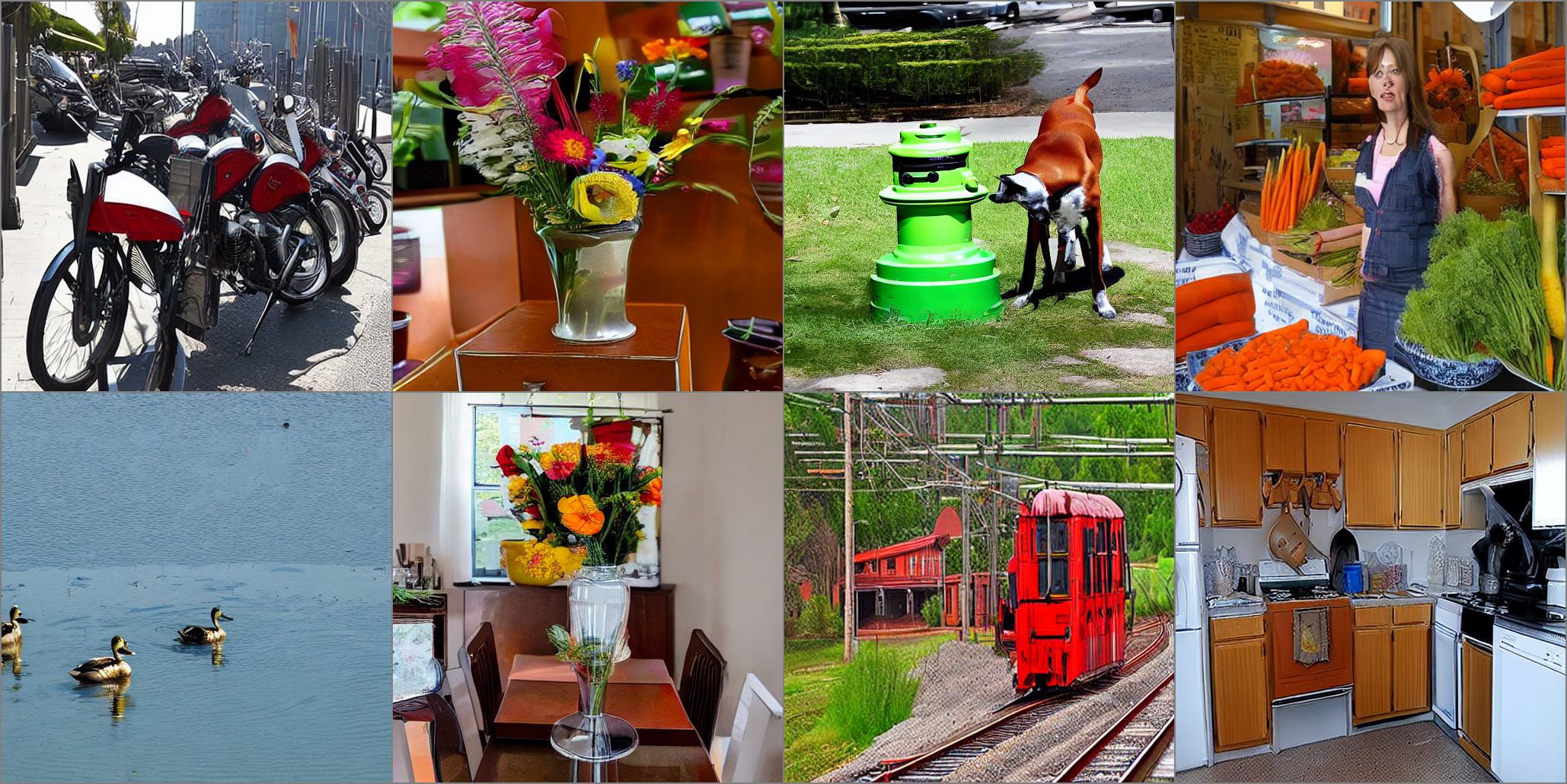}
  \caption{ControlNet}
  \label{fig:sub33}
\end{subfigure}%
\\
\begin{subfigure}{0.8\textwidth}
  \centering
  \includegraphics[width=0.95\linewidth]{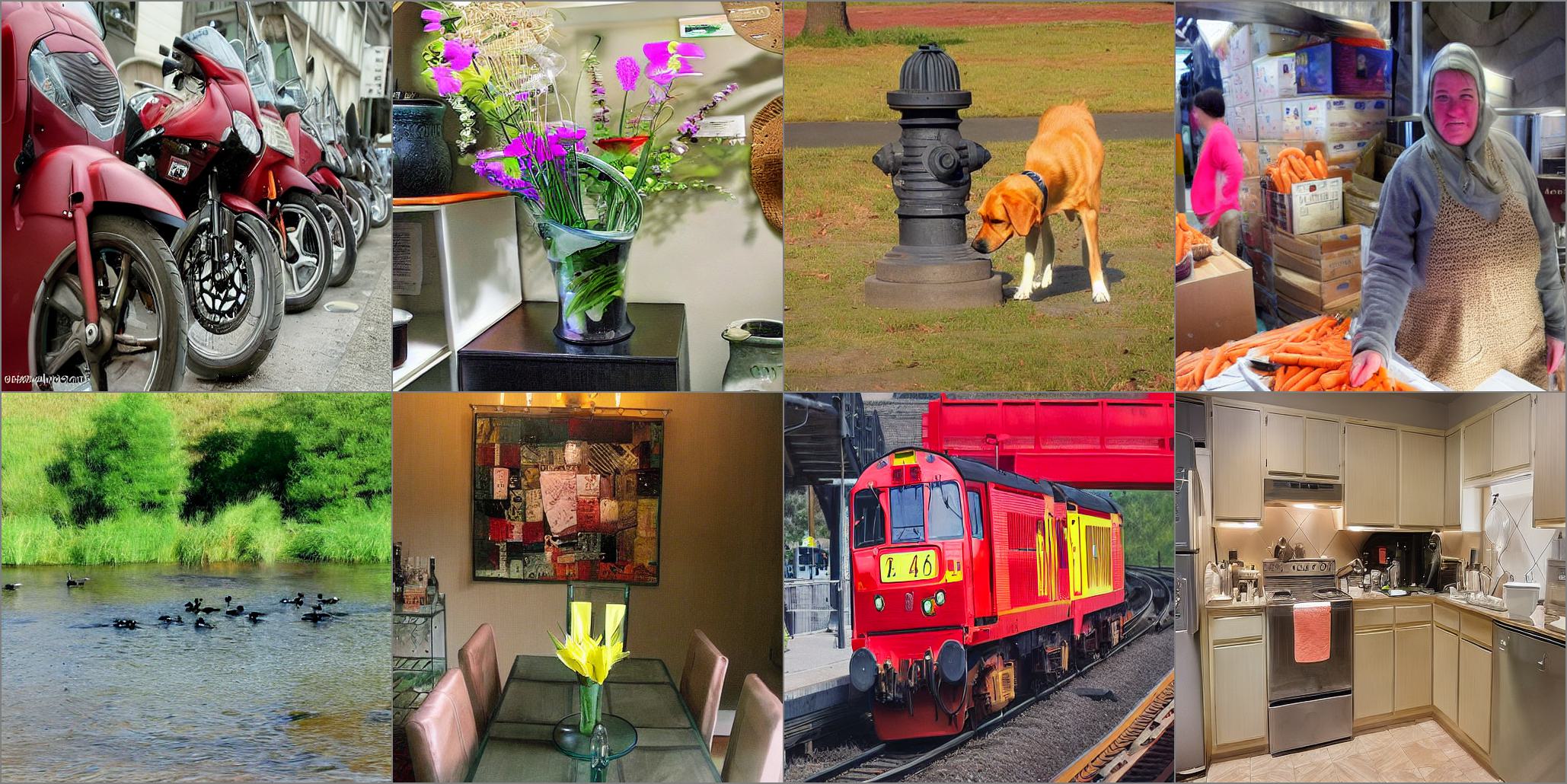}
  \caption{Diffscaler}
  \label{fig:sub34}
\end{subfigure}%
\caption{Non-cherry picked Conditional comparisons}
\label{fig:test31}
\end{figure}

\begin{figure}
\centering
\begin{subfigure}{0.8\textwidth}
  \centering
  \includegraphics[width=0.95\linewidth]{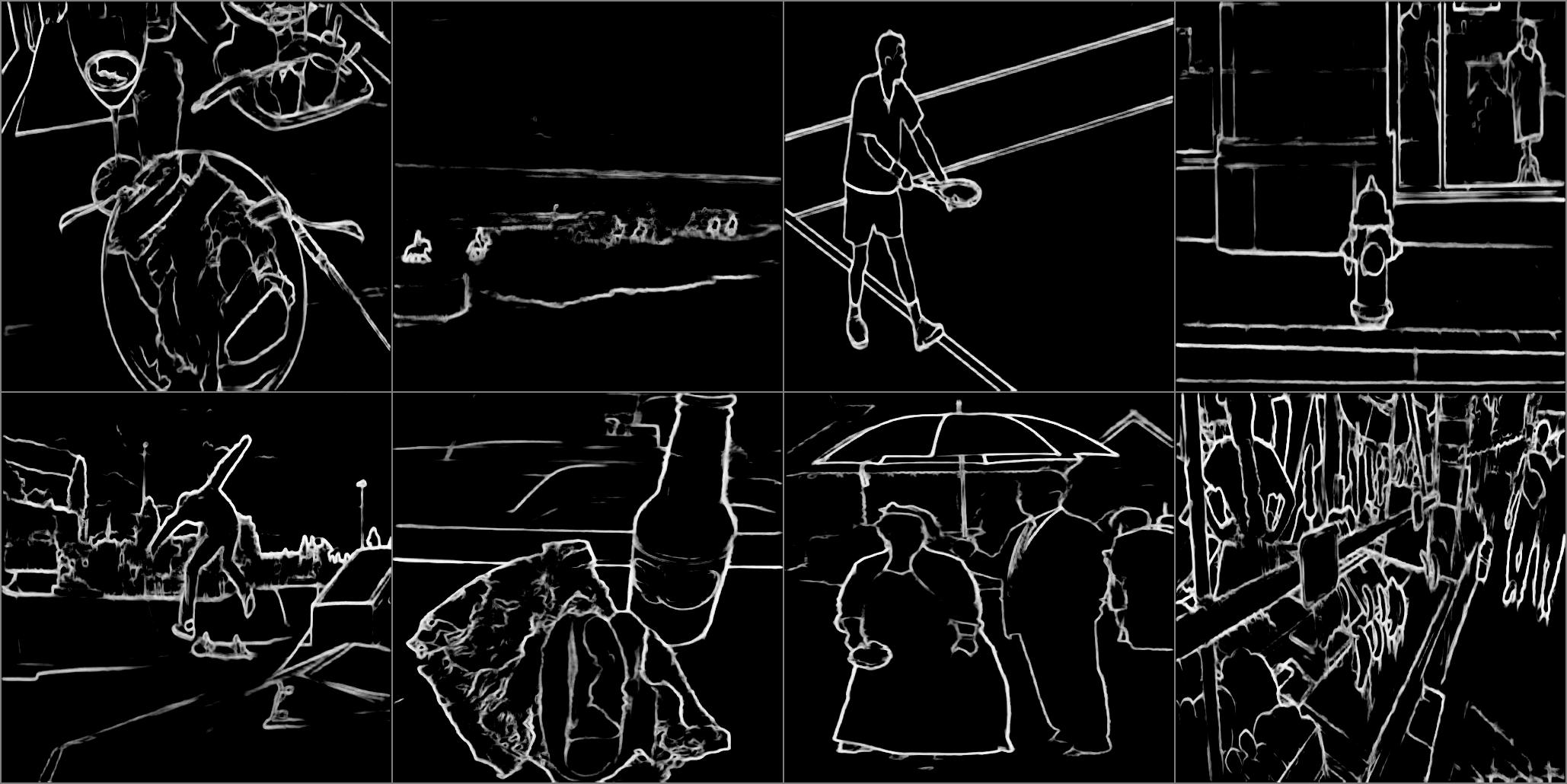}
  \caption{hed maps}
  \label{fig:sub41}
\end{subfigure}%
\\
\begin{subfigure}{0.8\textwidth}
  \centering
  \includegraphics[width=0.95\linewidth]{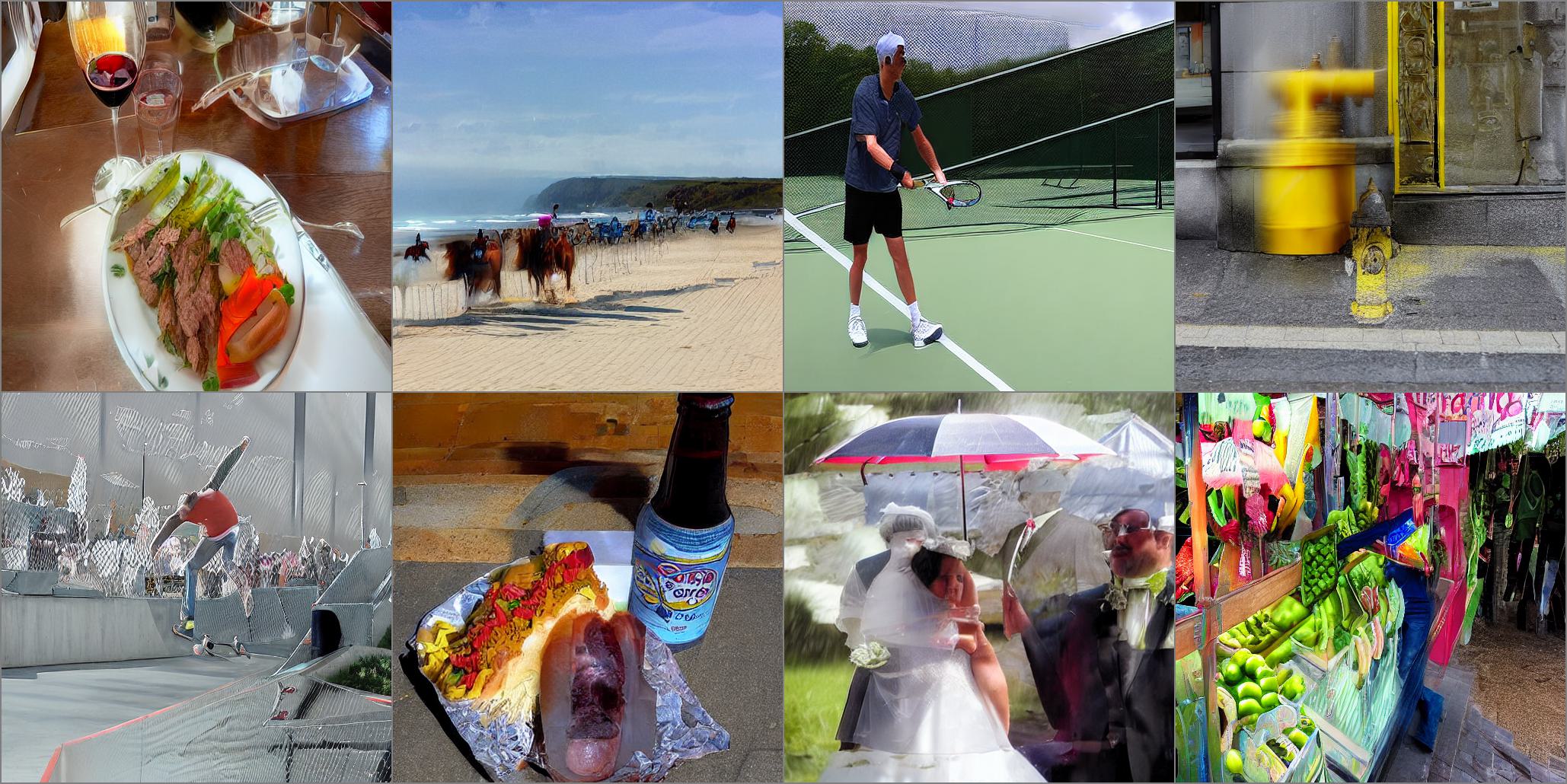}
  \caption{DiffFit}
  \label{fig:sub42}
\end{subfigure}%
\\
\begin{subfigure}{0.8\textwidth}
  \centering
  \includegraphics[width=0.95\linewidth]{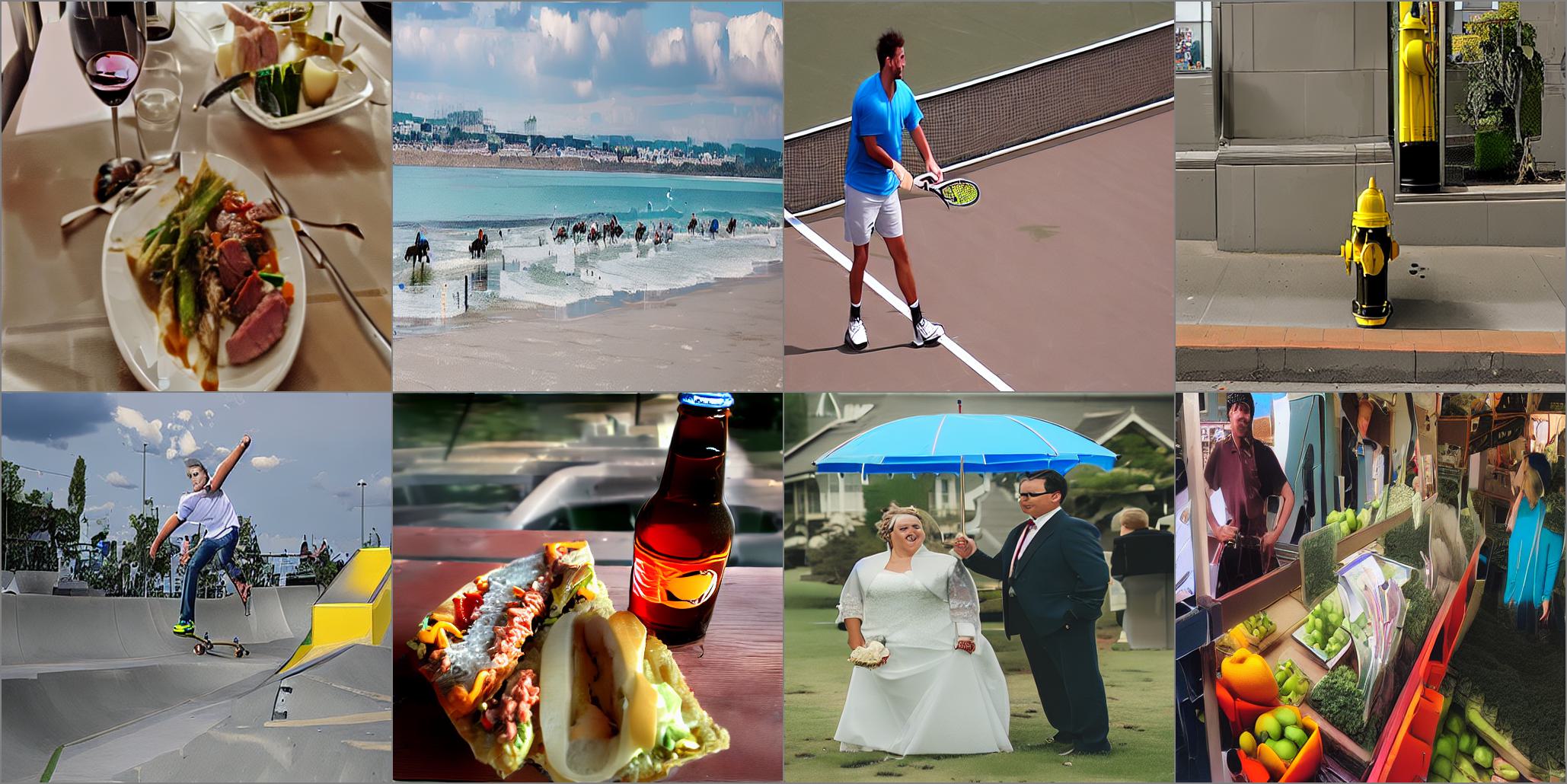}
  \caption{ControlNet}
  \label{fig:sub43}
\end{subfigure}%
\\
\begin{subfigure}{0.8\textwidth}
  \centering
  \includegraphics[width=0.95\linewidth]{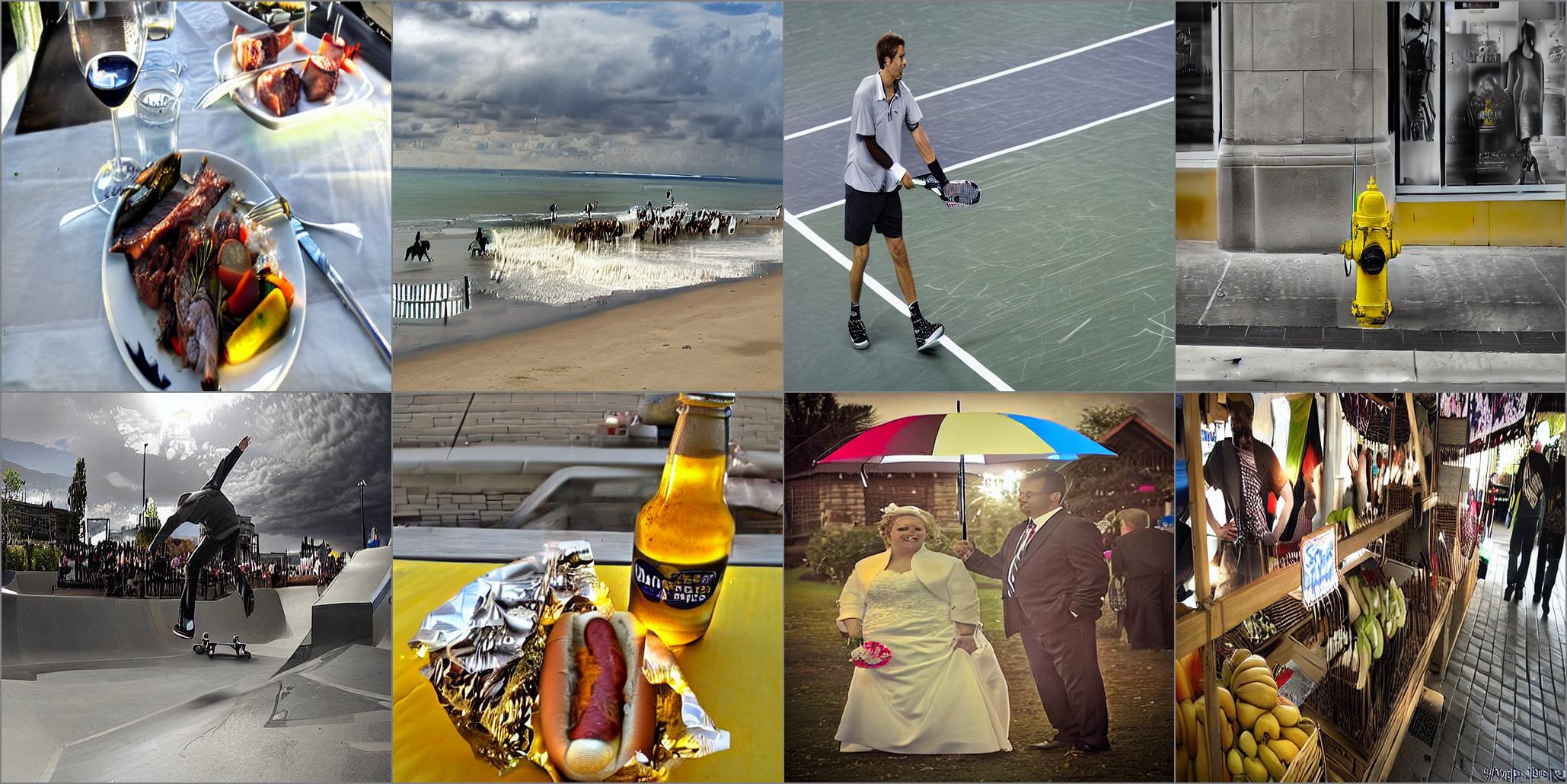}
  \caption{Diffscaler}
  \label{fig:sub44}
\end{subfigure}%
\caption{Non-cherry picked Conditional generation samples.}
\label{fig:test41}
\end{figure}

\begin{figure}
\centering
\begin{subfigure}{0.8\textwidth}
  \centering
  \includegraphics[width=0.95\linewidth]{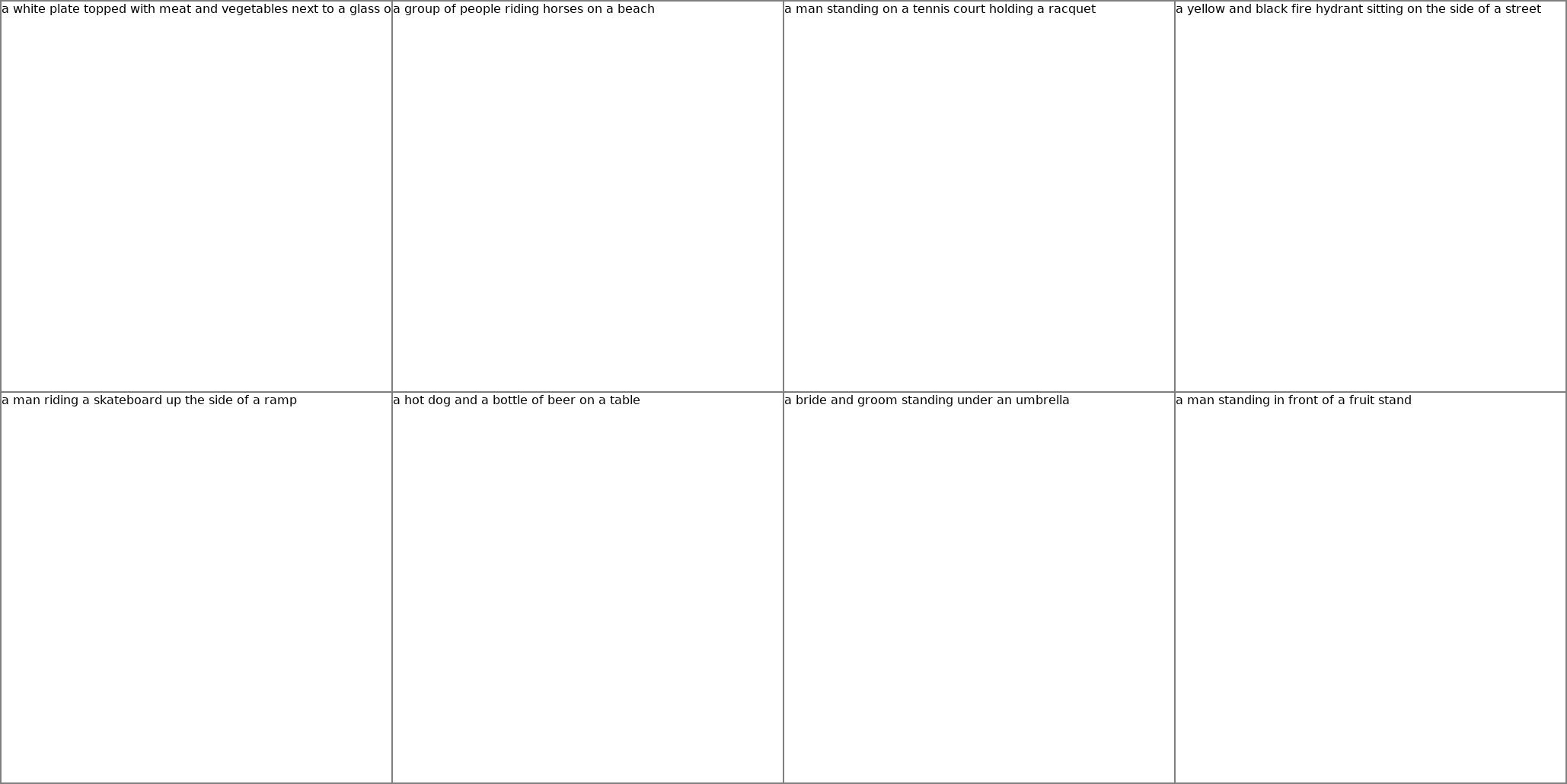}
  \caption{Hed Maps}
  \label{fig:sub51}
\end{subfigure}%
\\
\begin{subfigure}{0.8\textwidth}
  \centering
  \includegraphics[width=0.95\linewidth]{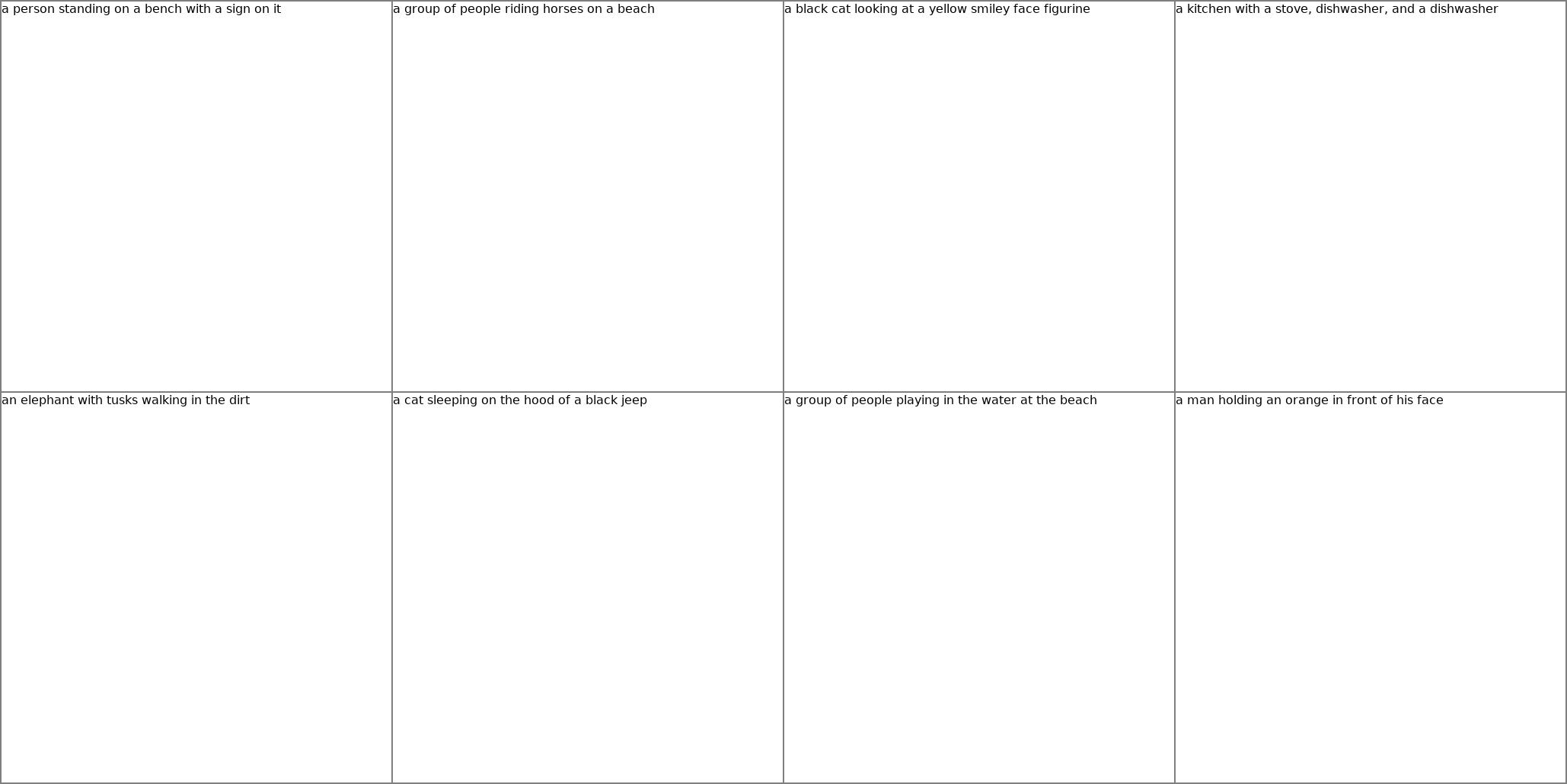}
  \caption{Depth Maps}
  \label{fig:sub52}
\end{subfigure}%
\\
\begin{subfigure}{0.8\textwidth}
  \centering
  \includegraphics[width=0.95\linewidth]{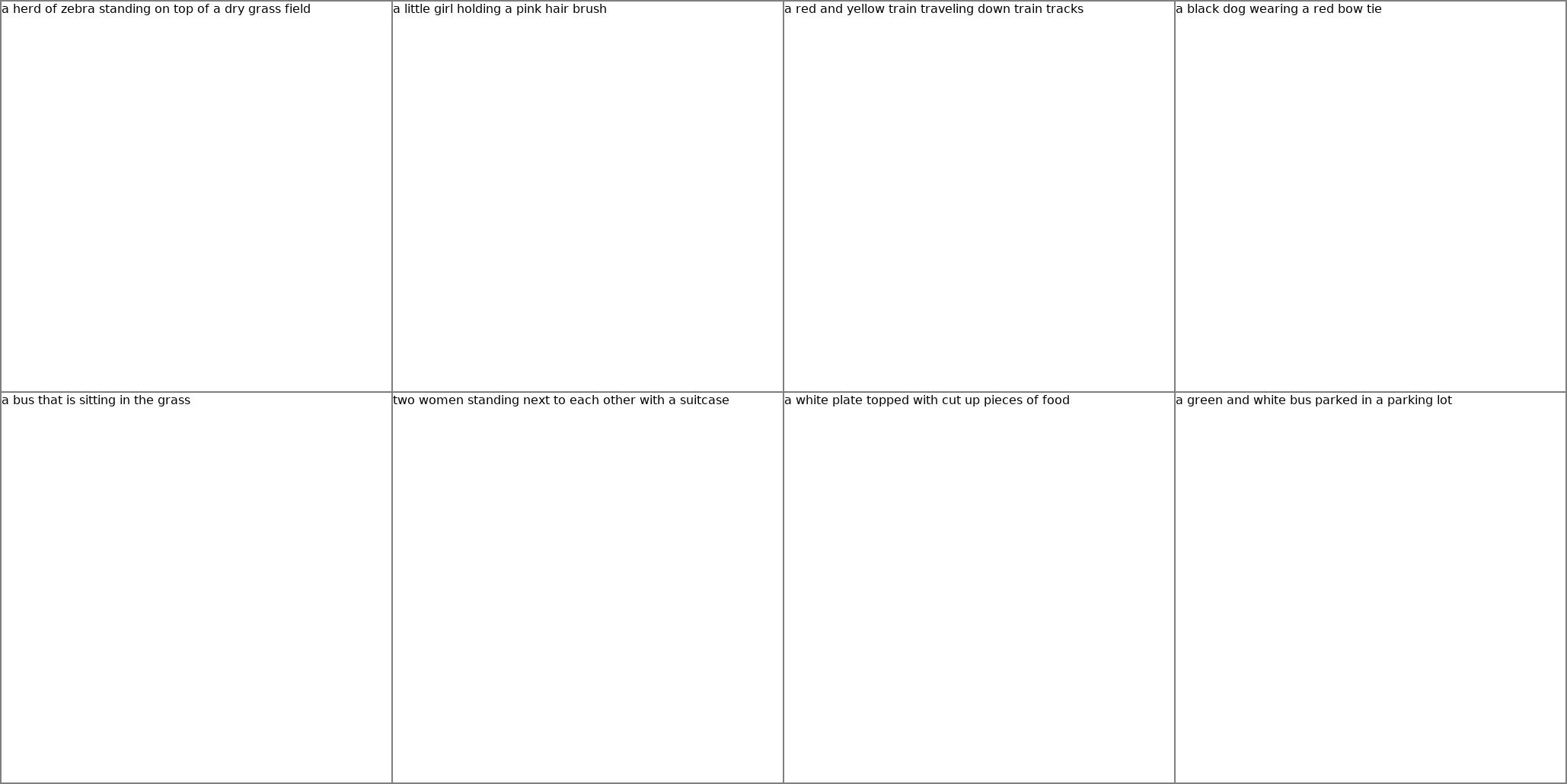}
  \caption{Segmentation Maps}
  \label{fig:sub53}
\end{subfigure}%
\\
\begin{subfigure}{0.8\textwidth}
  \centering
  \includegraphics[width=0.95\linewidth]{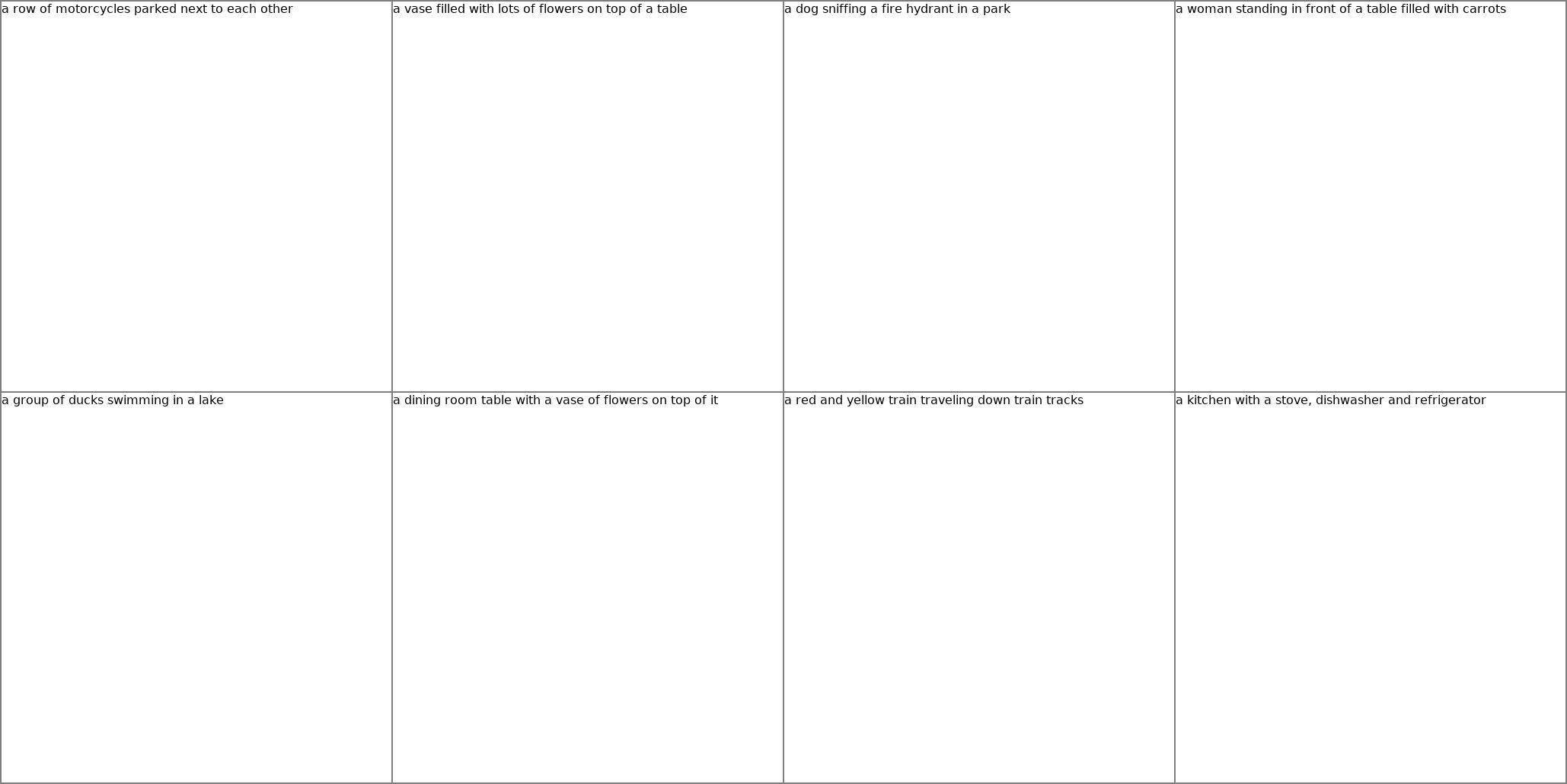}
  \caption{Canny maps}
  \label{fig:sub54}
\end{subfigure}%
\caption{Caption reference for conditional illustrations}
\label{fig:test51}
\end{figure}

\begin{figure}
\centering
\begin{subfigure}{.5\textwidth}
  \centering
  \includegraphics[width=0.95\linewidth]{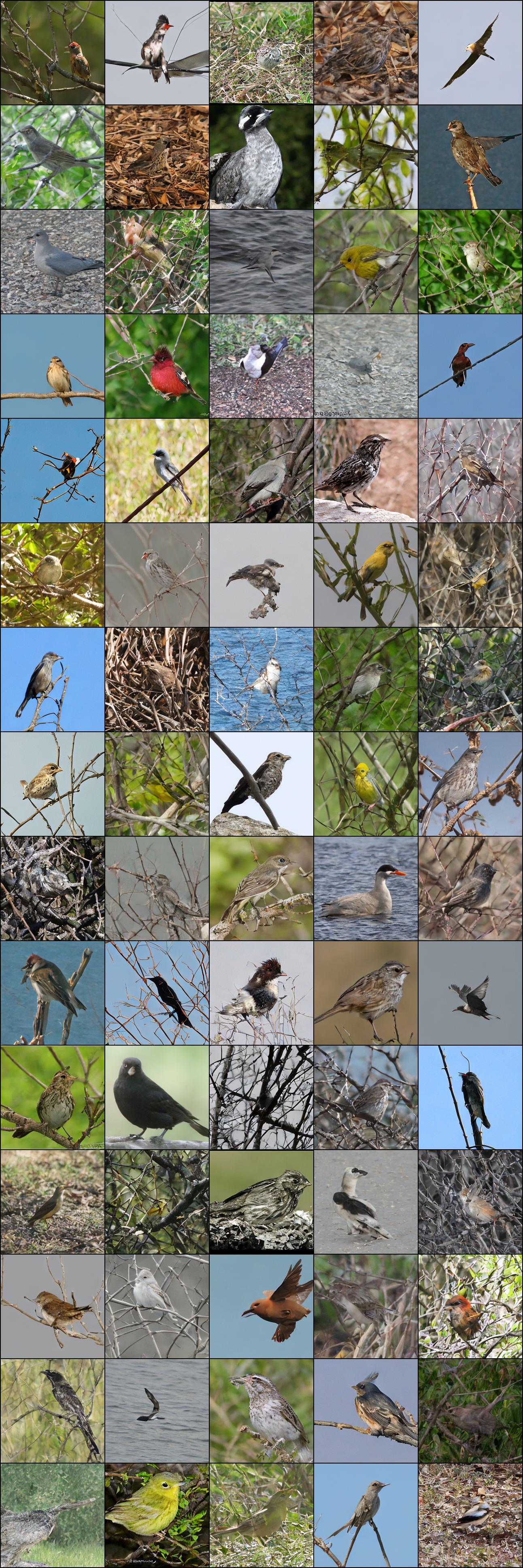}
  \caption{Diffscaler-DiT}
  \label{fig:sub61}
\end{subfigure}%
\begin{subfigure}{.5\textwidth}
  \centering
  \includegraphics[width=0.95\linewidth]{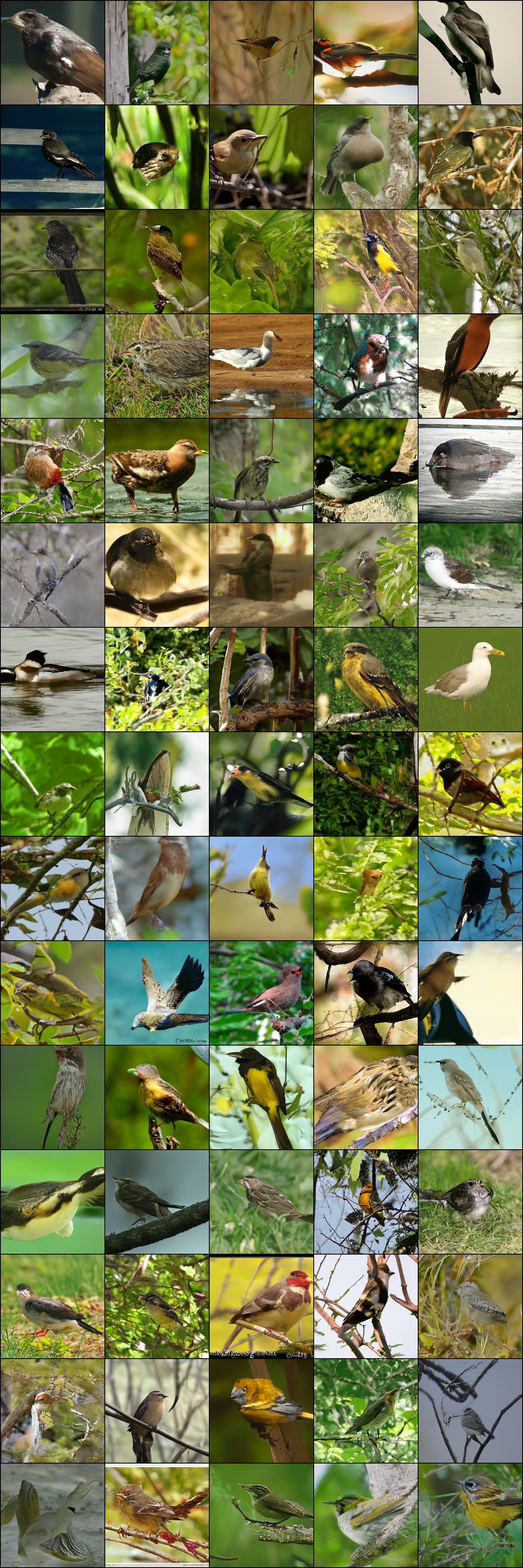}
  \caption{DiffScaler-CNN}
  \label{fig:sub62}
\end{subfigure}
\caption{Non-cherry picked Unconditonal generation samples on CUB-200 Dataset}
\label{fig:test61}
\end{figure}

\begin{figure}
\centering
\begin{subfigure}{.5\textwidth}
  \centering
  \includegraphics[width=0.95\linewidth]{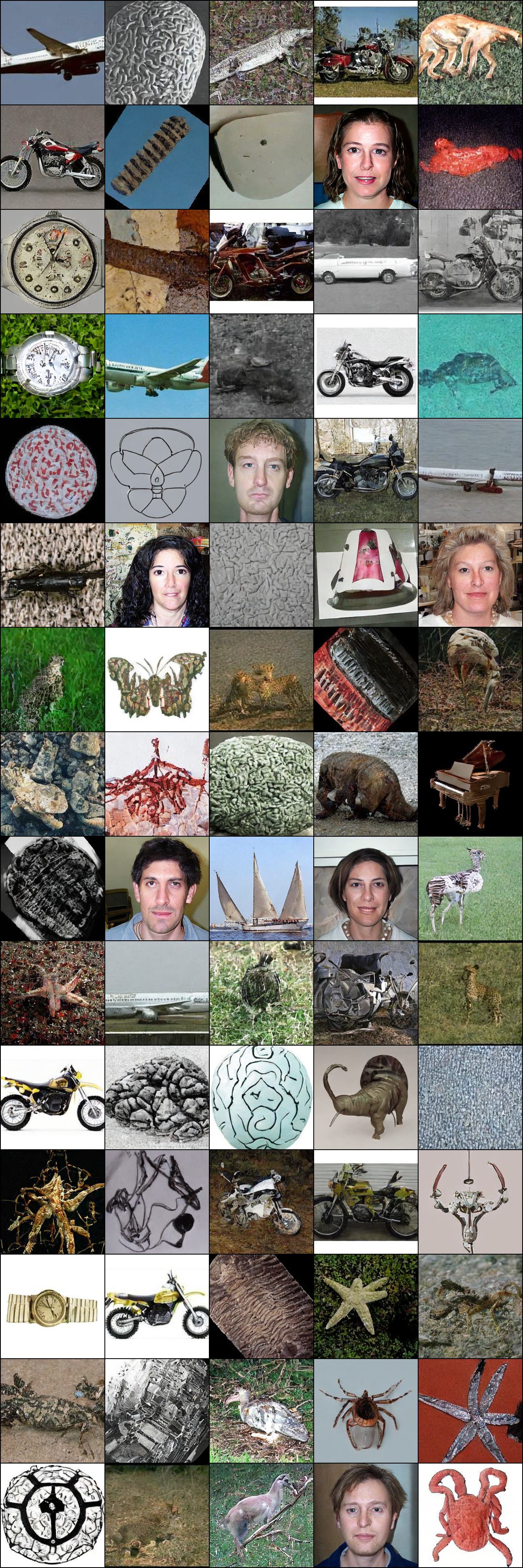}
  \caption{Diffscaler-DiT}
  \label{fig:sub71}
\end{subfigure}%
\begin{subfigure}{.5\textwidth}
  \centering
  \includegraphics[width=0.95\linewidth]{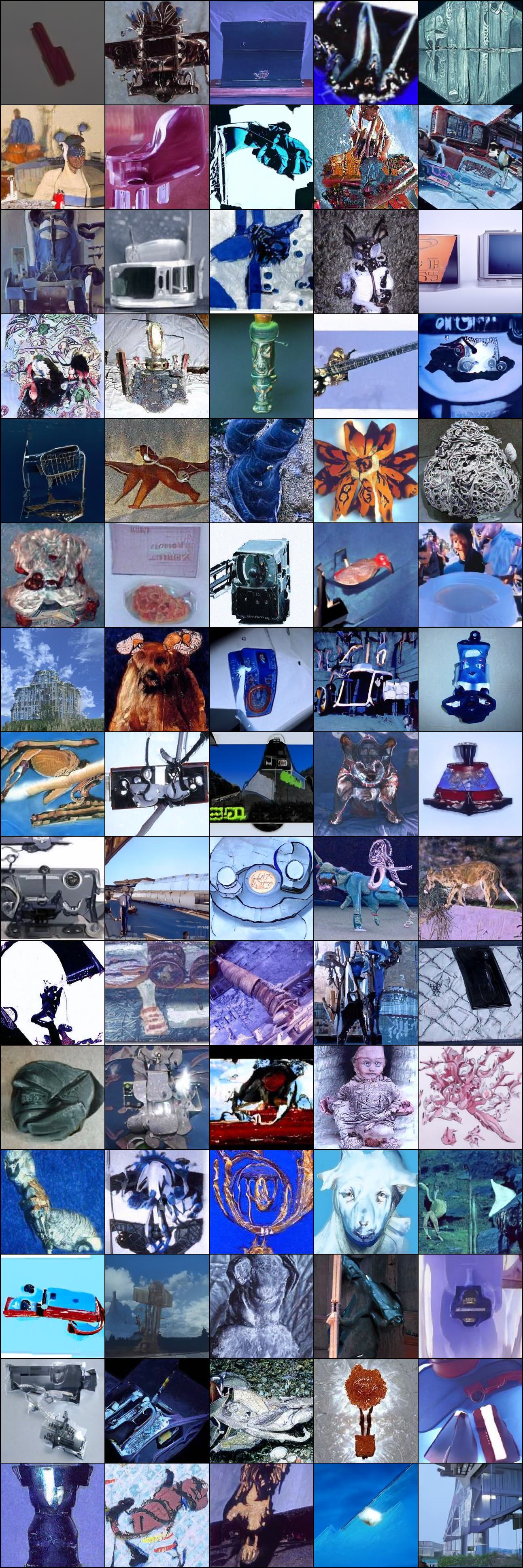}
  \caption{DiffScaler-CNN}
  \label{fig:sub72}
\end{subfigure}
\caption{Non-cherry picked Unconditonal generation samples on Caltech-101 Dataset}
\label{fig:test71}
\end{figure}
\begin{figure}
\centering
\begin{subfigure}{.5\textwidth}
  \centering
  \includegraphics[width=0.95\linewidth]{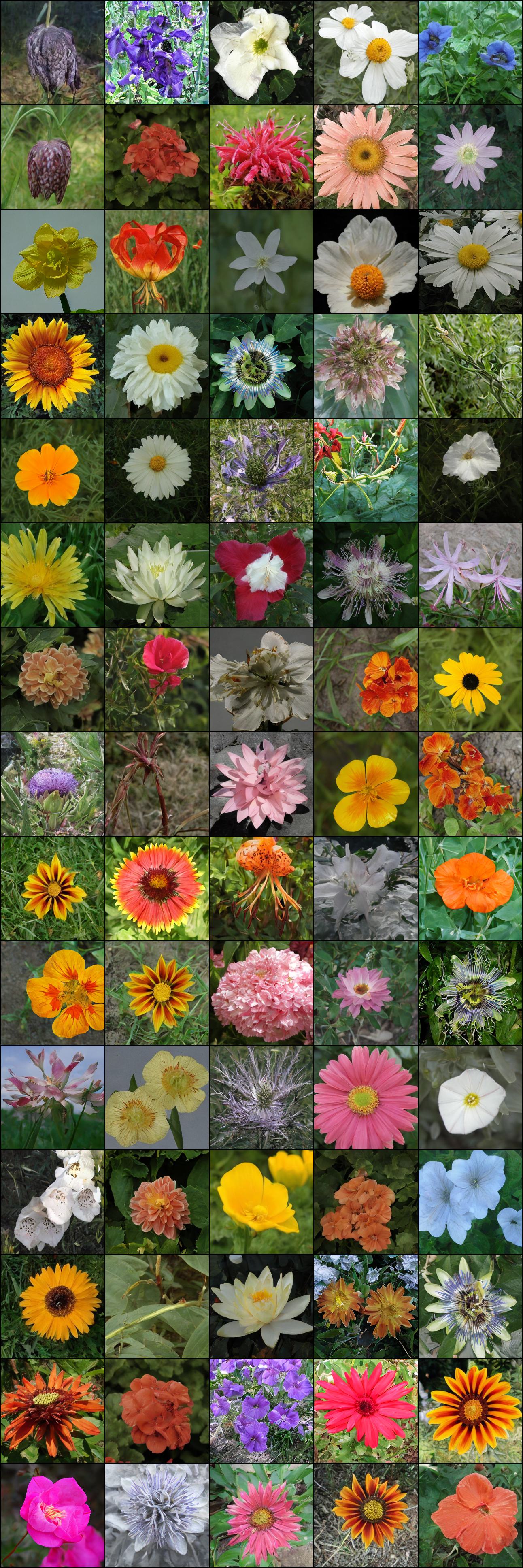}
  \caption{Diffscaler-DiT}
  \label{fig:sub81}
\end{subfigure}%
\begin{subfigure}{.5\textwidth}
  \centering
  \includegraphics[width=0.95\linewidth]{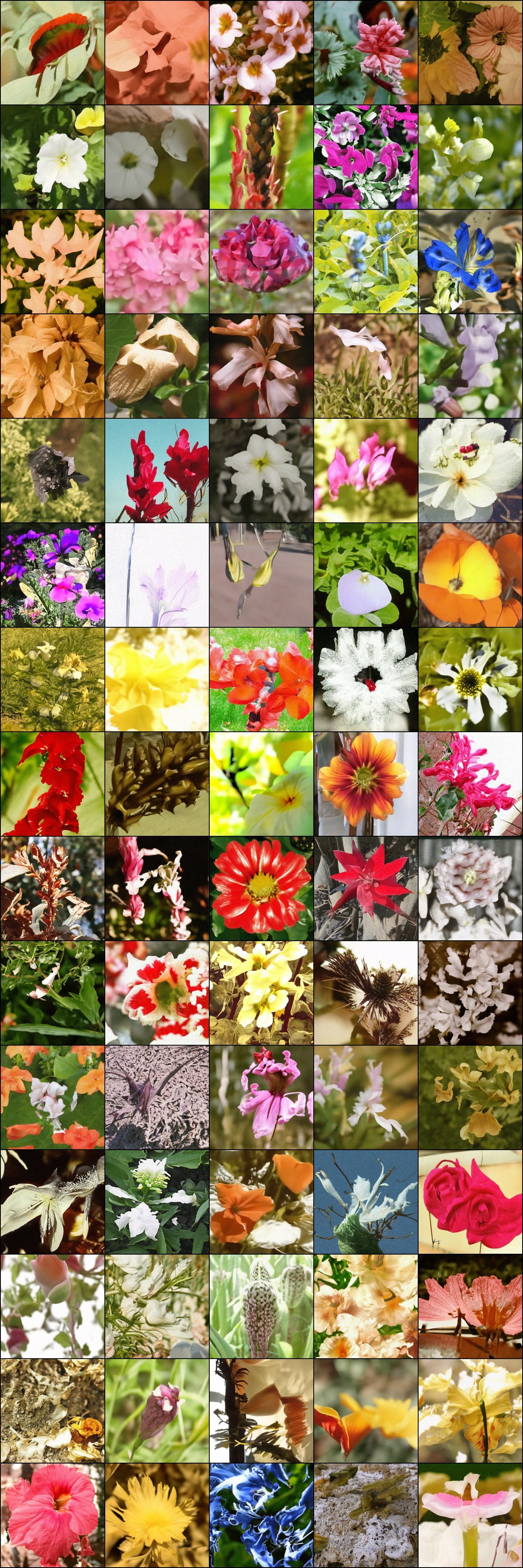}
  \caption{DiffScaler-CNN}
  \label{fig:sub82}
\end{subfigure}
\caption{Non-cherry picked Unconditonal generation samples on Oxford flowers Dataset}
\label{fig:test81}
\end{figure}

\begin{figure}
\centering
\begin{subfigure}{.5\textwidth}
  \centering
  \includegraphics[width=0.95\linewidth]{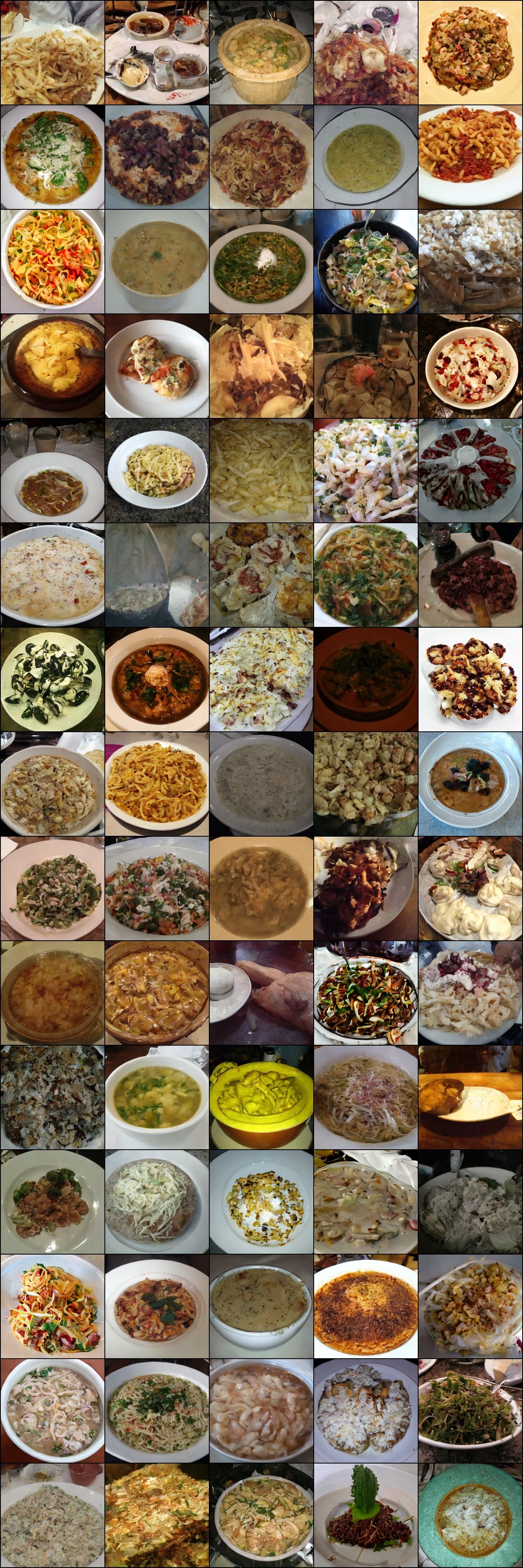}
  \caption{Diffscaler-DiT}
  \label{fig:sub91}
\end{subfigure}%
\begin{subfigure}{.5\textwidth}
  \centering
  \includegraphics[width=0.95\linewidth]{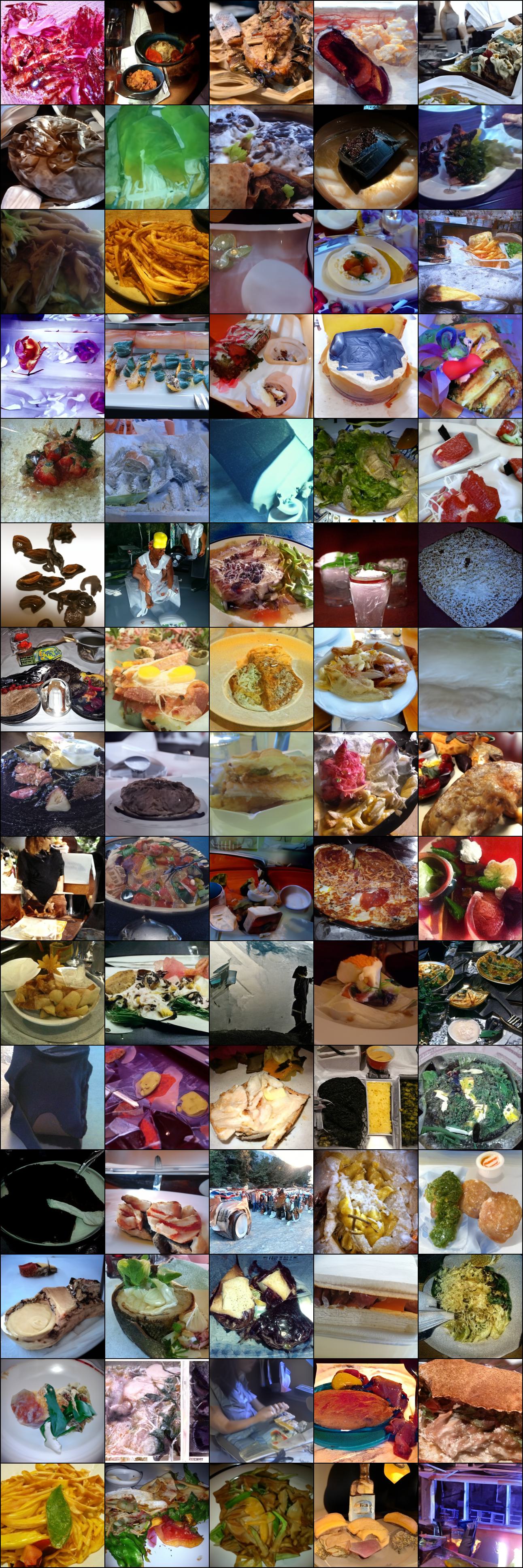}
  \caption{DiffScaler-CNN}
  \label{fig:sub92}
\end{subfigure}
\caption{Non-cherry picked Unconditonal generation samples on Food-101 Dataset}
\label{fig:test91}
\end{figure}


\begin{figure}
\centering
\begin{subfigure}{.5\textwidth}
  \centering
  \includegraphics[width=0.95\linewidth]{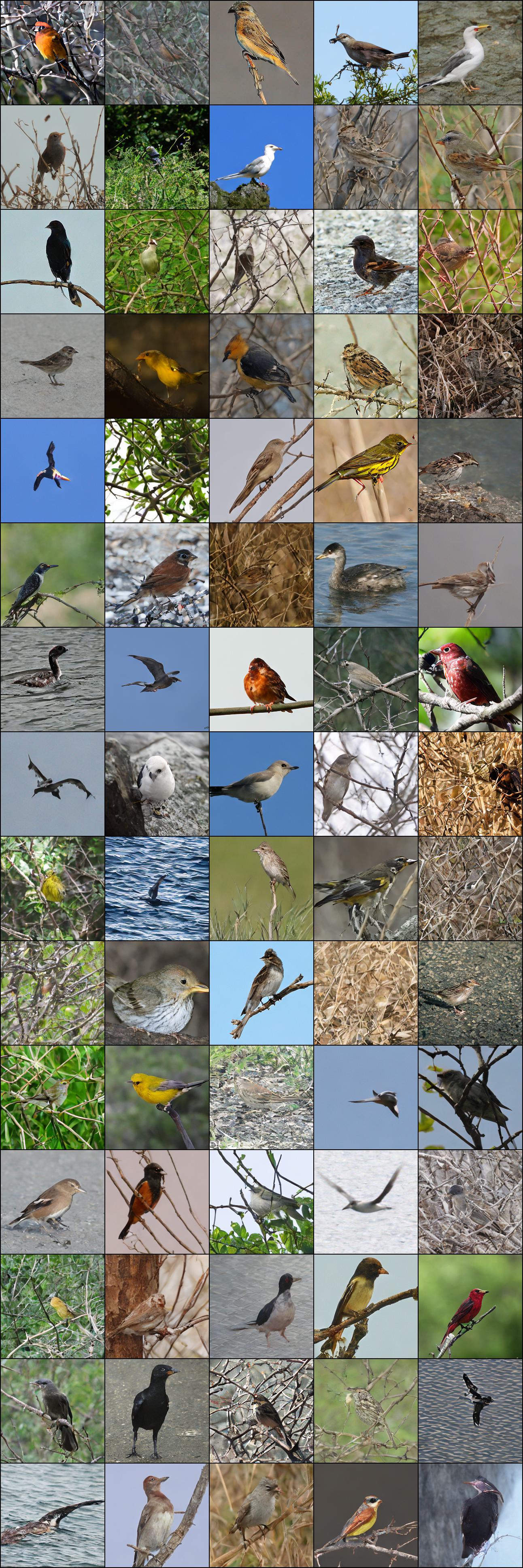}
  \caption{Diffscaler-DiT}
  \label{fig:sub101}
\end{subfigure}%
\begin{subfigure}{.5\textwidth}
  \centering
  \includegraphics[width=0.95\linewidth]{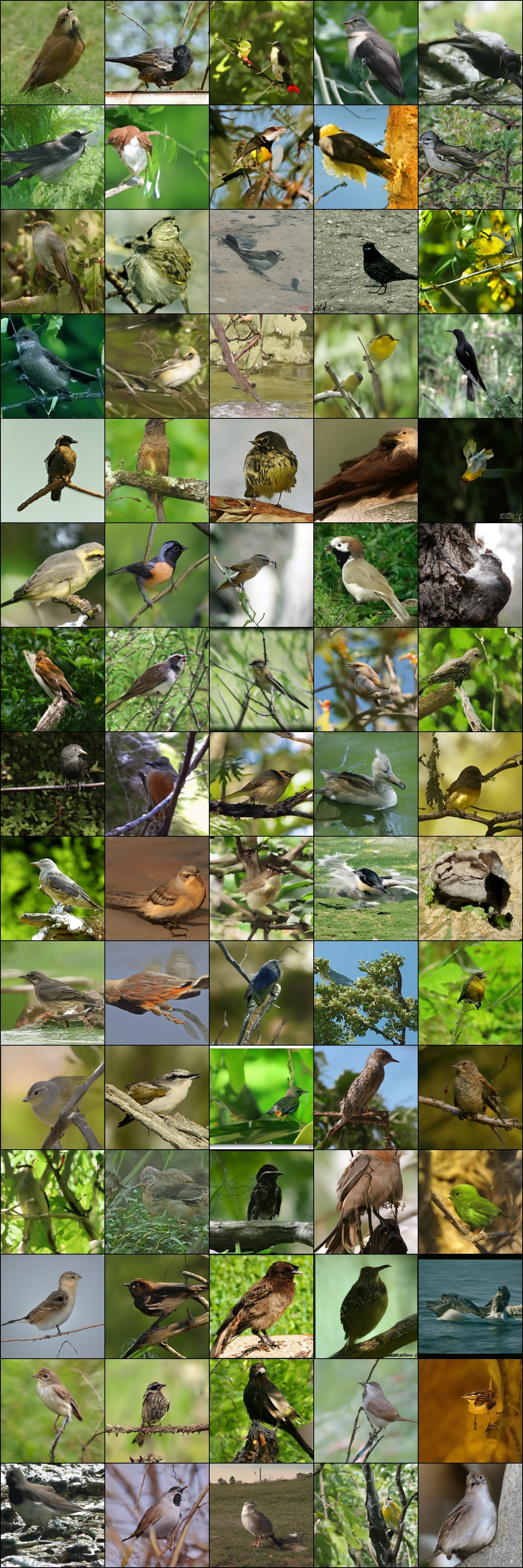}
  \caption{DiffScaler-CNN}
  \label{fig:sub102}
\end{subfigure}
\caption{Non-cherry picked Unconditonal generation samples on CUB-200 Dataset}
\label{fig:test101}
\end{figure}

\begin{figure}
\centering
\begin{subfigure}{.5\textwidth}
  \centering
  \includegraphics[width=0.95\linewidth]{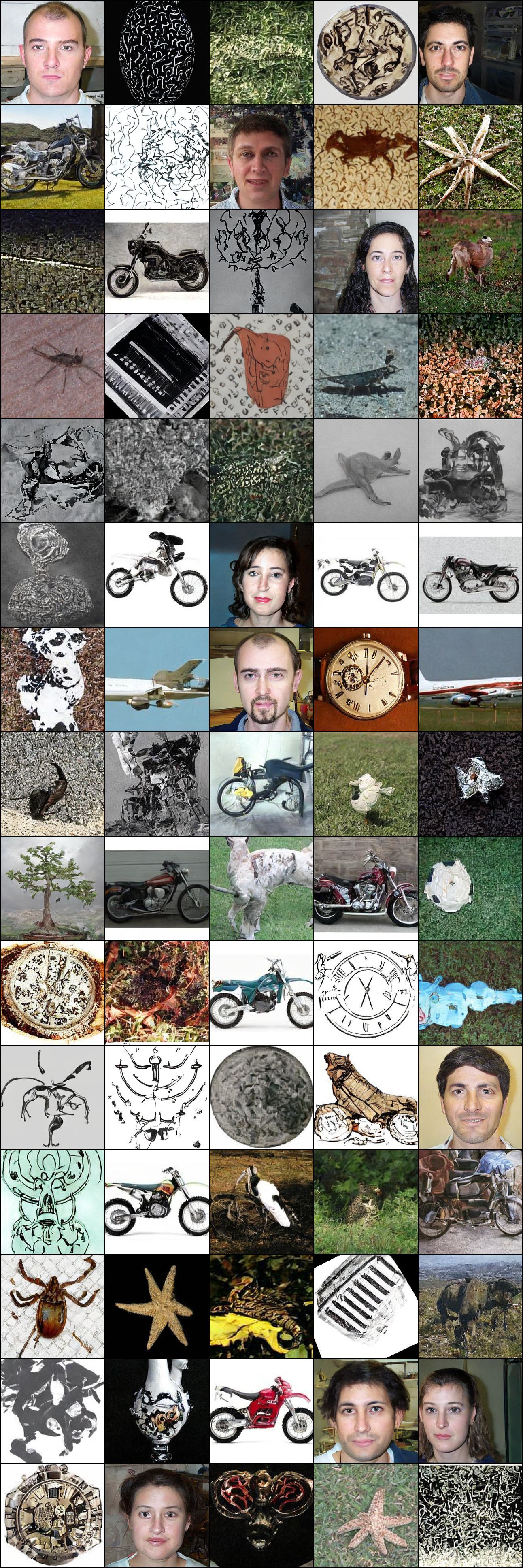}
  \caption{Diffscaler-DiT}
  \label{fig:sub111}
\end{subfigure}%
\begin{subfigure}{.5\textwidth}
  \centering
  \includegraphics[width=0.95\linewidth]{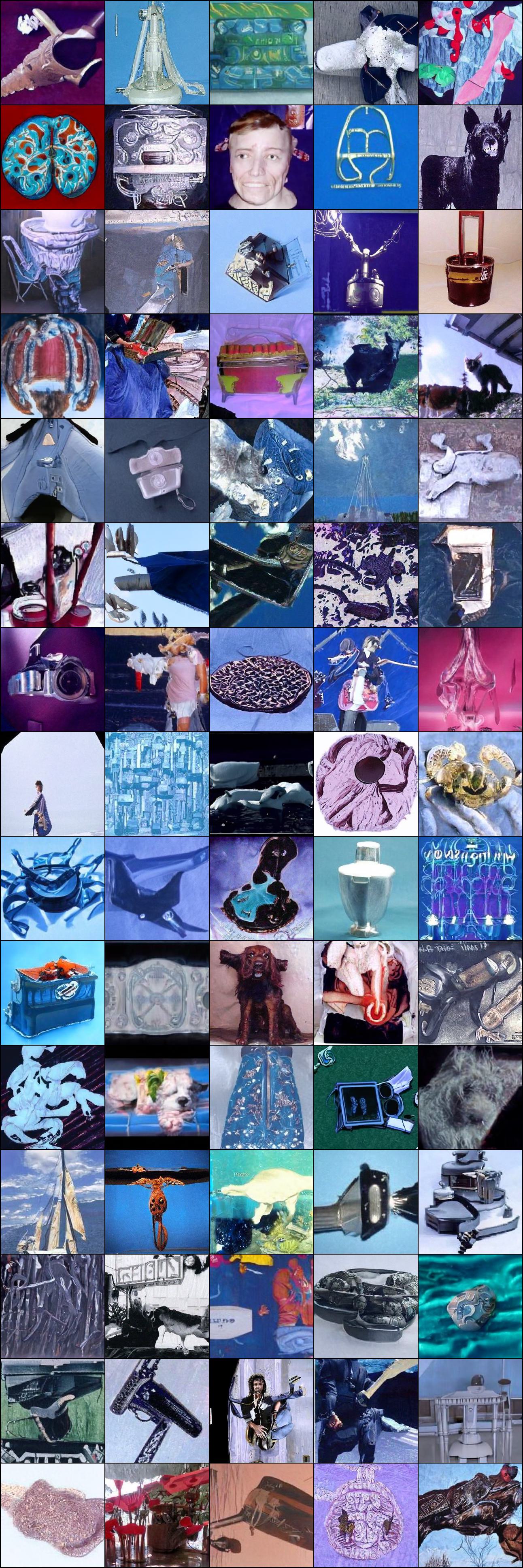}
  \caption{DiffScaler-CNN}
  \label{fig:sub112}
\end{subfigure}
\caption{Non-cherry picked Unconditonal generation samples on Caltech-101 Dataset}
\label{fig:test111}
\end{figure}
\begin{figure}
\centering
\begin{subfigure}{.5\textwidth}
  \centering
  \includegraphics[width=0.95\linewidth]{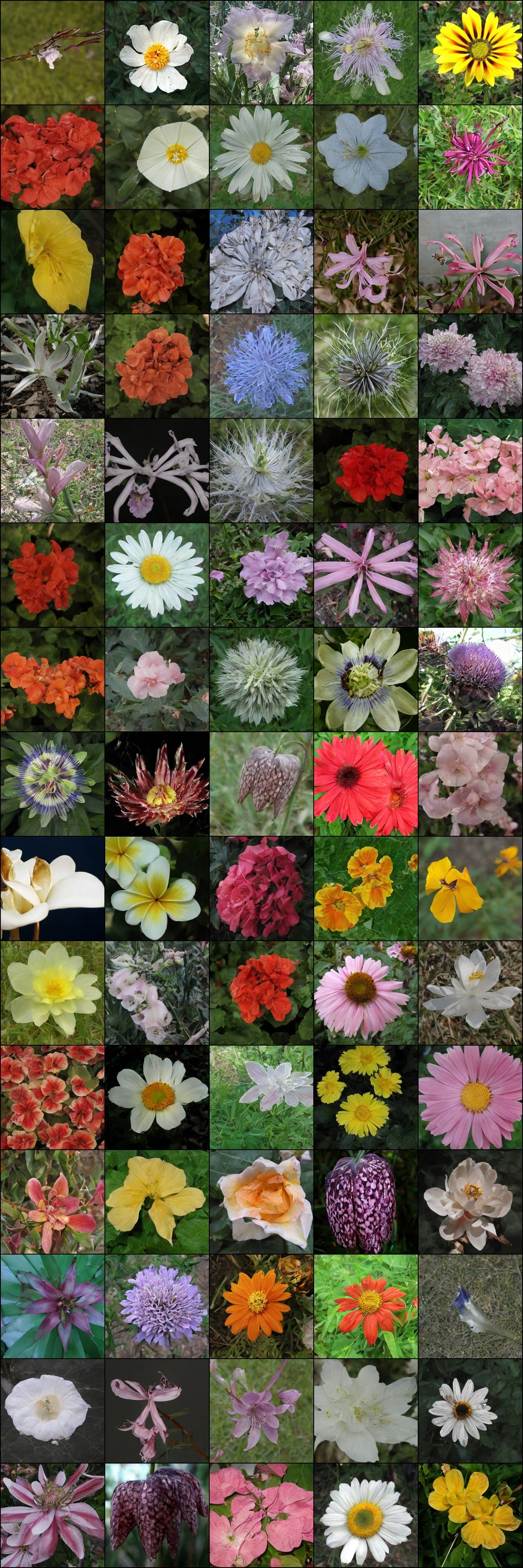}
  \caption{Diffscaler-DiT}
  \label{fig:sub121}
\end{subfigure}%
\begin{subfigure}{.5\textwidth}
  \centering
  \includegraphics[width=0.95\linewidth]{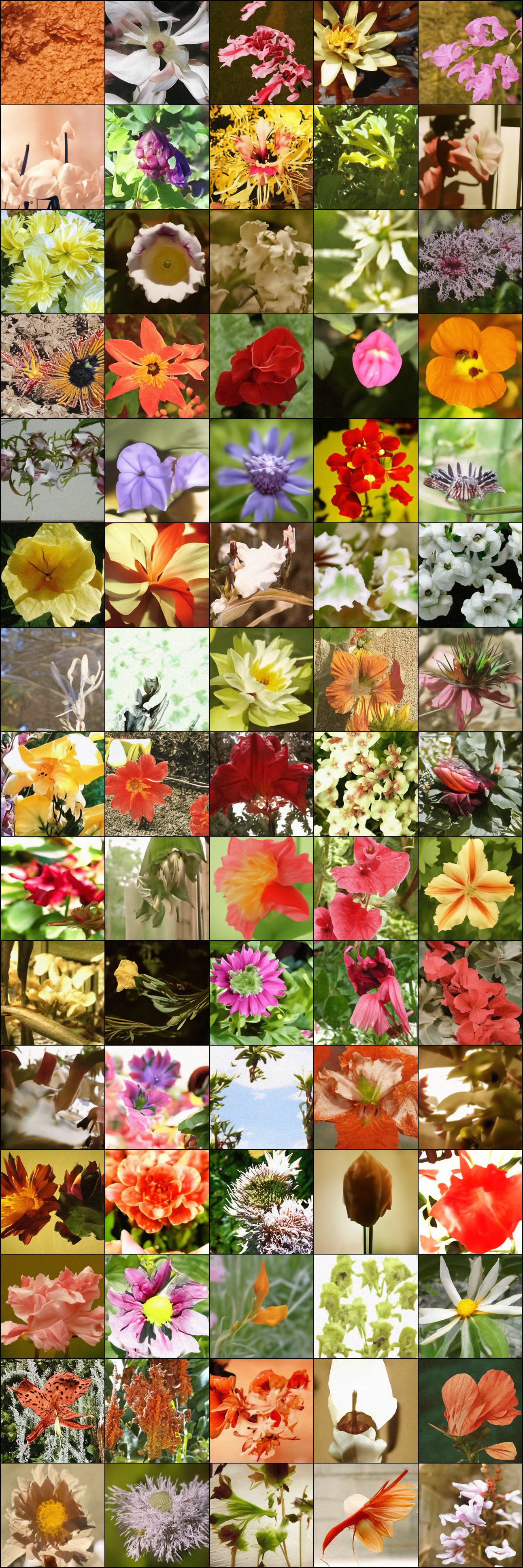}
  \caption{DiffScaler-CNN}
  \label{fig:sub122}
\end{subfigure}
\caption{Non-cherry picked Unconditonal generation samples on Oxford flowers Dataset}
\label{fig:test121}
\end{figure}

\begin{figure}
\centering
\begin{subfigure}{.5\textwidth}
  \centering
  \includegraphics[width=0.95\linewidth]{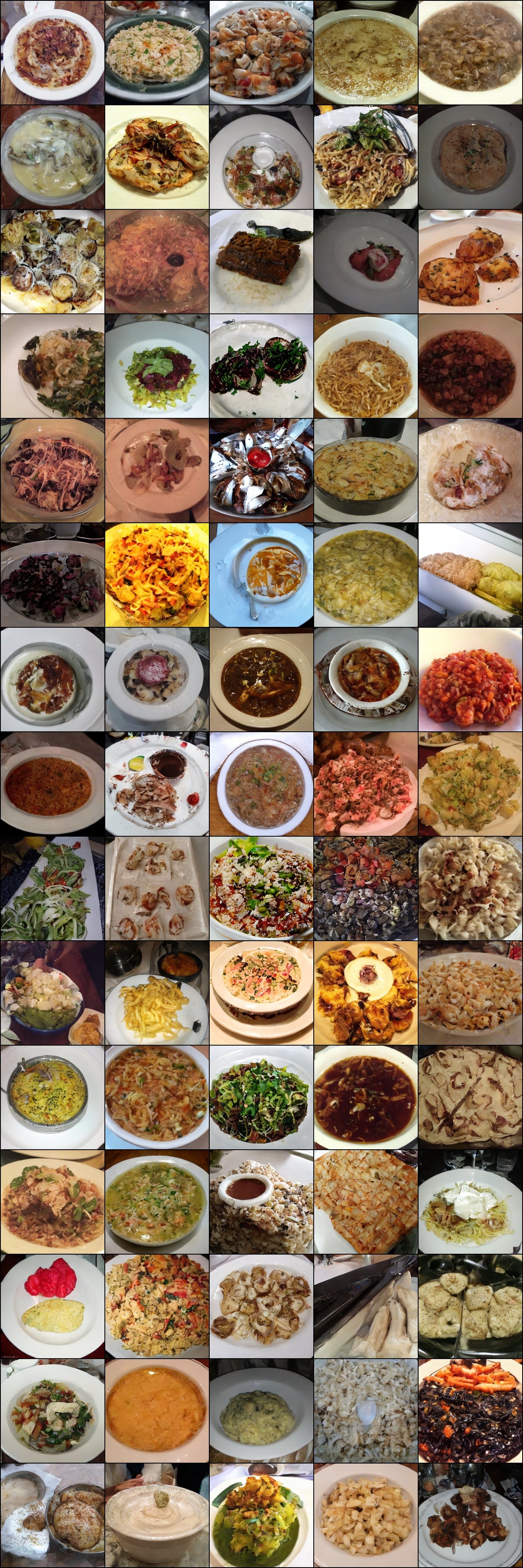}
  \caption{Diffscaler-DiT}
  \label{fig:sub131}
\end{subfigure}%
\begin{subfigure}{.5\textwidth}
  \centering
  \includegraphics[width=0.95\linewidth]{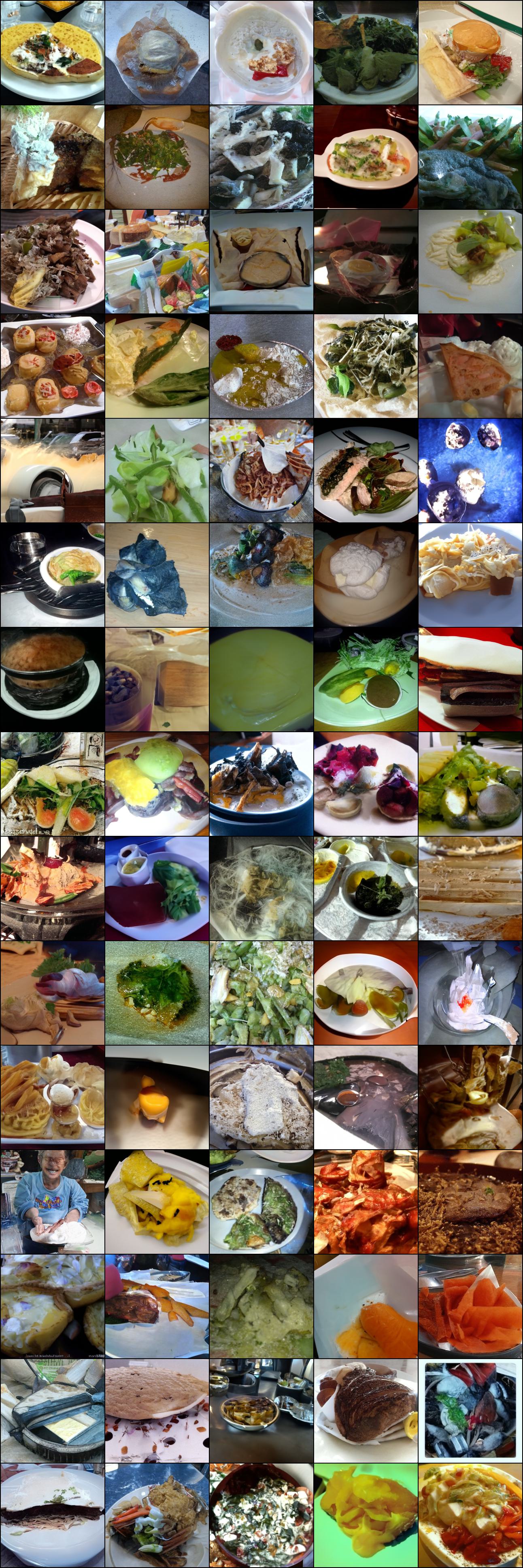}
  \caption{DiffScaler-CNN}
  \label{fig:sub132}
\end{subfigure}
\caption{Non-cherry picked Unconditonal generation samples on Food-101 Dataset}
\label{fig:test131}
\end{figure}


\end{document}